%% file: main.tex
\documentclass[journal]{IEEEtran}
\usepackage[hidelinks]{hyperref}
\usepackage{orcidlink}
\newcommand{\corrauthor}{\textsuperscript{\textdagger}}
\input{preamble}

\begin{document}

\title{Ordinal-Aware Calibration for Ordinal Classification}

\author{
Daehwan Kim\,\orcidlink{0009-0006-7856-7850},
Haejun Chung\corrauthor\,\orcidlink{0000-0001-8959-237X},
and Ikbeom Jang\corrauthor\,\orcidlink{0000-0002-6901-983X}%
\thanks{Daehwan Kim and Haejun Chung are with Hanyang University,
Seoul, Republic of Korea
(e-mail: officialhwan@hanyang.ac.kr; haejun@hanyang.ac.kr).}%
\thanks{Ikbeom Jang is with Hankuk University of Foreign Studies,
Yongin, Republic of Korea
(e-mail: ijang@hufs.ac.kr).}%
\thanks{\textsuperscript{\textdagger}Haejun Chung and Ikbeom Jang are the corresponding authors.}%
}
\maketitle

\begin{abstract}
Deep neural networks frequently produce overconfident, miscalibrated predictions. In ordinal classification, predictions must also adhere to a \emph{unimodal} and \emph{order-consistent} structure, a requirement that has dominated prior work while overlooking \emph{calibration}. 
We formalize this joint challenge as \emph{ordinal calibration} for the first time and propose the Ordinal loss for Calibration and Unimodality (ORCU). Unlike incremental modular combinations, ORCU provides a concise, principled unification of distance-aware soft encoding and an ordinal-aware log-barrier extension on adjacent logit gaps.
This coupling is theoretically synergetic: soft-encoded targets pre-condition the optimization landscape by providing a finite, full-support ordinal anchor for well-posed refinement, while the log-barrier extension preserves non-vanishing adjacent-gap gradients in the unimodality-feasible interior. This enables the log-barrier to refine confidence without disrupting the ordinal geometry. ORCU requires no architectural changes or post-hoc calibration. Across four ordinal benchmarks, it achieves state-of-the-art calibration without compromising accuracy. We establish a new reliability standard by providing a reproducible benchmark over 10 specialized and general-purpose loss functions. Our code is available at \href{https://github.com/labhai/ORCU}{https://github.com/labhai/ORCU}.
\end{abstract}


\input{main_body}
\balance
\bibliographystyle{IEEEtran}
\bibliography{main}
\newpage
\input{appendix}

\end{document}

%% file: preamble.tex
\usepackage[T1]{fontenc}
\usepackage{graphicx}
\usepackage{booktabs}
\usepackage{amsmath,amssymb,amsfonts,mathtools}
\usepackage{amsthm}
\usepackage{bm}
\usepackage{microtype}
\usepackage[dvipsnames,table]{xcolor}
\usepackage{multirow}
\usepackage{hhline}
\usepackage{diagbox}
\usepackage{array}
\usepackage{adjustbox}
\usepackage{enumitem}
\usepackage{colortbl}
\usepackage{pifont}
\usepackage{url}
\usepackage[caption=false,font=footnotesize]{subfig}
\usepackage{stfloats}
\usepackage{placeins}
\usepackage{balance}

\newtheorem{proposition}{Proposition}

\newtheorem{theorem}{Theorem}
\newtheorem{corollary}{Corollary}

\newsavebox\CBox
\newcommand*\textBF[1]{\sbox\CBox{#1}\resizebox{\wd\CBox}{\ht\CBox}{\textbf{#1}}}

\providecommand{\citep}[1]{\cite{#1}}
\providecommand{\Cref}[1]{\ref{#1}}

%% file: main_body.tex
\section{Introduction}
\label{sec:introduction}
Recent advances in ordinal classification (e.g., medical diagnosis and rating assessments) have largely pursued predictive accuracy by encoding label order into learning objectives. Yet ordered labels often quantify graded severity or quality in decision-critical settings, where downstream actions hinge not only on the predicted class but also on whether confidence reliably reflects correctness. In such settings, deep neural networks (DNNs) must output probabilities whose magnitudes match the true likelihood of being correct; strong ordinal accuracy can still coexist with systematically miscalibrated confidence \cite{guo2017calibration}. This motivates a central question: ``\textit{In ordered label spaces, is a prediction trustworthy if it is unimodal and order-consistent alone, or must its confidence also be well-calibrated?}'' In this work, we define the joint requirement of (i)~unimodal, order-consistent predictive distributions and (ii)~well-calibrated confidence as \emph{ordinal calibration}. Existing methods treat these requirements in isolation; ordinal objectives enforce ordered probability shapes without ensuring confidence reliability, whereas calibration-oriented objectives are order-agnostic and can violate ordinal structure, leaving their coupling unresolved in ordinal learning. To the best of our knowledge, this is the first study to explicitly pose and formalize \emph{ordinal calibration} in ordinal classification using a single, simple loss-level objective that jointly addresses calibration and ordinal structure.

Unlike nominal classification, ordinal classification assumes a natural ordering among class labels. Existing approaches, including regression-based~\cite{fu2008human, gu2014incremental, hamsici2015multiple, pan2018mean, ijcai2018p150, li2019bridgenet, zhong2023ordinal}, classification-based~\cite{fernandez2014ordinal, liu2020unimodal, polat2022class, vargas2022unimodal, rosati2024learning}, and ranking-based approaches~\cite{Niu_2016_CVPR, chen2017using, cao2020rank, shi2023deep} capture this structure, but largely overlook calibration and can yield miscalibrated confidence estimates. This limitation is critical because ordinal prediction tasks inherently require well-calibrated probability estimates~\cite{jiang2012calibrating,neumann2018relaxed,moon2020confidence,kompa2021second}, thereby motivating us to revisit calibration explicitly under ordinal constraints.

Calibration aligns predicted confidence with the true likelihood of correctness, formally \(P\bigl(Y = y \mid \hat{P} = p\bigr) = p\), where \(\hat{P}\) denotes the model's confidence. In high-stakes applications, such as medical diagnosis and autonomous driving, well-calibrated confidence allows low-confidence cases to trigger additional verification, supporting safety and reliability; this consideration is particularly important for ordinal labels, where decisions rely on trustworthy probability estimates.

Beyond calibration, unimodality in output distributions serves as a vital inductive bias for robust ordinal classification. A standard modeling assumption is that the target conditional distribution $P_{Y \mid X}$ is \emph{unimodal} \cite{li2022unimodal,vargas2022unimodal,dey2024conformal,yamasaki2025approximately}, where probabilities decrease monotonically away from the mode. While complex real-world cues can rarely exhibit multimodality \cite{shen2017label}, unconstrained models often produce irrational contradictions--e.g., assigning the highest and 2nd highest probabilities to distant labels. Treating unimodality as a critical inductive bias rather than a hard architectural constraint can thus prevent such inconsistencies, ensuring that predictive mass remains concentrated around the most likely label, while maintaining the flexibility required to model ambiguous or complex samples.

\input{figures/tex/fig1_intro}
These considerations motivate enforcing \emph{ordinal calibration} at training time.
Fig.~\ref{figure:intro} contrasts the failure modes of order-agnostic calibration and calibration-agnostic ordinal objectives, indicating the need for a single loss that refines confidence while preserving unimodal ordering. We therefore propose \emph{\textbf{OR}dinal loss for \textbf{C}alibration and \textbf{U}nimodality} (ORCU), a loss-level objective that couples distance-aware soft encoding with an ordinal log-barrier regularizer on adjacent logit gaps, providing non-vanishing feasible-interior gradients for confidence refinement under unimodality constraints. Defined purely at the loss level, ORCU leaves the architecture and training procedure intact and obviates post-hoc calibration, yielding credible predictions in safety-critical settings. The appendix includes theoretical derivations, implementation details, and extended results.

We make the following \textbf{key contributions}: 
(i)~We formally define \emph{ordinal calibration} as a distinct criterion for ordered label spaces, requiring simultaneous confidence alignment and structured unimodality. 
(ii)~We propose ORCU, a loss function that enforces calibration and unimodality in classification through a principled coupling of distance-aware soft encoding and an ordinal-aware regularizer. We provide a rigorous gradient analysis demonstrating that this synergy overcomes the optimization bottlenecks typical of log-barrier methods in high-dimensional spaces, enabling stable structural refinement without post-hoc or architectural overhead.
(iii)~We establish the first comprehensive calibration benchmark for ordinal classification, evaluating 10 existing loss functions across four public datasets to provide a reproducible foundation for future research in trustworthy ordinal classification.

\vspace{-0.2cm}

\section{Related Work}
\label{sec:relatedworks}

\subsection{Formulations for Ordinal Classification}
Predicting a target involves mapping an input $x \in \mathcal{X}$ to an output $y \in \mathcal{Y}$ via a function $\mathcal{M}: \mathcal{X} \rightarrow \mathcal{Y}$. In nominal classification, labels in $\mathcal{Y}$ are treated as unordered categories. In contrast, ordinal classification assumes a natural ordering among labels, denoted as $y_1 \prec y_2 \prec \dots \prec y_C$ (where $\prec$ denotes class ordering). Previous studies have proposed three main approaches to incorporate this structure into model design: regression-based, ranking-based, and classification-based methods. Regression-based methods such as cumulative link models \cite{fu2008human, pan2018mean, ijcai2018p150, li2019bridgenet} treat labels as continuous targets and minimize $\ell_1$ or $\ell_2$ loss. However, their scalar outputs lack probabilistic semantics, making them unsuitable for confidence estimation. Ranking-based methods such as CORAL and CORN \cite{Niu_2016_CVPR, chen2017using, cao2020rank, shi2023deep} decompose the task into binary comparisons to model label order, but their threshold-based design does not yield normalized class probabilities, limiting their applicability to calibration. In contrast, classification-based methods \cite{liu2020unimodal, polat2022class, vargas2022unimodal} treat ordinal categories as discrete classes and use a softmax classifier to produce class-wise probabilities. This enables ordinal structure learning and meaningful confidence estimation while providing interpretable predictions. Therefore, we adopt the classification-based formulation, which supports probabilistic calibration while respecting ordinal semantics.

\subsection{Ordinal Losses Neglecting Calibration}
The limitations of Cross-Entropy (CE) loss in modeling ordinal structure have driven several adaptations within classification-based ordinal regression. Soft ORDinal (SORD) encoding \cite{diaz2019soft} introduces a soft label distribution that reflects class proximity, encouraging smoother predictions aligned with label order. Class Distance Weighted CE (CDW-CE) \cite{polat2022class} applies distance-based penalties to discourage large deviations from the target label. In contrast, CO2 \cite{albuquerque2021ordinal} incorporates a regularization term to promote unimodality in the predicted distribution. Probabilistic Ordinal Embeddings (POE) \cite{li2021learning} further combine probabilistic label representations with architectural changes to encode uncertainty and order jointly. While these methods enhance ordinal structure or predictive smoothness, they largely overlook calibration\textemdash despite its critical role in ensuring reliable confidence estimates for ordinal tasks. We address this gap by reframing ordinal classification from a calibration-centric standpoint, proposing a principled approach that jointly enforces ordinal structure and confidence reliability.

\subsection{Calibration Losses Neglecting Ordinality}
Calibration, which aligns predicted confidence with actual accuracy, is typically addressed through either post-hoc \cite{platt1999probabilistic, guo2017calibration, tomani2021post, tomani2022parameterized} or regularization-based approaches. It is increasingly considered a crucial factor\textemdash sometimes more important than accuracy\textemdash in high-stakes multi-class classification tasks~\cite{guo2017calibration, minderer2021revisiting, dong2025survey}. Post-hoc techniques recalibrate predictions using a hold-out validation set, whereas regularization embeds calibration objectives directly into the training loss. Such regularizers include Label Smoothing (LS) \cite{szegedy2016rethinking}, which softens targets to mitigate overconfidence; FLSD \cite{mukhoti2020calibrating} re-weights samples by prediction confidence; MbLS \cite{liu2022devil} and MDCA \cite{hebbalaguppe2022stitch} extend margin-based smoothing over the entire logit space; and ACLS \cite{park2023acls} further adapts the smoothing strength per sample. However, the calibration studies to date have focused on nominal classification and have not accounted for ordinal structure. In ordinal label spaces, these objectives are typically order-agnostic, so unimodality and order-consistent probability shapes are not guaranteed and \emph{ordinal calibration} remains unaddressed. We introduce the first calibration-driven perspective on ordinal classification and argue for calibration at the loss level during training rather than via post-hoc procedures.

\vspace{-0.2cm}

\section{Unified Loss for Ordinal Calibration}
\label{sec:methods}
Ordinal calibration requires an ordered probability shape and confidence that remains responsive to input evidence. Since both arise from the same logits, they should be treated jointly rather than as independent objectives. A fixed unimodal target provides order but restricts sample-wise confidence, whereas order-agnostic confidence adjustment can distort the ordered shape. ORCU couples these roles at the loss level. $\mathcal L_{\text{SoftCE}}$ provides a finite, distance-aware unimodal target, while $\mathcal L_{\text{LBE}}$ keeps adjacent-logit update directions active even after the ordinal inequalities become feasible. We first derive this coupling and then analyze how it refines confidence within the resulting ordinal geometry.

\begin{figure*}[!t]
\begin{center}
\includegraphics[width=0.95\textwidth]{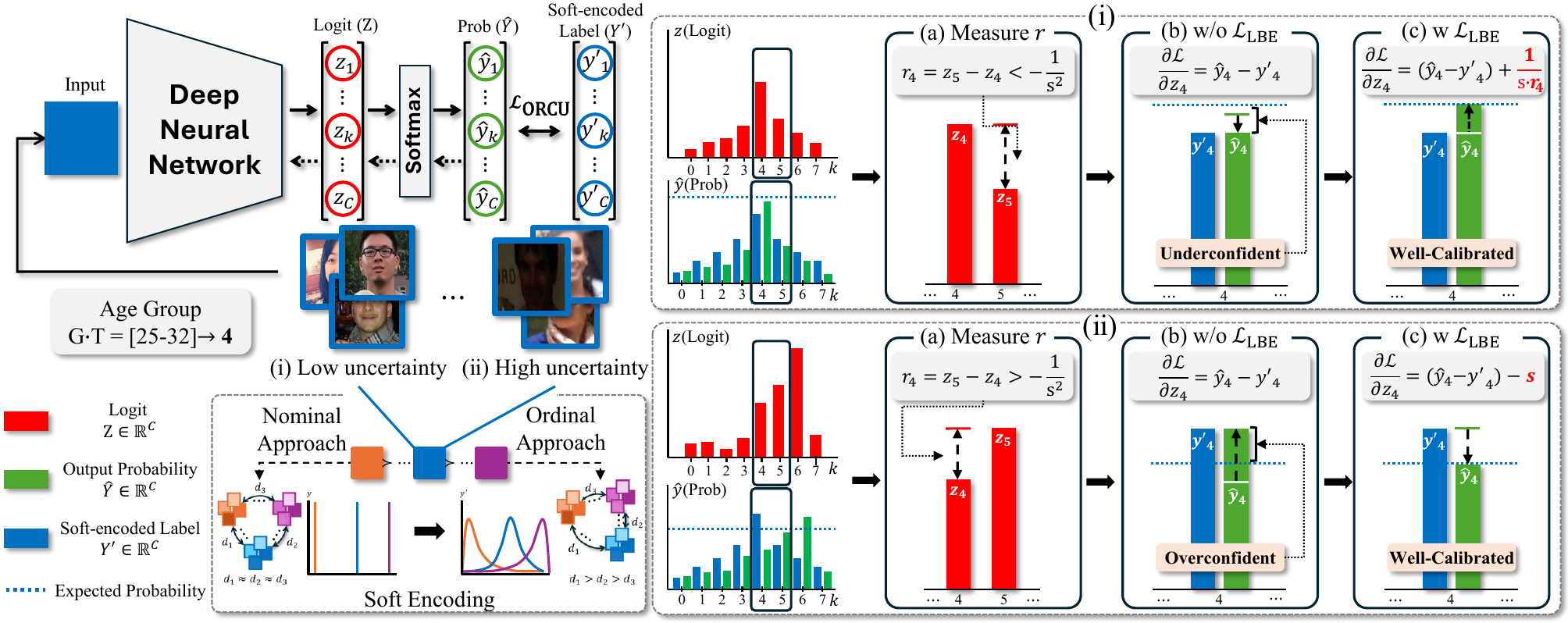}
\end{center}
\vspace{-0.3cm}
\caption{Illustration of ORCU on Adience (8 age classes) for two adjacent-logit configurations, both for $k\ge y_n$. Each case shows (a) the directed gap $r$ as an input-specific ordinal uncertainty proxy and its LBE response, (b) the update without $\mathcal L_{\text{LBE}}$, which remains tied to the fixed soft target, and (c) the LBE-assisted update through the same adjacent relation. The gap $r$ indicates whether the local ordering around the target is well separated, near-boundary, or violated. The expected-probability bars schematically illustrate confidence correction. Section~\ref{sec:cal_alignment} evaluates this refinement, while Section~\ref{sec:gradAnal} and Table~\ref{table:gradient} provide the gradient analysis.}
\vspace{-0.4cm}
\label{figure:main}
\end{figure*}

\subsection{Cross-Entropy Limitations and Ordinal Soft Encoding}
\label{subsec:limitation_of_ce}
\label{subsec:sord}
One-hot cross-entropy (CE) is not a suitable probability target for ordinal calibration because it ignores class distance and has no finite logit-space optimum. Let $N$ be the number of samples and $C$ the number of classes. For sample $n$, let $y_n\in\{1,\ldots,C\}$ be the true class, $\mathbf y_n\in\mathbb R^C$ its one-hot vector, $\mathbf z_n=[z_{n,1},\ldots,z_{n,C}]\in\mathbb R^C$ the logits, and $\hat{\mathbf y}_n=\operatorname{softmax}(\mathbf z_n)$. The loss $\mathcal L_{\mathrm{CE}}=-\sum_{n=1}^N\sum_{k=1}^C y_{n,k}\log\hat y_{n,k}$ has per-sample gradient $\partial\mathcal L_{\mathrm{CE}}/\partial z_{n,k}=\hat y_{n,k}-y_{n,k}$. For $k\neq y_n$, a descent step taken directly in logit space increases the target-to-nontarget margin by $\eta[(1-\hat y_{n,y_n})+\hat y_{n,k}]>0$. The infimum is approached only as $\hat y_{n,y_n}\to1$, which requires the relative margins to diverge and drives the prediction toward a near-delta distribution (Appx.~IV-A). Thus, one-hot CE neither distinguishes adjacent from distant classes nor provides a finite ordinal probability target.

Distance-aware soft encoding provides such a target. We adopt SORD encoding~\cite{diaz2019soft} and replace the one-hot label with a distribution whose mass decays with ordinal distance from $y_n$. Specifically,
{
\begin{equation}
\label{eq:sord_target}
y'_{n,k}
=\frac{\exp[-\phi(y_n,k)]}{\sum_{j=1}^C\exp[-\phi(y_n,j)]},
\qquad
\phi(y_n,k)=(y_n-k)^2.
\end{equation}}
The resulting vector $\mathbf y'_n\in\mathbb R^C$ is a valid probability distribution that is strictly unimodal with a unique mode at $y_n$. The Soft-Encoded Cross-Entropy (SoftCE) loss is
{
\begin{equation}
\label{eq:softce}
\mathcal{L}_{\text{SoftCE}}
= -\sum_{n=1}^N \sum_{k=1}^C y'_{n,k}\log\bigl(\hat y_{n,k}\bigr).
\end{equation}}
Its gradient, $\hat y_{n,k}-y'_{n,k}$, vanishes at the finite probability optimum $\hat{\mathbf y}_n=\mathbf y'_n$. Moreover, $\mathcal L_{\text{SoftCE}}$ is strictly proper with respect to $\mathbf y'_n$ (Appx.~IV-B), so it provides a well-defined unimodal probability target rather than an unattainable simplex vertex. This full-support anchor makes barrier-based ordinal refinement well-posed. Without such an anchor, the barrier term remains unbounded below along ordinal rays that sharpen the logit profile around $y_n$ (Appx.~IV-B). However, $\mathbf y'_n$ is fixed by the class label and distance metric. Samples sharing the same label are therefore driven toward the same distributional width regardless of evidence strength, which can narrow the attainable confidence range and lead to underconfidence. SoftCE supplies the required ordinal anchor, but not the sample-specific confidence adaptation needed within this induced ordinal geometry.

\subsection{Ordinal-Aware Log-Barrier Regularization}
\label{subsec:regularization}
The remaining task is to make confidence adaptable without losing the ordering already encoded by SoftCE. Since the required ordering is expressed through adjacent relations, we define the directed gap
\begin{equation}
\label{eq:directed_gap}
r_{n,k}=
\begin{cases}
z_{n,k}-z_{n,k+1}, & k<y_n,\\
z_{n,k+1}-z_{n,k}, & k\ge y_n.
\end{cases}
\end{equation}
By construction, $r_{n,k}\le0$ if and only if the adjacent pair follows the required monotone direction away from $y_n$. The boundary $r_{n,k}=0$ separates ordered and violating adjacent relations. More negative values indicate a larger ordering margin, negative values close to zero indicate weak local ordering, and positive values indicate an ordinal violation. Since this signed gap is computed from the current sample logits, it serves as a sample-specific ordinal uncertainty proxy through which the regularizer adapts its response.

A useful regularizer for these adjacent ordinal constraints must remain active at finite feasible gaps, attenuate its response for well-separated ordered pairs, strengthen the correction near the ordinal boundary, and avoid unbounded gradients after a violation. We use the log-barrier extension (LBE)~\cite{belharbi2019non, kervadec2022constrained} on the directed adjacent gaps in Eq.~\eqref{eq:directed_gap}, which gives the following ordinal regularization term
\begin{equation}
\label{eq:lbe}
\mathcal{L}_{\text{LBE}}
= \sum_{n=1}^N \sum_{k=1}^{C-1}
\begin{cases}
\hat I\bigl(z_{n,k}-z_{n,k+1}\bigr), & k<y_n,\\
\hat I\bigl(z_{n,k+1}-z_{n,k}\bigr), & k\ge y_n,
\end{cases}
\end{equation}
\begin{equation}
\label{eq:lbe_function}
\hat I(r)=
\begin{cases}
-\dfrac{1}{s}\log(-r), & r\le-\dfrac{1}{s^2},\\[3pt]
sr-\dfrac{1}{s}\log\!\left(\dfrac{1}{s^2}\right)+\dfrac{1}{s}, & \text{otherwise}.
\end{cases}
\end{equation}
In the barrier branch $r\le-1/s^2$, $\hat I'(r)=-1/(sr)$ is nonzero at every finite gap and decays algebraically as the pair moves deeper into the feasible region. As $r\to-1/s^2$ from below, its magnitude increases continuously to $s$, placing a stronger correction on relations with small ordering margins. For $r>-1/s^2$, the linear continuation fixes the slope at $s$, providing a bounded correction for near-boundary and violating pairs. The barrier scale $s$ controls both the transition margin and the maximum corrective slope. Thus, LBE supplies the finite-gap update signal, while the directed gaps align that signal with the ordinal direction away from $y_n$.

\noindent\textbf{ORdinal Loss for Calibration and Unimodality (ORCU).}
The complete objective is
{
\begin{equation}
\label{eq:orcu}
\mathcal L_{\text{ORCU}}=\mathcal L_{\text{SoftCE}}+\mathcal L_{\text{LBE}}.
\end{equation}}
The two terms have distinct roles. SoftCE determines where probability mass should lie across the ordered classes, while LBE changes relative adjacent logits according to the current sample. Their joint optimization retains the distance-aware ordinal reference while allowing the output width to vary across inputs. This is the intended coupling, rather than an order-agnostic confidence penalty appended to an ordinal loss.

\subsection{Gradient Analysis for Ordinal Calibration}
\label{sec:gradAnal}
\begin{figure*}[!t]
\centering
\subfloat[Interior $|I'|$]{%
    \includegraphics[width=0.235\textwidth]{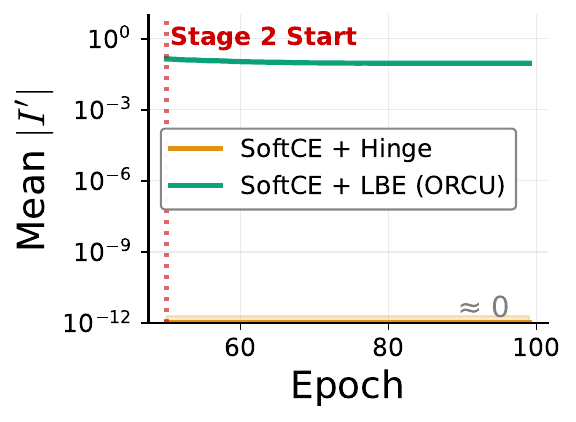}%
    \label{fig:sub_a}}
\hfill
\subfloat[Zero-grad (\%)]{%
    \includegraphics[width=0.235\textwidth]{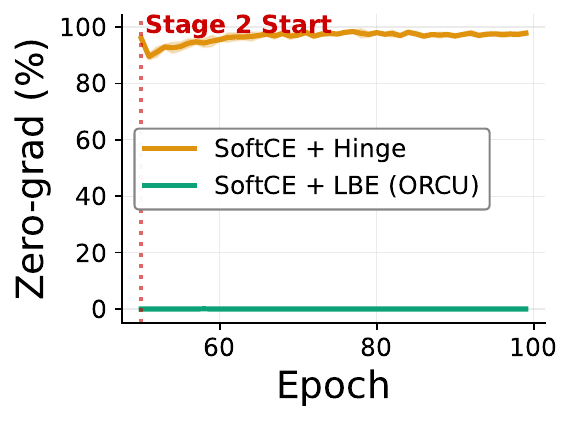}%
    \label{fig:sub_b}}
\hfill
\subfloat[Validation SCE]{%
    \includegraphics[width=0.235\textwidth]{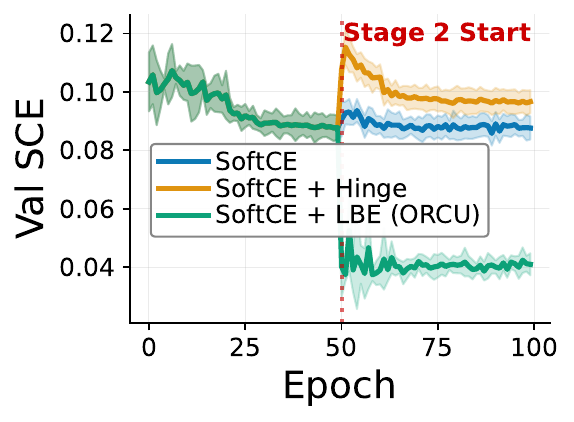}%
    \label{fig:sub_c}}
\hfill
\subfloat[Target-feasible SCE]{%
    \includegraphics[width=0.235\textwidth]{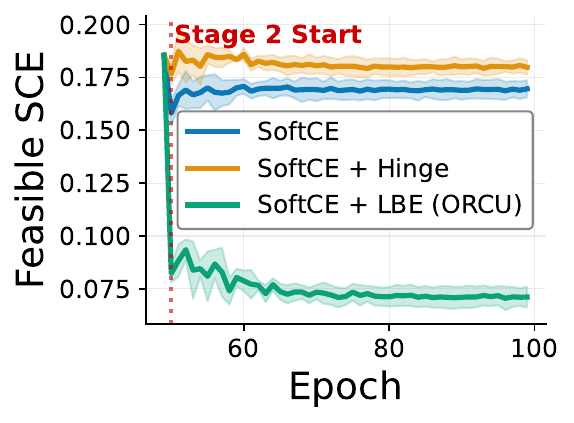}%
    \label{fig:sub_d}}
\caption{Two-stage continuation diagnostic. This continuation is used only for analysis; the main ORCU experiments jointly optimize both loss terms. Stage~1 trains with $\mathcal L_{\text{SoftCE}}$ for 50 epochs, and Stage~2 continues from the same checkpoint for 50 epochs with $\mathcal L_{\text{SoftCE}}$, $\mathcal L_{\text{SoftCE}}+\mathcal L_{\text{Hinge}}$, or $\mathcal L_{\text{SoftCE}}+\mathcal L_{\text{LBE}}$ (ORCU). The set $\mathcal F=\{n:r_{n,k}\le-1/s^2,\ \forall k\}$ is selected at Stage~1 and frozen across branches. The plots report (a) train mean interior $|I'(r)|$, (b) train zero-gradient rate of the regularizer w.r.t. logits, (c) full-validation SCE, and (d) SCE on $\mathcal F$. Hinge saturates, whereas LBE preserves interior gradients and lowers both full-validation and target-feasible SCE.}
\vspace{-0.2cm}
\label{fig:2stage}
\end{figure*}

The choice of LBE becomes most apparent after the ordinal inequalities are satisfied. A hinge penalty on the directed gap, $\hat I_{\mathrm{hinge}}(r)=\max(0,r+1/s^2)$, has zero derivative for $r<-1/s^2$ and therefore becomes inactive once most pairs enter the strict interior. Smooth penalties such as SoftPlus~\cite{dugas2000incorporating} avoid exact saturation, but their gradients decay exponentially as $r\to-\infty$ and become negligible when unimodality is strongly satisfied (Table~\ref{table:ablation}). LBE instead retains the algebraically decaying interior signal defined above. Starting from the same SoftCE checkpoint, Fig.~\ref{fig:2stage}~(a,b) confirms persistent gradient coverage for LBE and near-complete saturation for the hinge; Fig.~\ref{fig:2stage}~(c,d) and Section~\ref{sec:cal_alignment} then evaluate whether this active interior signal translates into confidence refinement.

The coupling is expressed directly by the total gradient
\begin{equation}
\label{eq:orcu_gradient}
\frac{\partial\mathcal L_{\text{ORCU}}}{\partial z_{n,k}}
=(\hat y_{n,k}-y'_{n,k})
+\frac{\partial\mathcal L_{\text{LBE}}}{\partial z_{n,k}},
\end{equation}
where the SoftCE term pulls the prediction toward the distance-aware target, while the LBE term changes relative logits according to the current adjacent gaps. These gaps have an exact structural interpretation. For $k<y_n$, $e^{r_{n,k}}=\hat y_{n,k}/\hat y_{n,k+1}$; for $k\ge y_n$, $e^{r_{n,k}}=\hat y_{n,k+1}/\hat y_{n,k}$. Thus, $e^{r_{n,k}}$ is the farther-to-closer adjacent probability ratio with respect to the target. A positive gap means that the farther class receives more probability than its closer neighbor, and is therefore exactly an adjacent violation of the required unimodal ordering around $y_n$. As $r$ increases, the local relation moves from a well-separated ordered configuration, through near-boundary ambiguity, to explicit violation; in this ordinal sense, $r$ acts as a local, input-specific uncertainty proxy for the LBE response.

\input{tables/tb1_gradient}

For readability, Table~\ref{table:gradient} reports the contribution of a single adjacent pair. For $1<k<C$, the complete derivative sums the two neighboring contributions (Appx.~V). Under a direct logit-space step induced by one LBE pair, $r_{n,k}^{\mathrm{new}}=r_{n,k}-2\eta\hat I'(r_{n,k})<r_{n,k}$ because $\hat I'(r)>0$. Each pairwise update therefore reduces a violation or increases the ordering margin of an already feasible relation. The update magnitude is small but nonzero deep in the interior, rises continuously near the boundary, and remains bounded after the transition.

Summed across neighboring pairs, these updates change distributional width in a structured way. At the target index, the available LBE contributions increase $z_{n,y_n}$ under gradient descent, counteracting the fixed-width tendency of the soft target. Away from the target, each contribution raises the logit closer to $y_n$ relative to its farther neighbor; the opposite side behaves symmetrically. Large negative gaps attenuate the change for well-separated ordered samples, whereas near-boundary or violating gaps produce stronger redistribution toward the target and its ordinal neighborhood. Softmax normalization converts these sample-dependent relative-logit updates into an adaptive output width, while $\mathcal L_{\text{SoftCE}}$ keeps the prediction anchored to the distance-aware target.

\noindent\textbf{Theoretical properties.}
The theory clarifies why the two loss terms must be coupled rather than used separately. In the per-sample free-logit relaxation, the LBE objective alone is unbounded below, and the same remains true when LBE is combined with one-hot CE along ordinal rays that sharpen the logit profile around $y_n$. SoftCE prevents this unbounded descent because its full-support target grows linearly along those rays, whereas LBE decreases only logarithmically (Appx.~IV-B). After fixing the uniform-shift degree of freedom, the resulting objective is strictly convex and coercive on the shift-free logit subspace. Hence ORCU has a unique finite per-sample minimizer for every $s>0$, and this minimizer is strictly unimodal with mode $y_n$ for every $s>0$ (Appx.~IV-C). Thus, the barrier scale is not needed to obtain ordinal feasibility; it controls how strongly the finite SoftCE anchor is refined.

The optimality condition also identifies the form of this refinement. Let $p_s^\star=\operatorname{softmax}(z_s^\star)$ denote the free-logit ORCU equilibrium, let $a=\mathbf y'_n$, and set $q_{s,k}=\hat I'(r_{s,k}^\star)$ with $0<q_{s,k}\le s$. In cumulative form, stationarity gives, for thresholds $k=1,\ldots,C-1$,
\begin{equation}
\label{eq:cdf_refinement_main}
F_{p_s^\star}(k)=
\begin{cases}
F_a(k)-q_{s,k}, & k<y_n,\\
F_a(k)+q_{s,k}, & k\ge y_n,
\end{cases}
\end{equation}
where $F_p(k)=\sum_{j=1}^k p_j$ denotes the cumulative mass up to threshold $k$. Equation~\eqref{eq:cdf_refinement_main} shows that every cumulative ordinal threshold moves from the soft target toward the observed label. The equilibrium increases probability mass on intervals containing the observed label and strictly improves the realized ranked probability score relative to the SoftCE anchor, with the deviation controlled by the barrier scale (Appx.~IV-D). These results characterize the per-sample geometry induced by ORCU; the following diagnostic then isolates its confidence-refinement effect after the unimodal ordering has formed.

\subsection{Confidence Refinement after Unimodal Ordering}
\label{sec:cal_alignment}
On the fixed Stage~1 target-feasible set $\mathcal F$ ($\sim$68\% of validation), the defining inequalities imply $z_{n,1}<\cdots<z_{n,y_n}>\cdots>z_{n,C}$, making $y_n$ the unique logit maximum at the selection checkpoint. Thus, each selected prediction is already correct and satisfies the required unimodal ordering before continuation. Freezing the same index set across branches isolates confidence refinement from changes in set membership or newly formed unimodal ordering.

\begin{table}[!t]
\caption{{Fixed target-feasible continuation on Image Aesthetics (IA; 5-fold mean). Act. denotes nonzero-gradient logit coverage; $A_{\mathrm{TF}}$ denotes average target-feasible calibration alignment; all branches retain 100\% unimodality on $\mathcal F$.}}
\label{table:cal_alignment}
\centering
\footnotesize
\setlength{\tabcolsep}{3.2pt}
\renewcommand{\arraystretch}{1.08}
\begin{tabular}{lcccc}
\toprule
Branch & Act. (\% $\uparrow$) & $A_{\mathrm{TF}}$ ($\uparrow$) & SCE ($10^{-2}$ $\downarrow$) & ECE (\% $\downarrow$) \\
\midrule
SORD  & --   & --     & 16.87 & 41.62 \\
Hinge & 3.99 & 0.0626 & 18.22 & 45.74 \\
ORCU  & \textbf{100.0} & \textbf{0.0866} & \textbf{7.25} & \textbf{16.71} \\
\bottomrule
\end{tabular}
\end{table}

At each evaluated state in a continuation branch, let $\hat k_n=\arg\max_j\hat y_{n,j}$, $c_n=\hat y_{n,\hat k_n}$, and $b_n=\mathbf 1[\hat k_n=y_n]$. For $\mathcal L_{\mathrm{reg}}\in\{\mathcal L_{\mathrm{Hinge}},\mathcal L_{\mathrm{LBE}}\}$, let $\mathbf g_n=-\nabla_{\mathbf z_n}\mathcal L_{\mathrm{reg}}$. Away from top-logit ties, a sufficiently small step that preserves $\hat k_n$ gives
\begin{align}
 a_n
 &= (b_n-c_n)\left\langle\nabla_{\mathbf z_n}c_n,\mathbf g_n\right\rangle \notag\\
 &= -\left.\frac{\partial}{\partial\eta}\frac{1}{2}
 \Bigl[b_n-c_n\bigl(\mathbf z_n+\eta\mathbf g_n\bigr)\Bigr]^2\right|_{\eta=0},
 \label{eq:cal_alignment}\\[-2pt]
 A_{\mathrm{TF}}
 &= \frac{1}{|\mathcal F|}\sum_{n\in\mathcal F}a_n.
 \label{eq:cal_alignment_average}
\end{align}
Equation~\eqref{eq:cal_alignment} is the negative directional derivative of the scaled squared correctness-confidence gap $\tfrac12(b_n-c_n)^2$, equivalently the scaled top-label Brier loss. Hence $a_n>0$ indicates that the regularizer direction locally reduces this squared gap, and $A_{\mathrm{TF}}$ is the corresponding average target-feasible calibration alignment over $\mathcal F$. We report active-logit coverage and SCE/ECE on the same fixed set.

Table~\ref{table:cal_alignment} shows that the hinge branch is active on only 3.99\% of logits and worsens SCE/ECE relative to SORD. ORCU, by contrast, retains 100\% activity, attains the largest $A_{\mathrm{TF}}$, and yields the lowest errors under the same 100\% unimodality condition. These results isolate the role of LBE after the unimodal ordering has formed: SoftCE supplies the ordinal probability target, but its fixed width limits input-specific confidence variation; LBE mitigates this underconfidence by keeping confidence adaptable within the same ordered geometry. Fig.~\ref{fig:2stage}~(c) and Section~V show the aggregate gains.

\section{Experimental Setup}
\label{sec_setup}
\subsection{Datasets and Implementation Details}
\label{subsec:dataset_and_implementation}
We evaluated ORCU on four public datasets (Appx.~II-A) with ordinal labels across aesthetic assessment, age estimation, and medical diagnosis: Image Aesthetics (IA; 5 levels)~\cite{schifanella:beauty:icwsm15}, Adience (8 age groups)~\cite{eidinger2014age}, LIMUC (4-level Mayo scores)~\cite{polat2022labeled}, and a balanced subset of Diabetic Retinopathy (DR; 5 severity levels).
We used ImageNet~\cite{deng2009imagenet}-pretrained ResNet-50~\cite{he2015deepresiduallearningimage} and additionally evaluated ResNet-34 and ResNet-101 for robustness. For soft encoding, we use squared distance metric and set the barrier scale to $s=3.0$ for all vision tasks and 0.05 for text to accommodate the differing logit scales of ResNet and BERT backbones. Sensitivity analysis (see Appx.~II-B) confirms that performance remains stable across datasets.

\vspace{-0.2cm}

\subsection{Performance Metrics}
Evaluating calibration remains essential in ordinal classification tasks, where models must provide accurate predictions alongside reliable confidence estimates. Among available metrics, Static Calibration Error (SCE) and Adaptive Calibration Error (ACE) are particularly suitable for such tasks, as they handle class imbalance and provide class-wise analysis. SCE extends Expected Calibration Error (ECE) by computing per-class calibration errors, offering finer granularity in multi-class settings~\cite{Nixon_2019_CVPR_Workshops}. ACE further improves reliability by using adaptive binning to distribute predictions evenly, addressing skewed distributions and sparsity~\cite{Nixon_2019_CVPR_Workshops}. While ECE remains the most widely adopted metric for general calibration evaluation, it measures overall calibration across fixed bins and lacks the class-wise resolution needed for ordinal tasks. We report all three metrics\textemdash ECE, SCE, and ACE\textemdash for comprehensive evaluation.

To assess whether the model captures ordinal structure, we report \%Unimodality (\%Unimodal)~\cite{cardoso2023unimodal}, which measures how often predicted distributions are unimodal around their predicted modal class. For classification performance, we include accuracy (Acc), Mean Absolute Error (MAE), and Quadratic Weighted Kappa (QWK)~\cite{ben2008comparison}. QWK reflects ordinal structure by penalizing misclassifications in proportion to their distance from the target. See Appx.~III for definitions and details.

\vspace{-0.1cm}

\input{tables/tb2_main}

\section{Results and Analyses}
\label{sec_results}
The proposed loss, ORCU, sets a new benchmark for \emph{ordinal calibration} by coupling calibration with ordinal relationships. We evaluate against 10 losses (Tables~\ref{table:main_results} and \ref{table:classification_results}): ordinal-focused SORD~\cite{diaz2019soft} (i.e., $\mathcal{L}_{\text{SoftCE}}$ only), CDW-CE~\cite{polat2022class}, CO2~\cite{albuquerque2021ordinal}, and POE~\cite{li2021learning}; calibration-oriented LS~\cite{szegedy2016rethinking}, FLSD~\cite{mukhoti2020calibrating}, MbLS~\cite{liu2022devil}, MDCA~\cite{hebbalaguppe2022stitch}, and ACLS~\cite{park2023acls}; and the baseline CE. ORCU mitigates miscalibration (Fig.~\ref{figure:reliability_diagram}) while yielding coherent embeddings (Fig.~\ref{figure:tsne}) and input-adaptive output distributions (Fig.~\ref{figure:comparison}). An ablation study confirms the roles of regularization and distance in $\mathcal{L}_{\text{ORCU}}$ (Table~\ref{table:ablation}). ORCU achieves near-perfect calibration without post-hoc correction (Table~\ref{table:posthoc}) and generalizes to non-visual domains such as text classification (Table~\ref{table:text_classification}). Overall, ORCU consistently achieves \emph{ordinal calibration} through its coupled loss design.

\subsection{Joint Calibration and Ordinal Optimization}
\label{sec:4.3.1}
\noindent\textbf{Overall Performance Evaluation:} ORCU effectively balances calibration and ordinal relationship learning, consistently
outperforming methods that focus on only one of these objectives. As shown in Table~\ref{table:main_results}, ORCU significantly reduces the calibration errors in all datasets while effectively capturing ordinal relationships between labels. Compared to SORD, ORCU improves ordinal alignment and markedly enhances SCE and ACE, reflecting the effectiveness of $\mathcal{L}_{\text{LBE}}$ via informative feasible-interior learning signals. While ECE remains widely used, its reliance on binning~\cite{roelofs2022mitigating} and binary formulation~\cite{silva2023classifier} limits its effectiveness for multi-class and ordinal settings. In contrast, SCE and ACE address these limitations by providing detailed class-specific calibration assessments and accommodating imbalanced datasets, making them more suitable for ordinal tasks. Although LS performs well on ECE, it struggles with SCE and ACE, reflecting its inability to meet the calibration requirements of ordinal tasks. ORCU, by contrast, achieves consistently strong results across all metrics, with particularly robust performance on SCE and ACE. This supports the efficacy of its ordinal-aware regularization in jointly ensuring calibration and ordinal coherence.

Although existing loss functions for ordinal tasks capture label relationships to some extent, they often fail to outperform baseline methods (e.g., CE) and prove insufficient in providing reliable confidence estimates. Similarly, calibration-focused approaches struggle to reduce calibration errors effectively in ordinal tasks and, in some cases, underperform compared to CE. These challenges emphasize the importance of a unified approach that combines ordinal relationship learning with calibration. ORCU addresses this need through its ordinal-aware regularization, which enforces unimodality and improves confidence alignment, resolving the dual challenges of ordinal classification. Performance evaluations on ResNet-34 and ResNet-101, presented in Appx.~VI-A, further demonstrate its generalizability.

\input{tables/tb3_classification}

Calibration often trades off with accuracy~\cite{minderer2021revisiting}, yet ORCU delivers markedly improved calibration while preserving competitive classification performance. On LIMUC (Table~\ref{table:classification_results}), ORCU achieves the highest Acc and QWK and the lowest MAE among all baselines, confirming that the calibration gains in Table~\ref{table:main_results} incur no degradation in classification performance. Additional datasets and Grad-CAM~\cite{selvaraju2017grad} visualizations are reported in Appx.~VI-B and Appx.~VI-C4, respectively.

\input{figures/tex/fig3_reliability_diagram}
\input{tables/tb4_posthoc}

\noindent\textbf{Calibration Reliability Analysis:} 
The proposed ORCU attains reliable calibration in ordinal settings by integrating ordinal awareness into its learning objective\textemdash a property absent from conventional formulations. Fig.~\ref{figure:reliability_diagram} presents a comparative visualization of calibration performance across a comprehensive set of benchmark losses, encompassing both ordinal- and calibration-driven approaches. CE and LS (Fig.~\ref{figure:reliability_diagram}~(a,f)) exhibit pronounced overconfidence, with predictions consistently falling below the calibration line. In contrast, SORD (Fig.~\ref{figure:reliability_diagram}~(b)) employs a soft-encoded representation to capture ordinal dependencies, yet produces underconfident predictions with a restricted confidence range. This behavior is consistent with the rigidity of distance-based soft targets, which can limit per-input confidence modulation once the unimodal geometry is formed. The calibration-focused baselines (Fig.~\ref{figure:reliability_diagram}~(g--j)) further display persistent overconfidence even under explicit confidence regularization, demonstrating that neglecting ordinal dependencies inherently limits their reliability. The proposed ORCU (Fig.~\ref{figure:reliability_diagram}~(k)) effectively addresses these deficiencies through its ordinal-aware regularization, yielding well-calibrated and trustworthy confidence estimates. Overall, these results indicate that when labels are intrinsically ordered, reliable calibration requires embedding ordinal structure in the learning objective, an inductive bias that enables $\mathcal{L}_{\text{ORCU}}$ to achieve ordinal-aware calibration. Full results are available in Appx.~VI-C1.

\noindent\textbf{Ordinal Structure and Consistency Assessment:}
ORCU (Fig.~\ref{figure:tsne}~(d)) produces well-ordered embeddings aligned with label order, reflecting its ordinal-aware regularization. Fig.~\ref{figure:tsne} presents t-SNE~\cite{zhang2023improving} embeddings on LIMUC, illustrating how different loss functions capture ordinal relationships. We include CE, SORD, and LS for their competitive performance (Table~\ref{table:main_results}). CE and LS (Fig.~\ref{figure:tsne}~(a, c)) yield dispersed embeddings with limited ordinal structure. SORD (Fig.~\ref{figure:tsne}~(b)) retains partial ordinal information but lacks structural coherence. These results emphasize ORCU's advantage in capturing ordinal relationships.

\input{figures/tex/fig4_5}

\noindent\textbf{Adaptive Confidence Shaping:}
Critically, ORCU generates well-calibrated, input-adaptive confidence shaping via its ordinal-aware regularization, aligning confidence with structural order. Fig.~\ref{figure:comparison} compares softmax outputs of ORCU and SORD on the Adience dataset for correct and incorrect predictions. In correct cases (Fig.~\ref{figure:comparison}~(a)), SORD yields near-fixed distributions across the upper- and lower-row samples, assigning comparable confidence despite differing uncertainty, indicating limited input-specific confidence modulation. In incorrect cases (Fig.~\ref{figure:comparison}~(b)), it produces symmetric distributions centered on the predicted class, allocating similar confidence to equidistant labels and neglecting ordinal asymmetry, where proximity to the ground truth should imply progressively higher confidence. 
This behavior stems from its fixed, metric-driven formulation, which enforces alignment to a predetermined distance-based structure and constrains adaptive confidence shaping. Unlike SORD's rigid outputs, ORCU modulates distribution width through the sample-specific adjacent-gap state \(r\), a local ordinal-structure proxy that tracks whether adjacent relations around the target are well separated, near the ordinal boundary, or violated (Appx.~IV-E). Even when the top prediction is incorrect, ORCU shifts probability mass toward the target side, supporting the regularizer's complementary role. These observations indicate that the gain reflects ordinal structural refinement rather than simple score smoothing.

\vspace{-0.2cm}

\subsection{Further Evaluation}
\noindent\textbf{Post-hoc Calibration:}
ORCU eliminates the need for a separate post-hoc calibration stage by enforcing calibration at training time through a loss-level, ordinal-aware objective. Table~\ref{table:posthoc} shows that applying Temperature Scaling (TS) or Dirichlet Calibration (DC) to ORCU does not improve ECE. While TS yields marginal gains for several baselines, their ECE remains worse than ORCU without post-hoc adjustment. Post-hoc calibration also requires a held-out validation split and dataset-specific fitting, adding computational and operational overhead. In contrast, ORCU uses a single fixed barrier scale hyperparameter chosen once and shared across datasets, avoiding auxiliary calibration and additional tuning while ensuring well-calibrated predictions.

\input{tables/tb5_text}

\noindent\textbf{Domain Generalization:}
ORCU demonstrated promising generalizability beyond images, yielding calibrated, ordinal-aware predictions in a non-visual domain (Table~\ref{table:text_classification}). We validate this via a preliminary evaluation on SST-5~\cite{socher2013recursive}, a text dataset with linguistic and semantic uncertainty~\cite{liu2025uncertainty}. ORCU improved calibration while preserving ordinal consistency, suggesting cross-modal applicability.

\vspace{-0.2cm}
\input{tables/tb6_ablation}
\subsection{Ablation Study}
\label{sec:4.3.2}
Ablations (Table~\ref{table:ablation}) show that ORCU's complementary components are jointly necessary for unimodality and calibration. Substituting our regularizer with margin-activated hinge penalties, such as the ReLU-based MbLS and ACLS, yields limited calibration benefit: these penalties are inactive within the margin, provide no gradients in the non-violation region, and thus preclude effective confidence control once the unimodality inequalities are satisfied. This saturation behavior is consistent with our feasible-interior analysis and is reflected in the two-stage continuation diagnostic (Fig.~\ref{fig:2stage}). Lacking order awareness, they can disrupt the ordinal structure induced by $\mathcal{L}_{\text{SoftCE}}$, often degrading calibration. Smooth alternatives such as SoftPlus also fail to deliver effective feasible-interior refinement, since their gradients decay rapidly in the deep interior and empirically worsen calibration (as discussed in Section~\ref{sec:gradAnal}). In contrast, ORCU's ordinal log-barrier integrates informative feasible-region signals with stable corrections under violations, enabling joint optimization of unimodality and calibration. Among distance metrics, the squared metric performs best; both the regularizer and the choice of distance measure are essential for ORCU's intended behavior, with any alteration or removal of either consistently degrading performance. 

\vspace{-0.2cm}

\section{Discussion and Conclusion}
While this study focuses on image-based tasks, preliminary results on a text dataset suggest that ORCU may generalize to other modalities (e.g., text, audio, video) where ordinal inference and calibration are critical. Future work will explore its broader applicability and address residual miscalibration at extreme confidence levels, as shown in reliability diagrams.

This paper introduced the concept of \emph{ordinal calibration} for ordinal classification and proposed ORCU, a loss function that unifies soft ordinal encoding and ordinal-aware regularization within a single, principled framework. By design, ORCU addresses the dual challenges of miscalibration and structural inconsistency in ordinal classification without requiring architectural or post-hoc overhead. We established that the coupling of these components is theoretically robust, as the soft encoded distribution pre-conditions the logit space by providing a finite, full-support ordinal anchor for well-posed refinement, while the log-barrier extension preserves non-vanishing adjacent-gap gradients through the feasible interior. Our extensive experiments on four diverse benchmarks, in a reproducible comparison against 10 losses, demonstrate state-of-the-art calibration without compromising accuracy. This work provides a rigorous foundation for future research in trustworthy ordinal prediction, offering a path toward models that are not only highly accurate but also inherently consistent with the ordered nature of the real-world phenomena.

%% file: figures/tex/fig1_intro.tex
\begin{figure*}[!t]
\centering
\includegraphics[width=0.96\textwidth]{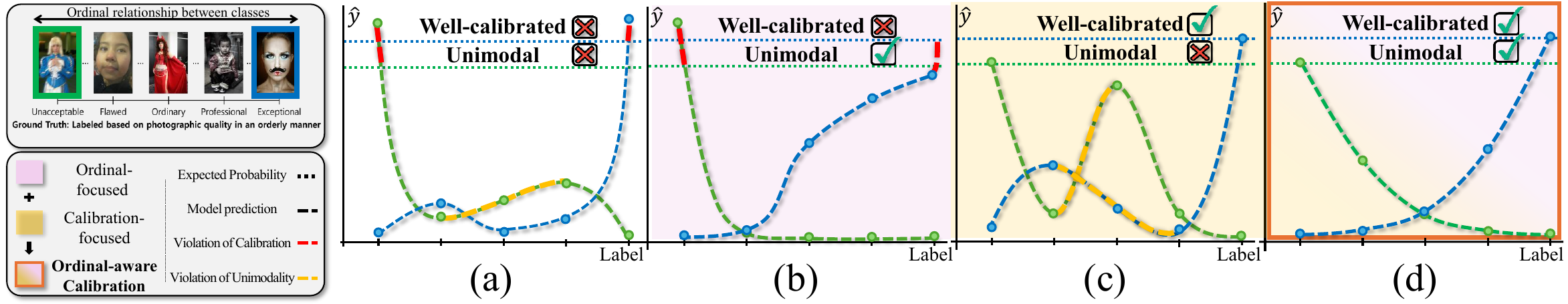}
\vspace{-0.3cm}
\caption{Comparison of predicted probability distributions for a 5-class ordinal classification task. Discrete probabilities are shown at class indices; interpolated curves are visual guides only. Green and blue denote ground-truth labels 1 and 5, respectively. (a) Baseline: neither unimodal nor calibrated. (b) Ordinal-focused loss: unimodal only. (c) Calibration-focused loss: calibrated only. (d) \( \mathcal{L}_{\text{ORCU}} \): both calibrated and unimodal.}
\vspace{-0.2cm}
\label{figure:intro}
\end{figure*}

%% file: tables/tb1_gradient.tex
\begin{table}[t!]
\renewcommand{\arraystretch}{0.9}
\caption{Gradient behavior of the ORCU loss with respect to logits, for $k < y_n$ and $k \geq y_n$. $\hat{y}_{n,k}$ and $y'_{n,k}$ denote the $k$-th components of the predicted probability and soft target, respectively; $s$ denotes the barrier scale of the regularizer. Note that $ \frac{\partial \mathcal{L}_{\text{ORCU}}}{\partial z_{n,k}} = \frac{\partial \mathcal{L}_{\text{SoftCE}}}{\partial z_{n,k}} + \frac{\partial \mathcal{L}_{\text{LBE}}}{\partial z_{n,k}} $. For clarity, this table presents the gradient behavior for a single adjacent logit pair; see Appx.~V for full derivations.}
\vspace{-0.2cm}
\centering
\resizebox{\columnwidth}{!}{%
        {
        \begin{tabular}{ccccc}
            \toprule
            & \multicolumn{2}{c}{\begin{tabular}[c]{@{}c@{}}$k < y_n$ ($ r = z_{n,k} - z_{n,k+1}$)\end{tabular}} 
            & \multicolumn{2}{c}{\begin{tabular}[c]{@{}c@{}}$k \geq y_n$ ($ r = z_{n,k+1} - z_{n,k}$)\end{tabular}} 
            \\ 
            \cmidrule(rl){2-3} \cmidrule(rl){4-5} 
            & \multicolumn{1}{c}{$ r \leq -\frac{1}{s^2} $} 
            & \multicolumn{1}{c}{$ r > -\frac{1}{s^2} $} 
            & \multicolumn{1}{c}{$ r \leq -\frac{1}{s^2} $} 
            & $ r > -\frac{1}{s^2} $ 
            \\ 
            \midrule
            \makebox[2em][c]{(a)} \makebox[4em]{$\frac{\partial \mathcal{L_\text{SoftCE}}}{\partial z_{n,k}}$}
            & \multicolumn{4}{c}{$\hat{y}_{n,k} - y'_{n,k}$} 
            \\ \midrule
            \makebox[2em][c]{(b)} \makebox[4em]{$\frac{\partial \mathcal{L_\text{LBE}}}{\partial z_{n,k}}$}
            & \multicolumn{1}{c}{$-\frac{1}{s\cdot r}$} 
            & \multicolumn{1}{c}{$s$} 
            & \multicolumn{1}{c}{$\frac{1}{s\cdot r}$} 
            & \multicolumn{1}{c}{$-s$} 
            \\ \midrule
            \makebox[2em][c]{(c)} \makebox[4em]{$\frac{\partial \mathcal{L_{\text{ORCU}}}}{\partial z_{n,k}}$}
            & \multicolumn{1}{c}{$(\hat{y}_{n,k}-y'_{n,k})-\frac{1}{s\cdot r}$} 
            & \multicolumn{1}{c}{$(\hat{y}_{n,k}-y'_{n,k})+s$} 
            & \multicolumn{1}{c}{$(\hat{y}_{n,k}-y'_{n,k})+\frac{1}{s\cdot r}$} 
            & \multicolumn{1}{c}{$(\hat{y}_{n,k}-y'_{n,k})-s$} 
            \\ \bottomrule
        \end{tabular}
        }
    }
    \vspace{-0.4cm}
    \label{table:gradient}
\end{table}

%% file: tables/tb2_main.tex
\begin{table*}[!t]
\renewcommand{\arraystretch}{1.0}
\caption{Calibration performance (SCE, ACE, and ECE) and ordinal structure representation (Unimodal) across various loss functions on four ordinal datasets. The table reports mean$_{\text{std}}$ across folds, highlighting the best in bold and the second-best underlined. Additional results for ResNet-34 and ResNet-101 are provided in Appx.~VI-A.}

\vspace{-0.15cm}
\centering
\resizebox{\textwidth}{!}{%
        {
        \begin{tabular}{rcccccccccccc}
        \toprule
        \multicolumn{1}{c}{\multirow{2}{*}{Metrics}} & \multicolumn{11}{c}{Methods} \\
        \cmidrule(lr){2-12}
        & CE
        & SORD 
        & CDW-CE
        & CO2
        & POE
        & LS
        & FLSD
        & MbLS
        & MDCA
        & ACLS
        & \cellcolor[gray]{.93} \textBF{ORCU (Ours)} \\
        \midrule
        \multicolumn{12}{c}{Image Aesthetics (\(n = 13{,}364\))} \\
        SCE($10^{-2}$) $\downarrow$ & \phantom{0}8.48$_{0.68}$ & \phantom{0}8.49$_{0.08}$ & \phantom{0}7.16$_{0.46}$ & \phantom{0}9.25$_{0.56}$ & \phantom{0}8.27$_{0.33}$ & \phantom{0}\underline{5.65}$_{0.34}$ & \phantom{0}7.52$_{0.46}$ & \phantom{0}7.85$_{0.15}$ & \phantom{0}8.32$_{0.49}$ & \phantom{0}8.01$_{0.42}$ & \cellcolor[gray]{.93} \textBF{\phantom{0}2.82$_{0.23}$} \\
        ACE($10^{-2}$) $\downarrow$ & \phantom{0}8.07$_{0.67}$ & \phantom{0}8.42$_{0.08}$ & \phantom{0}6.94$_{0.47}$ & \phantom{0}8.92$_{0.56}$ & \phantom{0}8.08$_{0.24}$ & \phantom{0}\underline{5.69}$_{0.28}$ & \phantom{0}7.19$_{0.41}$ & \phantom{0}7.59$_{0.13}$ & \phantom{0}7.95$_{0.46}$ & \phantom{0}7.77$_{0.36}$ & \cellcolor[gray]{.93} \textBF{\phantom{0}3.07$_{0.26}$} \\
        ECE(\%) $\downarrow$ & 20.57$_{1.62}$ & 18.46$_{0.26}$ & 17.51$_{1.36}$ & 22.69$_{1.51}$ & 20.14$_{0.93}$ & \phantom{0}\underline{9.91}$_{0.62}$ & 18.37$_{1.21}$ & 18.95$_{0.62}$ & 20.22$_{1.24}$ & 19.57$_{0.97}$ & \cellcolor[gray]{.93} \textBF{\phantom{0}2.57$_{0.40}$} \\
        Unimodal(\%) $\uparrow$ & 96.76$_{1.59}$ & \textBF{100.0$_{0.00}$} & \underline{99.71}$_{0.43}$ & 94.00$_{1.61}$ & 89.60$_{1.90}$ & 79.07$_{0.49}$ & 98.07$_{0.95}$ & 84.97$_{1.14}$ & 96.32$_{1.27}$ & 84.61$_{0.76}$ & \cellcolor[gray]{.93} \textBF{100.0$_{0.00}$} \\
        \midrule
        \multicolumn{12}{c}{Adience (\(n = 17{,}423\))} \\
        SCE($10^{-2}$) $\downarrow$ & \phantom{0}8.68$_{0.95}$ & \phantom{0}\underline{4.26}$_{0.77}$ & \phantom{0}7.76$_{0.53}$ & \phantom{0}9.12$_{1.03}$ & \phantom{0}7.05$_{0.82}$ & \phantom{0}5.64$_{0.85}$ & \phantom{0}8.25$_{1.00}$ & \phantom{0}7.36$_{0.86}$ & \phantom{0}8.72$_{1.03}$ & \phantom{0}7.48$_{0.90}$ & \cellcolor[gray]{.93} \phantom{0}\textBF{4.20$_{0.16}$} \\
        ACE($10^{-2}$) $\downarrow$ & \phantom{0}8.24$_{0.99}$ & \phantom{0}\underline{4.33}$_{0.71}$ & \phantom{0}7.43$_{0.51}$ & \phantom{0}8.58$_{0.96}$ & \phantom{0}6.75$_{0.77}$ & \phantom{0}5.56$_{0.85}$ & \phantom{0}7.80$_{0.95}$ & \phantom{0}7.03$_{0.82}$ & \phantom{0}8.28$_{1.08}$ & \phantom{0}7.13$_{0.92}$ & \cellcolor[gray]{.93} \phantom{0}\textBF{4.04$_{0.21}$} \\
        ECE(\%) $\downarrow$ & 33.64$_{4.01}$ & \phantom{0}\textBF{7.31$_{2.40}$} & 29.13$_{2.10}$ & 35.33$_{4.06}$ & 26.69$_{3.63}$ & 18.90$_{4.15}$ & 31.93$_{4.07}$ & 28.15$_{3.44}$ & 33.72$_{4.11}$ & 28.47$_{3.78}$ & \cellcolor[gray]{.93} \phantom{0}\underline{8.80}$_{2.27}$ \\
        Unimodal(\%) $\uparrow$ & 85.16$_{1.32}$ & \underline{98.76}$_{0.95}$ & 93.63$_{0.89}$ & 82.36$_{1.73}$ & 70.21$_{0.56}$ & 67.24$_{0.78}$ & 86.65$_{0.95}$ & 69.38$_{0.82}$ & 84.55$_{2.41}$ & 68.96$_{0.91}$ & \cellcolor[gray]{.93} \textBF{99.94$_{0.04}$} \\
        \midrule
        \multicolumn{12}{c}{LIMUC (\(n = 11{,}276\))} \\
        SCE($10^{-2}$) $\downarrow$ & \phantom{0}6.82$_{0.79}$ & \phantom{0}8.74$_{0.30}$ & \phantom{0}6.20$_{0.50}$ & \phantom{0}7.98$_{0.52}$ & \phantom{0}6.87$_{0.63}$ & \phantom{0}\underline{4.66}$_{0.30}$ & \phantom{0}5.60$_{0.80}$ & \phantom{0}6.72$_{0.40}$ & \phantom{0}6.24$_{0.65}$ & \phantom{0}6.70$_{0.67}$ & \cellcolor[gray]{.93} \phantom{0}\textBF{3.98$_{0.41}$} \\
        ACE($10^{-2}$) $\downarrow$ & \phantom{0}6.39$_{0.83}$ & \phantom{0}8.55$_{0.24}$ & \phantom{0}5.80$_{0.42}$ & \phantom{0}7.56$_{0.60}$ & \phantom{0}6.58$_{0.68}$ & \phantom{0}\underline{4.47}$_{0.30}$ & \phantom{0}5.21$_{0.78}$ & \phantom{0}6.35$_{0.44}$ & \phantom{0}5.88$_{0.65}$ & \phantom{0}6.32$_{0.68}$ & \cellcolor[gray]{.93} \phantom{0}\textBF{3.62$_{0.47}$} \\
        ECE(\%) $\downarrow$ & 12.95$_{1.70}$ & 16.36$_{0.64}$ & 11.90$_{0.96}$ & 15.44$_{1.18}$ & 13.41$_{1.39}$ & \phantom{0}\textBF{5.92$_{0.88}$} & 10.69$_{1.67}$ & 13.01$_{0.90}$ & 11.97$_{1.29}$ & 12.99$_{1.40}$ & \cellcolor[gray]{.93} \phantom{0}\underline{6.54}$_{0.73}$ \\
        Unimodal(\%) $\uparrow$ & 97.98$_{1.20}$ & \textBF{100.0$_{0.00}$} & \underline{99.86}$_{0.32}$ & 96.93$_{1.46}$ & 89.08$_{1.03}$ & 78.10$_{0.94}$ & 98.61$_{1.06}$ & 85.03$_{1.31}$ & 98.19$_{0.99}$ & 84.58$_{1.34}$ & \cellcolor[gray]{.93} \textBF{100.0$_{0.00}$} \\
        \midrule
        \multicolumn{12}{c}{Diabetic Retinopathy (\(n = 9{,}570\))} \\
        SCE($10^{-2}$) $\downarrow$ & \phantom{0}9.83$_{0.19}$ & \phantom{0}\underline{4.08}$_{0.17}$ & \phantom{0}9.29$_{0.52}$ & 11.94$_{0.61}$ & \phantom{0}7.53$_{0.97}$ & \phantom{0}6.23$_{0.55}$ & \phantom{0}7.56$_{0.56}$ & \phantom{0}9.03$_{0.58}$ & \phantom{0}9.48$_{0.80}$ & \phantom{0}9.71$_{0.42}$ & \cellcolor[gray]{.93} \phantom{0}\textBF{3.75$_{0.43}$} \\
        ACE($10^{-2}$) $\downarrow$ & \phantom{0}9.62$_{0.18}$ & \phantom{0}\underline{4.17}$_{0.19}$ & \phantom{0}9.09$_{0.58}$ & 11.71$_{0.61}$ & \phantom{0}7.36$_{0.94}$ & \phantom{0}6.10$_{0.48}$ & \phantom{0}7.40$_{0.50}$ & \phantom{0}8.69$_{0.61}$ & \phantom{0}9.27$_{0.77}$ & \phantom{0}9.52$_{0.44}$ & \cellcolor[gray]{.93} \phantom{0}\textBF{3.67$_{0.30}$} \\
        ECE(\%) $\downarrow$ & 23.48$_{0.57}$ & \phantom{0}\underline{4.27}$_{0.23}$ & 21.51$_{1.65}$ & 28.67$_{1.66}$ & 17.72$_{2.65}$ & 14.39$_{1.39}$ & 17.90$_{1.85}$ & 21.38$_{1.64}$ & 22.45$_{2.14}$ & 23.32$_{1.19}$ & \cellcolor[gray]{.93} \phantom{0}\textBF{3.22$_{0.80}$} \\
        Unimodal(\%) $\uparrow$ & 90.34$_{0.97}$ & 99.00$_{0.04}$ & \textBF{99.99$_{0.01}$} & 87.68$_{1.39}$ & 92.03$_{2.22}$ & 78.79$_{1.27}$ & 93.01$_{0.82}$ & 84.26$_{1.27}$ & 91.13$_{1.52}$ & 84.54$_{1.07}$ & \cellcolor[gray]{.93} \underline{99.96}$_{0.02}$ \\
        \bottomrule
        \end{tabular}
        }
    }
    \vspace{-0.15cm}
    \label{table:main_results}
\end{table*}

%% file: tables/tb3_classification.tex
\begin{table*}[!t]
\renewcommand{\arraystretch}{0.85}
\caption{Classification performance (Acc, QWK, and MAE) across various loss functions. The table reports mean$_{\text{std}}$ across folds, highlighting the best in bold and the second-best underlined. Results are presented for the LIMUC. Full results are provided in Appx.~VI-B.}
\vspace{-0.15cm}
\centering
\resizebox{\textwidth}{!}{%
        {
        \begin{tabular}{rcccccccccccc}
        \toprule
        \multicolumn{1}{c}{\multirow{2}{*}{Metrics}} & \multicolumn{11}{c}{Methods} \\
        \cmidrule(lr){2-12}
        & CE
        & SORD
        & CDW-CE
        & CO2
        & POE
        & LS
        & FLSD
        & MbLS
        & MDCA
        & ACLS
        & \cellcolor[gray]{.93} \textBF{ORCU (Ours)} \\
        \midrule
        Acc(\%) $\uparrow$
        & 77.02$_{0.66}$
        & 77.49$_{0.60}$
        & \underline{77.73}$_{0.58}$
        & 76.62$_{0.76}$
        & 76.60$_{0.74}$
        & 76.33$_{0.55}$
        & 77.40$_{0.55}$
        & 76.55$_{0.79}$
        & 76.92$_{0.48}$
        & 76.83$_{0.89}$
        & \cellcolor[gray]{.93} \textBF{78.38}$_{0.73}$ \\
        QWK($10^{-2}$) $\uparrow$
        & 84.61$_{0.90}$
        & 85.39$_{0.62}$
        & \underline{85.51}$_{0.69}$
        & 84.11$_{0.81}$
        & 83.92$_{1.11}$
        & 84.02$_{0.86}$
        & 84.93$_{0.77}$
        & 84.09$_{0.81}$
        & 84.13$_{0.60}$
        & 84.54$_{0.92}$
        & \cellcolor[gray]{.93} \textBF{86.21}$_{0.67}$ \\
        MAE($10^{-2}$) $\downarrow$
        & 23.87$_{0.83}$
        & 23.14$_{0.63}$
        & \underline{22.88}$_{0.67}$
        & 24.32$_{0.86}$
        & 24.49$_{1.00}$
        & 24.67$_{0.67}$
        & 23.47$_{0.72}$
        & 24.49$_{0.87}$
        & 24.15$_{0.49}$
        & 24.07$_{0.99}$
        & \cellcolor[gray]{.93} \textBF{22.10}$_{0.82}$ \\
        \bottomrule
        \end{tabular}
        }
    }
\label{table:classification_results}
\end{table*}

%% file: figures/tex/fig3_reliability_diagram.tex
\begin{figure*}[!t]
\begin{center}
\includegraphics[width=0.9\textwidth]{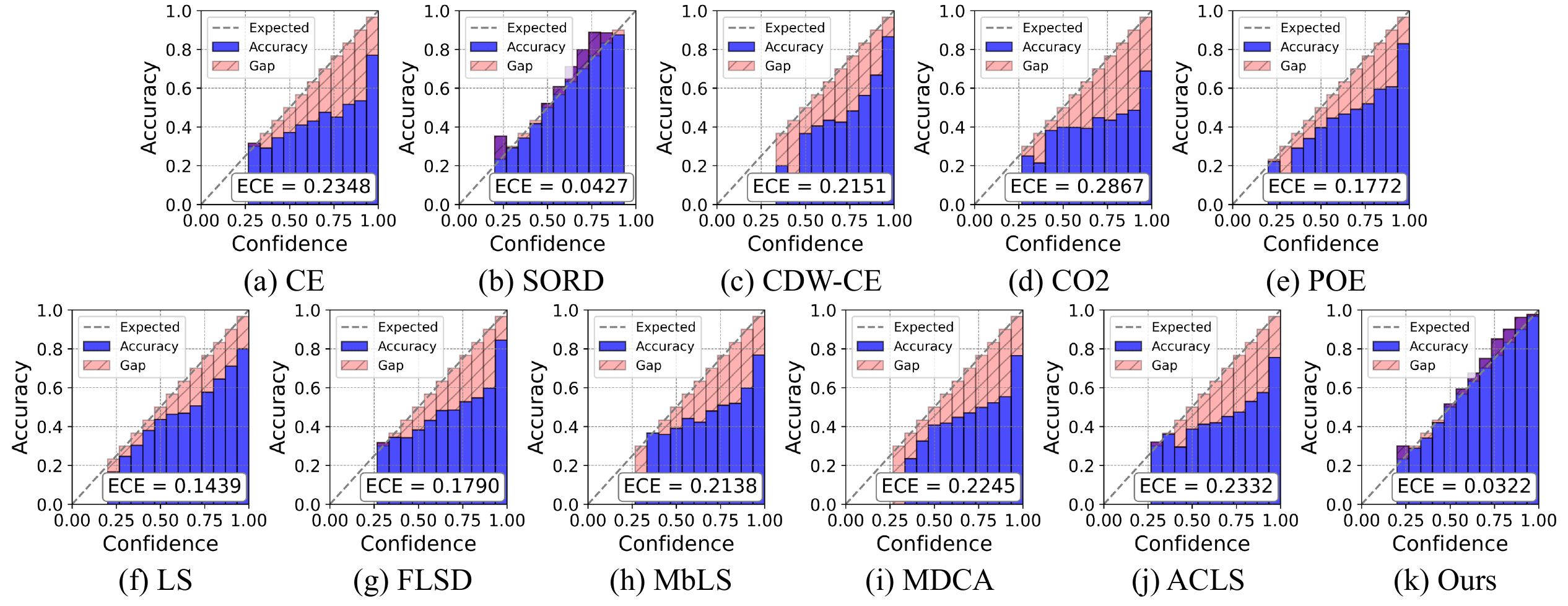}
\end{center}
\vspace{-0.4cm}

\caption{Reliability diagrams for ordinal loss functions, illustrating model confidence and the calibration gap between confidence and accuracy on the DR test split. Accuracy bars extending above the expected line indicate underconfidence (\( P(Y=y \mid \hat{P}=p)>p \)), while those below it indicate overconfidence (\(P(Y=y \mid \hat{P}=p)<p\)). ECE is computed using 15 bins. Extended analyses are provided in Appx.~VI-C1.}
\vspace{-0.3cm}

\label{figure:reliability_diagram}
\end{figure*}

%% file: tables/tb4_posthoc.tex
\begin{table*}[!t]
\renewcommand{\arraystretch}{0.65}
\caption{ECE after Post-hoc (PH) calibration with Temperature Scaling (TS)~\cite{guo2017calibration} and Dirichlet Calibration (DC)~\cite{kull2019beyond} on IA, without POE for incompatibility. Results are reported as mean$_{\text{std}}$. Best in bold; experimental details are provided in Appx.~II-B.}
\vspace{-0.3cm}
\centering
\resizebox{\textwidth}{!}{%
        {
        \begin{tabular}{rccccccccccc}
        \toprule
        \multicolumn{1}{c}{\multirow{2}{*}{ECE(\%)}} & \multicolumn{10}{c}{Methods} \\
        \cmidrule(lr){2-11}
        & CE & SORD & CDW-CE & CO2 & LS & FLSD & MbLS & MDCA & ACLS & \cellcolor[gray]{.93} \textBF{ORCU (Ours)} \\
        \midrule
        w/o PH 
        & 20.57$_{1.62}$
        & 18.46$_{0.26}$
        & 17.51$_{1.36}$
        & 22.69$_{1.51}$
        & \phantom{0}9.91$_{0.62}$
        & 18.37$_{1.21}$
        & 18.95$_{0.62}$
        & 20.22$_{1.24}$
        & 19.57$_{0.97}$
        & \cellcolor[gray]{.93} \phantom{0}\textBF{2.57}$_{0.40}$ \\
        w/ TS
        & \phantom{0}3.21$_{0.46}$
        & \phantom{0}3.32$_{1.18}$
        & \phantom{0}4.09$_{0.64}$
        & \phantom{0}3.84$_{0.99}$
        & \phantom{0}4.19$_{1.22}$
        & \phantom{0}2.70$_{0.85}$
        & \phantom{0}4.15$_{1.14}$
        & \phantom{0}3.49$_{1.12}$
        & \phantom{0}4.46$_{1.59}$
        & \cellcolor[gray]{.93} \phantom{0}5.53$_{2.99}$ \\
        w/ DC
        & 18.22$_{1.85}$
        & 26.12$_{3.42}$
        & 30.95$_{1.51}$
        & 13.86$_{1.92}$
        & 17.54$_{0.74}$
        & 17.73$_{2.55}$
        & 17.84$_{2.55}$
        & 17.96$_{3.31}$
        & 18.60$_{1.73}$
        & \cellcolor[gray]{.93} 34.13$_{2.23}$ \\
        \bottomrule
        \end{tabular}
        }
    }
    \vspace{-0.2cm}
    \label{table:posthoc}
\end{table*}

%% file: figures/tex/fig4_5.tex
\begin{figure}[!t]
\centering
\includegraphics[width=0.8\columnwidth]{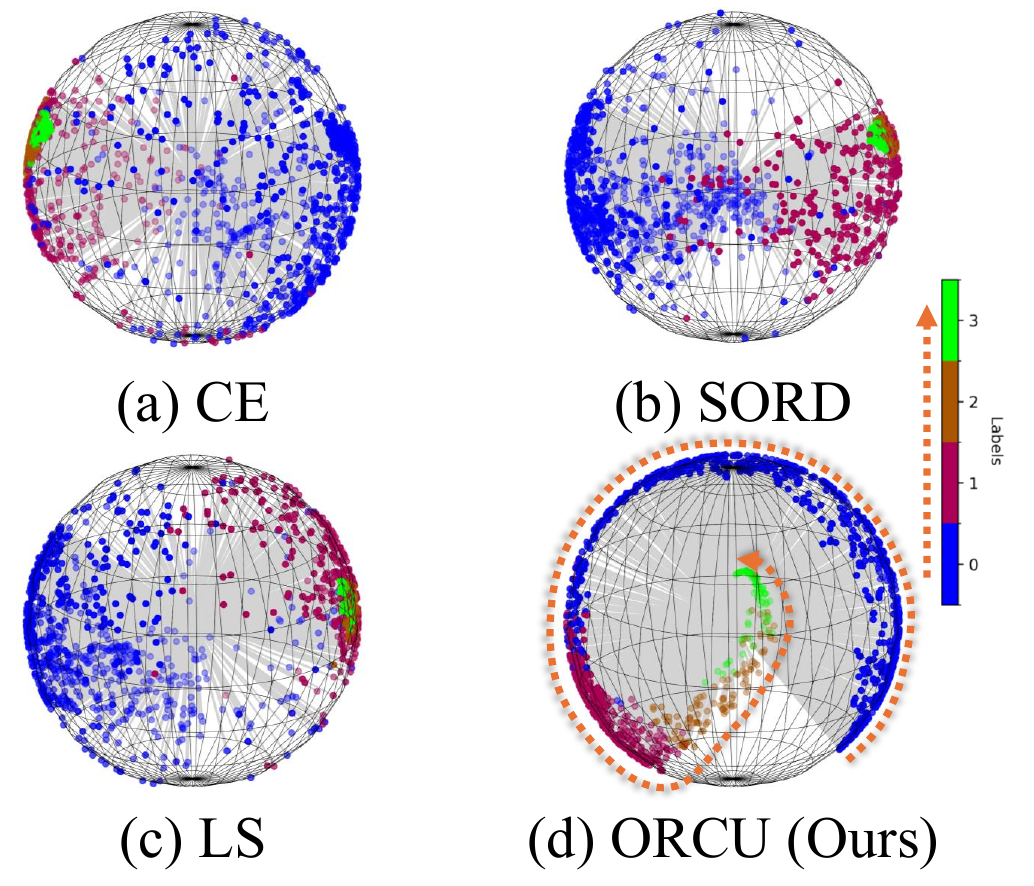}
\caption{t-SNE of ResNet-50 feature embeddings on  LIMUC for CE, SORD, LS, and ORCU. Colors denote class labels, and the orange dashed line represents feature alignment, illustrating effective ordinal learning. See Appx.~VI-C2 for other datasets.}
\label{figure:tsne}
\end{figure}

\begin{figure}[!t]
\centering
\includegraphics[width=0.8\columnwidth]{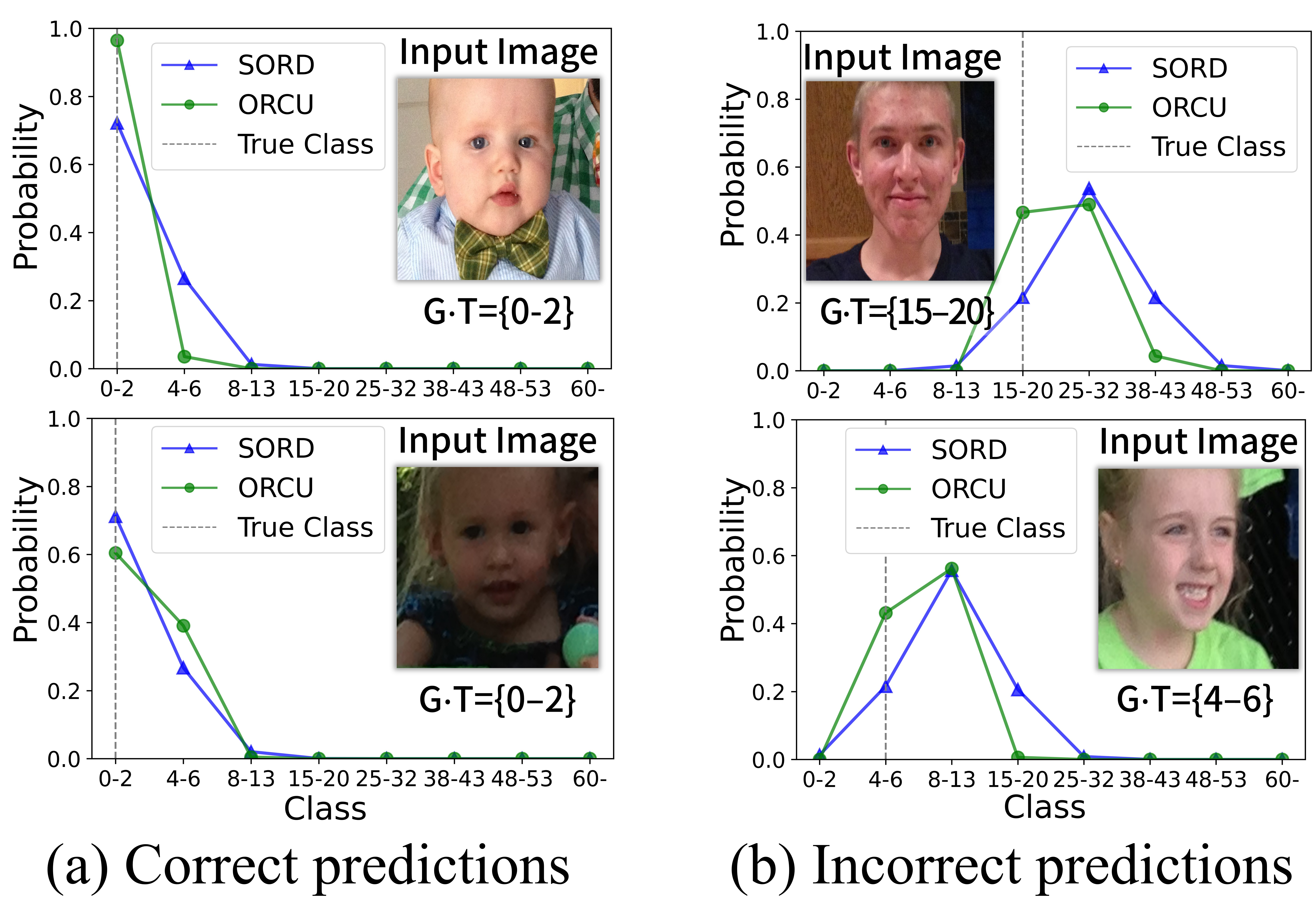}
\caption{Test-split output distributions on Adience for (a) correct and (b) incorrect predictions. Each subfigure compares \textcolor{Green}{ORCU} with \textcolor{blue}{SORD}; the gray dashed line marks the ground-truth age group. Comparisons with CE and LS: Appx.~VI-C3.}
\label{figure:comparison}
\end{figure}

%% file: tables/tb5_text.tex
\begin{table}[!t]
\renewcommand{\arraystretch}{0.7}
\caption{Text classification on the SST-5 dataset for domain generalization. Best and second-best results are in bold and underlined. Experimental details: see Appx.~II-B.}
\vspace{-0.2cm}
\centering
\resizebox{\columnwidth}{!}{%
        {
        \Huge
        \begin{tabular}{rccccccccccc}
        \toprule
        \multicolumn{1}{c}{\multirow{2}{*}{Metrics}} & \multicolumn{10}{c}{Methods} \\
        \cmidrule(lr){2-11}
        & CE
        & SORD
        & CDW-CE
        & CO2
        & LS
        & FLSD
        & MbLS
        & MDCA
        & ACLS
        & \cellcolor[gray]{.93} \textBF{ORCU (Ours)} \\
        \midrule
        ECE(\%) $\downarrow$
        & \phantom{0}3.17 & \phantom{0}7.07 & 26.39 & 21.49 
        & \phantom{0}\underline{3.07} & \phantom{0}6.61 & \phantom{0}5.84 & \phantom{0}4.87 & \phantom{0}3.94 
        & \cellcolor[gray]{.93} \phantom{0}\textBF{2.96} \\
        Unimodal(\%) $\uparrow$
        & 98.22 & \textBF{100.0} & 98.40 & 96.84
        & 98.86 & \underline{99.56} & 99.55 & 99.49 & 99.45 
        & \cellcolor[gray]{.93} \textBF{100.0} \\
        MAE($10^{-2}$) $\downarrow$
        & 76.92 & \underline{73.30} & 76.11 & 76.70
        & 76.52 & 75.20 & 75.61 & 80.00 & 75.07 
        & \cellcolor[gray]{.93} \textBF{72.85} \\
        \bottomrule
        \end{tabular}
        }
    }
    \vspace{-0.2cm}
    \label{table:text_classification}
\end{table}

%% file: tables/tb6_ablation.tex
\begin{table}[t!]
\renewcommand{\arraystretch}{0.8}
\caption{Ablation study on the choice of regularization term and distance metrics in $\mathcal{L}_{\text{SoftCE}}$ on the DR begins from the $\mathcal{L}_{\text{SoftCE}}$-only baseline (SORD), comparing other regularizers (MbLS, MDCA, ACLS, and SoftPlus) under a fixed distance metric, followed by variations of it in ORCU. Results are reported as mean$_{\text{std}}$. Full results are in Appx.~VI-E.}
\vspace{-0.1cm}
\centering
\resizebox{\columnwidth}{!}{%
        {
        \begin{tabular}{lcccc}
            \toprule
            \multicolumn{1}{c}{Combination} & \multicolumn{4}{c}{Metrics} \\ 
            \cmidrule(lr){2-5}
            \makebox[5em][c]{$\mathcal{L}_{\text{SoftCE}{(\phi=\{\})}}$} \makebox[2em]{+} \makebox[4.8em][c]{$\mathcal{L}_{\text\{\}}$} & SCE ($10^{-2}$) $\downarrow$ & ACE ($10^{-2}$) $\downarrow$ & ECE (\%) $\downarrow$ & Unimodal(\%) $\uparrow$ \\ 
            \midrule
            \makebox[5em][c]{Squared} \makebox[2em]{+} \makebox[4.8em][c]{-} & 4.08$_{0.17}$ & 4.17$_{0.19}$ & \phantom{0}4.27$_{0.23}$ & 99.90$_{0.04}$ \\ 
            \makebox[5em][c]{} \makebox[2em]{+} \makebox[4.8em][c]{MbLS} & 3.95$_{0.24}$ & 3.99$_{0.18}$ & \phantom{0}4.65$_{0.66}$ & 99.27$_{0.18}$ \\
            \makebox[5em][c]{} \makebox[2em]{+} \makebox[4.8em][c]{MDCA} & 4.06$_{0.30}$ & 4.10$_{0.22}$ & \phantom{0}4.71$_{1.08}$ & 99.93$_{0.02}$ \\
            \makebox[5em][c]{} \makebox[2em]{+} \makebox[4.8em][c]{ACLS}  & 3.99$_{0.29}$ & 4.13$_{0.25}$ & \phantom{0}5.07$_{1.10}$ & 99.43$_{0.17}$\\  
            \makebox[5em][c]{} \makebox[2em]{+} \makebox[4.8em][c]{SoftPlus} & 5.80$_{0.34}$ & 5.53$_{0.27}$ & 10.13$_{1.28}$ & 99.96$_{0.01}$ \\ 
            \rowcolor[gray]{.93}\makebox[5em][c]{} \makebox[2em]{+} \makebox[4.8em][c]{LBE (\textBF{Ours})} & 3.75$_{0.43}$ & 3.67$_{0.30}$ & \phantom{0}3.22$_{0.80}$ & 99.96$_{0.02}$ \\ 
            \makebox[5em][c]{Absolute} \makebox[2em]{+} \makebox[4.8em][c]{} & 4.22$_{0.53}$ & 4.15$_{0.45}$ & \phantom{0}4.00$_{0.38}$ & 99.98$_{0.02}$ \\
            \makebox[5em][c]{Huber} \makebox[2em]{+} \makebox[4.8em][c]{} & 4.20$_{0.18}$ & 4.28$_{0.25}$ & \phantom{0}4.57$_{1.28}$ & 99.98$_{0.01}$ \\
            \makebox[5em][c]{Exponential} \makebox[2em]{+} \makebox[4.8em][c]{} & 5.35$_{0.27}$ & 5.32$_{0.29}$ & \phantom{0}6.68$_{1.74}$ & 99.98$_{0.01}$ \\
            \bottomrule
        \end{tabular}
        }
    }
    \label{table:ablation}
    \vspace{-0.3cm}
\end{table}

%% file: appendix.tex
\setcounter{page}{1}
\setcounter{section}{0}
\setcounter{figure}{0}
\setcounter{table}{0}
\begin{center}
    {\textbf{Ordinal-Aware Calibration for Ordinal Classification}
    }\\
    {\textemdash\ \textbf{Appendix} \textemdash}
\end{center}

\section{Overview}
\label{suppl_sec:overview}
This appendix provides additional details and analyses to complement the findings presented in the main paper. In particular, it includes the following sections: 

\textbf{Section~\ref{suppl_sec:datasets_and_implementation}} summarizes the datasets and implementation details. \textbf{Section~\ref{suppl_sec:metrics}} describes the performance metrics used to evaluate both calibration and classification performance. \textbf{Section~\ref{suppl_sec:theorectical_proof}} presents the theoretical analysis of ORCU, including the overconfidence induced by one-hot cross-entropy (Section~\ref{suppl_subsec:proof_ce_overconfidence}), the role of the full-support SoftCE anchor and the well-posedness of the ORCU objective (Section~\ref{supple_subsec:proof_sps}), the all-scale unimodality of the free-logit optimum with mode at the observed class (Section~\ref{suppl_subsec:all_scale_unimodality}), threshold-wise probability redistribution (Section~\ref{suppl_subsec:probability_redistribution}), and the interpretation of adjacent gaps as ordinal uncertainty proxies (Section~\ref{suppl_subsec:proof_uncertainty}). \textbf{Section~\ref{suppl_sec:gradients}} derives the gradients of \(\mathcal{L}_{\text{ORCU}}\), detailing how SoftCE provides target anchoring while LBE supplies ordinal structure and feasible-interior confidence refinement. \textbf{Section~\ref{suppl_sec:results}} presents extended results on classification and generalization performance (Sections~\ref{suppl_subsec:generalization_performance} and \ref{suppl_subsec:classification_performance}), visual analyses (Section~\ref{suppl_subsec:visual_analyses}) including reliability diagrams, t-SNE plots, softmax output distributions, and Grad-CAM visualizations, training curves (Section~\ref{suppl_subsec:training_curve}), as well as a comprehensive ablation study (Section~\ref{suppl_subsec:ablation}) assessing the contributions of key components in \(\mathcal{L}_{\text{ORCU}}\).

Together, these sections offer deeper insights into both the methodology and empirical outcomes, further validating the robustness of the proposed approach. 

\section{Datasets and Implementation Details}
\label{suppl_sec:datasets_and_implementation}
\input{tables/suppl_tb1_dataset}
\subsection{Datasets}
\label{suppl_subsec:datasets}
We evaluated the proposed $\mathcal{L}_{\text{ORCU}}$ loss function on four publicly available image datasets, selected for their ordinal characteristics and relevance to distinct application domains, as well as one text dataset, used in a preliminary study to assess ORCU's potential for domain generalization. Details on the label distributions for each dataset are summarized in Table~\ref{suppl_table:dataset}.

\noindent\textbf{Image Aesthetics dataset:} This dataset~\cite{schifanella:beauty:icwsm15}, derived from Flickr, includes 13,364 valid images annotated with 5 ordinal aesthetic score levels, reflecting subjective aesthetic judgments. For evaluation, we employed an 80/20 train-test split and performed five-fold cross-validation to ensure reliable performance measurement.

\noindent\textbf{Adience dataset:} The Adience dataset~\cite{eidinger2014age} comprises 26,580 images grouped into 8 ordinal age categories and is widely used for age estimation tasks where the inherent order of age groups is critical. Of these, 17,423 images are included in the predefined folds. We evaluated using five-fold cross-validation, leveraging the predefined folds to maintain a consistent train-test split ratio of approximately 80/20.

\noindent\textbf{LIMUC dataset:} This medical dataset~\cite{polat2022labeled} contains 11,276 images of ulcerative colitis patients labeled with 4 ordinal Mayo Endoscopic Scores (MES), representing disease severity. To ensure robust and clinically relevant evaluations, we adopted a subject-exclusive train-test split protocol \cite{paplham2024call} with an 85/15 ratio, alongside a ten-fold cross-validation approach \cite{polat2022class}. 

\noindent\textbf{Diabetic Retinopathy dataset:} This dataset comprises retinal fundus images from the Kaggle 2015 Diabetic Retinopathy Detection and APTOS 2019 Blindness Detection competitions. Each image is resized and cropped to a maximum dimension of $1024 \times 1024$ pixels. The images are labeled from 0 to 4 based on the severity of diabetic retinopathy. To address the class imbalance in the original dataset of 88,702 images, we performed uniform sampling based on the class with the smallest sample size (severity level 4), ensuring each class contained 1,914 images. This process resulted in a balanced subset of 9,570 images. For evaluation, we applied an 80/20 train-test split and conducted five-fold cross-validation.

\noindent\textbf{SST-5 dataset:} The Stanford Sentiment Treebank (SST-5)~\cite{socher2013recursive} dataset comprises 11,855 movie review phrases, each annotated with one of five ordinal sentiment labels: strongly negative, negative, neutral, positive, or strongly positive. Following prior work~\cite{castagnos2022simple} in text ordinal classification, we adopted a 7:1:2 split for training, validation, and test sets, respectively.

\subsection{Implementation Details}
\label{suppl_subsec:implementations}
For all experiments on image datasets, we used a ResNet-50 backbone~\cite{he2015deepresiduallearningimage} pretrained on ImageNet~\cite{russakovsky2015imagenet}. To assess the generalizability of our method, we additionally evaluated ResNet-34 and ResNet-101 under the same implementation settings, as described below.

\input{figures/tex/suppl_fig1_tfigure}
\subsubsection{Experiments on Image Data} \

\noindent\textbf{Learning Strategy:} A layer-wise learning rate scheme was adopted, assigning an initial rate of 0.01 to the fully connected layer and 0.001 to all others. Optimization was performed using AdamW \cite{loshchilov2019decoupledweightdecayregularization} with a weight decay of 0.01. Learning rates were adjusted dynamically via the ReduceLROnPlateau scheduler based on validation loss.

\noindent\textbf{Training Protocol:} Models were trained for 100 epochs with a batch size of 64. Input images were resized to $224 \times 224$ and augmented prior to training. All experiments were conducted using NVIDIA RTX 4090 GPUs.

\noindent\textbf{Data Augmentation:} Augmentation included horizontal and vertical flipping (each with probability 0.5), rotation (up to 20 degrees, probability 0.5), and normalization using dataset-specific training set statistics. All transformations were implemented with the Albumentations library \cite{Buslaev_2020}.

\noindent\textbf{Post-hoc Calibration:} 
We applied Temperature Scaling (TS)~\cite{guo2017calibration} and Dirichlet Calibration (DC)~\cite{kull2019beyond} using the validation set, optimizing all parameters to minimize the negative log-likelihood (NLL). We tuned TS via grid search over the $[0, 10]$ range with a step size of $0.1$. For DC, we appended a single-layer network to the logits while keeping the backbone parameters fixed, and selected calibration parameters through grid search over $\lambda \in \{0, 0.0025, 0.005, 0.01, 0.025, 0.05, 0.1, 0.25, \\ 0.5, 1, 2.5, 5, 10\}$ and $\mu \in \{0, 0.01, 0.1, 1, 10\}$.

\noindent\textbf{ORCU:} The $\mathcal{L}_{\text{ORCU}}$ loss employed a squared distance metric for soft encoding. The optimal value of the barrier scale $s$ was identified through validation experiments on the most balanced dataset, Diabetic Retinopathy, using $s \in \{1.0, 3.0, 5.0, 7.0, 10.0\}$. The hyperparameter barrier scale $s = 3.0$, which achieved the lowest SCE (Fig.~\ref{suppl_figure:t_dr}), was selected and subsequently used in all experiments. It also yielded the lowest ACE and ECE, and the same trend was observed on the validation set of the most imbalanced dataset, Image Aesthetics (Fig.~\ref{suppl_figure:t_ia}), supporting its generalizability across datasets.

\subsubsection{Experiment on Text Data}\ 

\noindent For the text data experiment, we used the BERT-tiny model~\cite{turc2019well}, following prior work in text-based ordinal classification~\cite{castagnos2022simple}.

\noindent\textbf{Tokenization and Preprocessing:} All input texts were tokenized using HuggingFace’s AutoTokenizer, with truncation and padding applied to a fixed maximum sequence length of 128 tokens.

\noindent\textbf{Training Configuration:} We trained the model for 100 epochs with a batch size of 256 using the AdamW optimizer~\cite{loshchilov2019decoupledweightdecayregularization} and a weight decay of 0.01. We set the initial learning rate to 5e$-$4 and decayed it linearly to zero throughout training without warmup. All experiments were conducted on NVIDIA RTX 4090 GPUs.

\noindent\textbf{ORCU:} For text data, we selected the barrier scale \(s\) based on validation performance. Unlike image-based tasks, lower values of \(s\) were more effective. We evaluated \(s\) over the range \(\{0.01, 0.05, 0.1, 0.5, 1.0\}\) and selected \(s = 0.05\), which achieved the lowest ECE on the validation set.

\section{Performance Metrics}
\label{suppl_sec:metrics}

This section describes the metrics used to evaluate both calibration and classification performance in our experiments. While Expected Calibration Error (ECE) is widely adopted for assessing calibration, it has limitations in multi-class and ordinal settings due to its coarse granularity and lack of class-specific evaluation. To address these limitations, we additionally report Static Calibration Error (SCE) and Adaptive Calibration Error (ACE), which are designed to provide more accurate calibration assessment in such contexts~\cite{Nixon_2019_CVPR_Workshops}. For classification performance, we include conventional metrics such as Accuracy (Acc) and Mean Absolute Error (MAE), as well as Quadratic Weighted Kappa (QWK)~\cite{ben2008comparison}, which captures ordinal consistency, and \%Unimodality~\cite{cardoso2023unimodal}, which quantifies the alignment of predictions with ordinal structure.

\subsection{Calibration Metrics}
\label{suppl_subsec:calibration_metrics}

We use three calibration metrics to evaluate how well predicted probabilities reflect true likelihoods: ECE, SCE, and ACE. Expected Calibration Error (ECE) is the most commonly used due to its simplicity. ECE partitions the $[0, 1]$ confidence interval into $B$ equally spaced bins. For each bin $b$, it computes the accuracy $\text{acc}(b)$ as the proportion of correctly predicted samples and the confidence $\text{conf}(b)$ as the average predicted probability in that bin. ECE is then defined as:
\begin{equation}
\text{ECE} = \sum_{b=1}^{B} \frac{n_b}{N} \big| \text{acc}(b) - \text{conf}(b) \big|,
\end{equation}
where $n_b$ is the number of predictions in bin $b$, and $N$ is the total number of predictions. However, ECE's reliance on fixed binning and lack of class-specific evaluation can lead to inaccurate estimates, especially in imbalanced or ordinal tasks.

Static Calibration Error (SCE) improves upon ECE by computing calibration error separately for each class and then averaging across all classes. This class-wise decomposition provides finer-grained insights and is particularly suitable for ordinal tasks, where the structure across labels matters. SCE is defined as:
\begin{equation}
\text{SCE} = \frac{1}{K} \sum_{k=1}^{K} \sum_{b=1}^{B} \frac{n_{b,k}}{N} \big| \text{acc}(b,k) - \text{conf}(b,k) \big|,
\end{equation}
where $n_{b,k}$ is the number of predictions for class $k$ in bin $b$, and $\text{acc}(b,k)$ and $\text{conf}(b,k)$ are the accuracy and confidence for class $k$ in that bin. By capturing class-specific calibration, SCE allows for a more detailed evaluation, essential for tasks with structured label spaces.

Adaptive Calibration Error (ACE) further refines this evaluation by employing adaptive binning, ensuring that each bin contains an equal number of predictions. This dynamic binning mitigates sparsity issues often encountered with fixed-bin approaches and yields more stable calibration estimates, especially in class-imbalanced settings. ACE is defined as:
\begin{equation}
\text{ACE} = \frac{1}{KR} \sum_{k=1}^{K} \sum_{r=1}^{R} \big| \text{acc}(r,k) - \text{conf}(r,k) \big|,
\end{equation}
where $R$ is the number of adaptive bins, and $\text{acc}(r,k)$ and $\text{conf}(r,k)$ denote the accuracy and confidence in range $r$ for class $k$.

Together, these metrics reveal the limitations of ECE in capturing calibration behavior under multi-class and ordinal conditions. While SCE improves granularity through class-specific decomposition, ACE enhances robustness via adaptive binning. The complementary use of ECE, SCE, and ACE thus enables a comprehensive and reliable assessment of calibration performance in ordinal classification tasks.

\subsection{Ordinal Classification Metrics}
\label{suppl_subsec:classification_metrics}

This section introduces the metrics Quadratic Weighted Kappa (QWK)~\cite{ben2008comparison} and \%Unimodality~\cite{cardoso2023unimodal}, both used to assess the classification performance of ORCU in capturing ordinal relationships.

QWK quantifies the agreement between predicted and target labels while explicitly considering the ordinal structure of the label space~\cite{ben2008comparison}. Unlike standard accuracy-based metrics, QWK applies a quadratic weighting scheme that heavily penalizes more significant discrepancies, making it especially effective for tasks where the relative distances between classes are meaningful. It is defined as:
\begin{equation}
\text{QWK} = 1 - \frac{\sum_{i=1}^{C} \sum_{j=1}^{C} w_{ij} O_{ij}}{\sum_{i=1}^{C} \sum_{j=1}^{C} w_{ij} E_{ij}},
\end{equation}
where \( C \) denotes the number of classes, \( w_{ij} = \frac{(i - j)^2}{(C - 1)^2} \) represents the penalty assigned to a prediction error between classes \( i \) and \( j \), \( O_{ij} \) is the observed count, and \( E_{ij} \) is the expected count under random chance. QWK ranges from \(-1\) (complete disagreement) to \(1\) (perfect agreement), with higher values indicating stronger alignment between predictions and ground truth. By accounting for both correctness and the severity of misclassification, QWK provides a robust and informative metric for ordinal classification.

\%Unimodality evaluates how well the predicted probability distributions conform to the unimodality property relative to the predicted class. Let \( \hat{y} \in \mathbb{R}^C \) denote the predicted probability distribution over \( C \) ordinal classes, and let \( y \in \{1, \dots, C\} \) be the ground-truth label.\;Let \( \hat{c} = \arg\max_{k \in \{1,\dots,C\}} \hat{y}_k \) denote the predicted class. A prediction satisfies unimodality if the probabilities increase monotonically up to class \( \hat{c} \) and decrease monotonically thereafter:
\begin{equation}
\hat{y}_{1} \leq \hat{y}_{2} \leq \dots \leq \hat{y}_{\hat{c}} \quad \text{and} \quad \hat{y}_{\hat{c}} \geq \hat{y}_{\hat{c}+1} \geq \dots \geq \hat{y}_{C}.
\end{equation}
Each prediction is assigned a binary indicator \( U \in \{0, 1\} \) based on whether this condition is met:
\begin{equation}
U = 
\begin{cases} 
1, & \text{if unimodality is satisfied}, \\
0, & \text{otherwise}.
\end{cases}
\end{equation}
The final \%Unimodality score is computed as the proportion of predictions that satisfy this property:
\begin{equation}
\% \text{Unimodality} = \frac{1}{N} \sum_{n=1}^{N} U_n,
\end{equation}
where \( N \) is the number of predictions and \( U_n \) indicates unimodality compliance for the \( n \)-th prediction. This metric directly evaluates the model’s ability to produce structured probability outputs that respect ordinal label relations.

QWK and \%Unimodality are particularly well-suited for evaluating the proposed ORCU loss, as they jointly assess its ability to model ordinal relationships during training and generate predictions that align with the underlying label structure.

\section{Structural Theory of ORCU}
\label{suppl_sec:theorectical_proof}
\begingroup
This section analyzes ORCU in the per-sample free-logit setting. We fix one training example, treat its logits \(\mathbf z\in\mathbb R^C\) as optimization variables, and denote the ground-truth class by \(y\). These statements characterize the geometry induced by the loss itself; they do not constitute a global convergence theorem for the shared neural-network parameterization or a population calibration guarantee.

\subsection{Logit-Space Overconfidence under One-Hot Cross-Entropy}
\label{suppl_subsec:proof_ce_overconfidence}
Let \(\mathbf e_y\) be the one-hot target and \(p=\operatorname{softmax}(\mathbf z)\). The one-hot CE loss is
\begin{equation}
\mathcal L_{\mathrm{CE}}(\mathbf z;y)=-\log p_y.
\end{equation}
Its logit gradient is \(\partial\mathcal L_{\mathrm{CE}}/\partial z_k=p_k-\mathbf1[k=y]\). Under a direct logit-space descent step with learning rate \(\eta>0\), the margin between the target and any non-target class satisfies
\begin{equation}
(z_y^+-z_k^+)-(z_y-z_k)
=\eta\bigl[(1-p_y)+p_k\bigr]>0,\qquad k\ne y.
\end{equation}
Thus, every loss-decreasing step enlarges all target-to-nontarget margins. Since
\begin{equation}
p_y=\frac{1}{1+\sum_{k\ne y} \exp(z_k-z_y)},
\end{equation}
driving \(\mathcal L_{\mathrm{CE}}\) to its infimum requires \(z_y-z_k\to+\infty\) for all \(k\ne y\), and hence \(p_y\to1\). One-hot CE therefore lacks a finite logit-space optimum and pushes the distribution toward a near-delta prediction, motivating a full-support ordinal target.

\subsection{SoftCE Anchor and All-Scale Well-Posedness}
\label{supple_subsec:proof_sps}
Let \(a=(a_1,\ldots,a_C)\in\operatorname{int}\Delta^{C-1}\) denote the soft-encoded target for class \(y\), with
\begin{equation}
a_k=\frac{\exp[-(y-k)^2]}{\sum_{j=1}^C\exp[-(y-j)^2]}.
\end{equation}
Then \(a\) is strictly unimodal with unique mode \(y\), i.e., \(a_1<\cdots<a_y>\cdots>a_C\). Define the directed adjacent-gap operator \(B_y\in\mathbb R^{(C-1)\times C}\) by
\begin{equation}
(B_y\mathbf z)_k=r_k=
\begin{cases}
z_k-z_{k+1}, & k<y,\\
z_{k+1}-z_k, & k\ge y.
\end{cases}
\end{equation}
The per-sample ORCU objective is
\begin{equation}
F_s(\mathbf z)=\log\sum_{j=1}^C e^{z_j}-a^\top\mathbf z+
\sum_{k=1}^{C-1}\hat I_s(r_k),\qquad r=B_y\mathbf z,
\end{equation}
where \(\hat I_s\) is the LBE function used in the main text and
\begin{equation}
\hat I_s'(r)=
\begin{cases}
-\dfrac{1}{sr}, & r\le-1/s^2,\\[4pt]
s, & r>-1/s^2.
\end{cases}
\label{suppl_eq:Iprime_theory}
\end{equation}

\begin{theorem}[SoftCE strict propriety]\label{thm:softce_proper_green}
The SoftCE component \(-\sum_k a_k\log p_k\) is strictly proper with respect to the designed soft target \(a\) and is uniquely minimized over the probability simplex at \(p=a\).
\end{theorem}
\begin{proof}
We have
\begin{equation}
-\sum_{k=1}^C a_k\log p_k=H(a)+D_{\mathrm{KL}}(a\|p),
\end{equation}
where \(H(a)\) is independent of \(p\). Since \(a\) has full support, the KL term is nonnegative and equals zero only at \(p=a\). The softmax preimages of \(a\) are \(\log a+c\mathbf1\), and these logits are strictly unimodal around \(y\) because \(a\) is strictly unimodal.
\end{proof}

\begin{proposition}[The full-support anchor is necessary]\label{prop:anchor_necessary_green}
LBE alone, and also one-hot CE combined with LBE, is unbounded below. The full-support SoftCE target supplies the coercive ordinal anchor needed for a finite refinement.
\end{proposition}
\begin{proof}
Consider the ordinal ray \(z_j(t)=-t|j-y|\), \(t>0\), which sharpens the logit profile around \(y\). Then every directed gap satisfies \(r_k(t)=-t\). For \(t>1/s^2\), all gaps lie in the log branch and
\begin{equation}
\sum_{k=1}^{C-1}\hat I_s(r_k(t))=-\frac{C-1}{s}\log t\to-\infty.
\end{equation}
Thus, LBE alone is unbounded below. Along the same ray, one-hot CE satisfies \(-\log p_y(z(t))\to0\), so CE+LBE is also unbounded below. In contrast,
\begin{equation}
\begin{aligned}
\mathcal L_{\mathrm{SoftCE}}(z(t);a)
&=\log\sum_j e^{z_j(t)}-a^\top z(t)\\
&\sim t\sum_{j=1}^C a_j|j-y|\to+\infty.
\end{aligned}
\end{equation}
SoftCE grows linearly along the ray, whereas the negative LBE growth is only logarithmic. Hence the SoftCE anchor prevents this unbounded descent.
\end{proof}

\begin{theorem}[All-scale finite equilibrium]\label{thm:all_s_wellposed_green}
For every \(s>0\), \(F_s\) has a unique finite minimizer on the shift-free subspace \(\mathcal H=\{\mathbf z:\mathbf1^\top\mathbf z=0\}\). Equivalently, the minimizer in \(\mathbb R^C\) is unique modulo uniform logit shifts.
\end{theorem}
\begin{proof}
Softmax probabilities and adjacent gaps are invariant to \(\mathbf z\mapsto \mathbf z+c\mathbf1\), so it suffices to work on \(\mathcal H\). The Hessian of SoftCE is \(\operatorname{diag}(p)-pp^\top\). For any nonconstant \(\mathbf v\),
\begin{equation}
\mathbf v^\top(\operatorname{diag}(p)-pp^\top)\mathbf v=\operatorname{Var}_{K\sim p}(v_K)>0,
\end{equation}
so SoftCE is strictly convex on \(\mathcal H\). The LBE term is convex because \(\hat I_s'\) is nondecreasing and the gaps are linear maps of \(\mathbf z\). Thus, \(F_s\) is strictly convex on \(\mathcal H\).

It remains to show coercivity. Let \(R(\mathbf z)=\max_j z_j-\min_j z_j\) and \(a_{\min}=\min_j a_j>0\). Since \(\log\sum_j e^{z_j}\ge\max_j z_j\),
\begin{equation}
\mathcal L_{\mathrm{SoftCE}}(\mathbf z;a)
\ge \max_j z_j-a^\top\mathbf z
\ge a_{\min}R(\mathbf z).
\end{equation}
For each adjacent gap, \(|r_k|\le R(\mathbf z)\), and the lowest possible LBE growth is logarithmic; hence for a finite constant \(C_s\),
\begin{equation}
\mathcal L_{\mathrm{LBE}}(\mathbf z;y)
\ge -\frac{C-1}{s}\log(1+R(\mathbf z))-C_s.
\end{equation}
Therefore,
\begin{equation}
F_s(\mathbf z)
\ge a_{\min}R(\mathbf z)-\frac{C-1}{s}\log(1+R(\mathbf z))-C_s\to+\infty
\end{equation}
as \(R(\mathbf z)\to\infty\). Coercivity gives existence, and strict convexity gives uniqueness on \(\mathcal H\).
\end{proof}

\subsection{All-Scale Unimodality of the Free-Logit Optimum}
\label{suppl_subsec:all_scale_unimodality}
\begin{theorem}[Strict unimodality with mode \(y\)]\label{thm:all_s_unimodal_green}
If \(a_1<\cdots<a_y>\cdots>a_C\), then for every \(s>0\) the unique minimizer \(\mathbf z_s^\star\) of \(F_s\) satisfies
\begin{equation}
(B_y\mathbf z_s^\star)_k<0,
\qquad k=1,\ldots,C-1.
\end{equation}
Consequently, \(\operatorname{softmax}(\mathbf z_s^\star)\) is strictly unimodal with mode \(y\).
\end{theorem}
\begin{proof}
Let \(p_s=\operatorname{softmax}(\mathbf z_s^\star)\), \(r_s=B_y\mathbf z_s^\star\), and \(q_{s,k}=\hat I_s'(r_{s,k})\). From \eqref{suppl_eq:Iprime_theory}, \(0<q_{s,k}\le s\). The stationarity condition is
\begin{equation}
p_s-a+B_y^\top\mathbf q_s=0,
\qquad\text{or}\qquad
p_s=a-B_y^\top\mathbf q_s.
\label{suppl_eq:stationarity_green}
\end{equation}
Set \(q_{s,0}=q_{s,C}=0\). Coordinate-wise,
\begin{equation}
p_{s,j}-a_j=
\begin{cases}
q_{s,j-1}-q_{s,j}, & j<y,\\
q_{s,y-1}+q_{s,y}, & j=y,\\
q_{s,j}-q_{s,j-1}, & j>y.
\end{cases}
\label{suppl_eq:coordinate_balance_green}
\end{equation}
Consider an edge \(k<y\). If \(r_{s,k}\ge0\), then this edge is in the linear branch and \(q_{s,k}=s\). Equation~\eqref{suppl_eq:coordinate_balance_green} gives \(p_{s,k}\le a_k\) and \(p_{s,k+1}\ge a_{k+1}\). Since \(a_k<a_{k+1}\), we obtain \(p_{s,k}<p_{s,k+1}\). But \(r_{s,k}=z_{s,k}-z_{s,k+1}\ge0\) implies \(p_{s,k}\ge p_{s,k+1}\), a contradiction. Hence \(r_{s,k}<0\) for all \(k<y\). The case \(k\ge y\) is symmetric. Assuming \(r_{s,k}=z_{s,k+1}-z_{s,k}\ge0\) gives \(p_{s,k}>p_{s,k+1}\), contradicting softmax order preservation. Thus all directed gaps are strictly negative.
\end{proof}

This result shows that large \(s\) is not required to obtain ordinal feasibility. The role of large \(s\) is to make the regularized optimum close to the SoftCE anchor. If the anchor logits \(\log a+c\mathbf1\) have directed gaps bounded above by \(-\alpha<0\), local strong convexity around the SoftCE optimum yields the standard stability estimate
\begin{equation}
\|\mathbf z_s^\star-(\log a+c\mathbf1)\|_2=O\!\left(\frac{1}{s\alpha}\right)
\end{equation}
for sufficiently large \(s\), after fixing the uniform shift. Thus, all-scale unimodality and large-\(s\) proximity play different roles.

\subsection{Threshold-Wise Probability Redistribution}
\label{suppl_subsec:probability_redistribution}
\begin{theorem}[Cumulative threshold refinement]\label{thm:cdf_refinement_green}
At the unique free-logit ORCU optimum, the probability vector \(p_s\) satisfies
\begin{equation}
p_s=a-B_y^\top\mathbf q_s,
\qquad 0<q_{s,k}\le s.
\end{equation}
If \(F_p(k)=\sum_{j=1}^k p_j\), then
\begin{equation}
F_{p_s}(k)=
\begin{cases}
F_a(k)-q_{s,k}, & k<y,\\
F_a(k)+q_{s,k}, & k\ge y.
\end{cases}
\label{suppl_eq:cdf_identity_green}
\end{equation}
Thus, \(q_{s,k}=|F_{p_s}(k)-F_a(k)|\) is exactly the cumulative probability mass shifted across threshold \(k\) toward the observed class.
\end{theorem}
\begin{proof}
The first identity is the stationarity condition \eqref{suppl_eq:stationarity_green}. Summing the coordinate balance \eqref{suppl_eq:coordinate_balance_green} from class \(1\) to \(k\) telescopes all interior terms and leaves only the boundary response \(q_{s,k}\), with the sign determined by whether the threshold lies to the left or right of \(y\).
\end{proof}

\begin{corollary}[Concentration around the observed class]\label{cor:target_concentration_green}
The probability of the observed class increases relative to the SoftCE anchor, so \(p_{s,y}>a_y\). More generally, every proper interval \([\ell,r]\) containing \(y\) gains probability mass.
\begin{equation}
P_{p_s}(\ell\le K\le r)=P_a(\ell\le K\le r)+q_{s,\ell-1}+q_{s,r},
\end{equation}
with \(q_{s,0}=q_{s,C}=0\).
\end{corollary}

\begin{corollary}[RPS improvement relative to the anchor]\label{cor:rps_green}
Let \(t_k=\mathbf1[y\le k]\) be the ordinal threshold outcome and define
\begin{equation}
\operatorname{RPS}(p,y)=\sum_{k=1}^{C-1}\bigl(F_p(k)-t_k\bigr)^2.
\end{equation}
Then \(\operatorname{RPS}(p_s,y)<\operatorname{RPS}(a,y)\). Specifically,
\begin{equation}
\begin{aligned}
\operatorname{RPS}(a,y)-\operatorname{RPS}(p_s,y)
={}&\sum_{k<y}q_{s,k}\bigl(2F_a(k)-q_{s,k}\bigr)\\
&+\sum_{k\ge y}q_{s,k}\bigl(2(1-F_a(k))-q_{s,k}\bigr)\\
&>0.
\end{aligned}
\end{equation}
\end{corollary}
\begin{proof}
For \(k<y\), the threshold outcome is \(0\) and \(F_{p_s}(k)=F_a(k)-q_{s,k}\), so the squared threshold error decreases by \(q_{s,k}(2F_a(k)-q_{s,k})\). Since \(F_{p_s}(k)\ge0\), \(q_{s,k}\le F_a(k)\), making the term positive. For \(k\ge y\), the threshold outcome is \(1\) and \(F_{p_s}(k)=F_a(k)+q_{s,k}\), yielding the second sum; positivity follows from \(F_{p_s}(k)\le1\).
\end{proof}

\begin{corollary}[KL-controlled deviation]\label{cor:kl_control_green}
The ORCU equilibrium remains controlled around the SoftCE anchor.
\begin{equation}
D_{\mathrm{KL}}(a\|p_s)\le\frac{C-1}{s}.
\end{equation}
Consequently, \(\|p_s-a\|_1\le\sqrt{2(C-1)/s}\) by Pinsker's inequality.
\end{corollary}
\begin{proof}
Let \(z_a=\log a+c\mathbf1\). Since \(\mathcal L_{\mathrm{SoftCE}}(z_s)-\mathcal L_{\mathrm{SoftCE}}(z_a)=D_{\mathrm{KL}}(a\|p_s)\), convexity gives
\begin{equation}
D_{\mathrm{KL}}(a\|p_s)
\le \left\langle \nabla\mathcal L_{\mathrm{SoftCE}}(z_s), z_s-z_a\right\rangle.
\end{equation}
By stationarity, \(\nabla\mathcal L_{\mathrm{SoftCE}}(z_s)=p_s-a=-B_y^\top\mathbf q_s\). Hence
\begin{equation}
D_{\mathrm{KL}}(a\|p_s)
\le -\mathbf q_s^\top B_y(z_s-z_a)
= -\mathbf q_s^\top r_s+\mathbf q_s^\top r_a.
\end{equation}
Because \(a_1<\cdots<a_y>\cdots>a_C\), we have \(r_a=B_yz_a<0\), so \(\mathbf q_s^\top r_a<0\). It remains to bound \(-q_{s,k}r_{s,k}\). By Theorem~\ref{thm:all_s_unimodal_green}, \(r_{s,k}<0\). If the edge is in the log branch, \(-q_{s,k}r_{s,k}=1/s\); if it is in the linear branch, then \(q_{s,k}=s\) and \(-r_{s,k}<1/s^2\), so \(-q_{s,k}r_{s,k}<1/s\). Summing over \(C-1\) edges gives the bound.
\end{proof}

\subsection{Adjacent Logit Gaps as Ordinal Uncertainty Proxies}
\label{suppl_subsec:proof_uncertainty}
For a model prediction \(\hat{\mathbf y}_n=\operatorname{softmax}(\mathbf z_n)\), define
\begin{equation}
 r_{n,k}=
 \begin{cases}
   z_{n,k}-z_{n,k+1}, & k<y_n,\\
   z_{n,k+1}-z_{n,k}, & k\ge y_n,
 \end{cases}
\end{equation}
and the adjacent violation
\begin{equation}
 v_{n,k}=
 \begin{cases}
   [\hat y_{n,k}-\hat y_{n,k+1}]_+, & k<y_n,\\
   [\hat y_{n,k+1}-\hat y_{n,k}]_+, & k\ge y_n.
 \end{cases}
\end{equation}
Let \(m_{n,k}=\hat y_{n,k}+\hat y_{n,k+1}\).

{
Define the local farther-side mass fraction
\[
\pi_{n,k}=
\begin{cases}
\hat y_{n,k}/m_{n,k}, & k<y_n,\\
\hat y_{n,k+1}/m_{n,k}, & k\ge y_n,
\end{cases}
=\frac{e^{r_{n,k}}}{1+e^{r_{n,k}}}.
\]
It is monotone in \(r_{n,k}\), equals \(1/2\) at the ordinal boundary, and exceeds \(1/2\) exactly when the adjacent relation places more mass on the ordinally farther side from \(y_n\). Thus, \(r_{n,k}\) gives a probability-level interpretation: increasing it moves the pair from well-separated ordering around \(y_n\) toward near-boundary ambiguity and, once positive, to violation.
}

\begin{proposition}\label{prop:violation_equivalence_green}
For every adjacent pair,
\begin{equation}
r_{n,k}>0\quad\Longleftrightarrow\quad v_{n,k}>0,
\end{equation}
and
\begin{equation}
\frac{v_{n,k}}{m_{n,k}}=\bigl[\tanh(r_{n,k}/2)\bigr]_+.
\end{equation}
\end{proposition}
\begin{proof}
For \(k<y_n\), \(\hat y_{n,k}/\hat y_{n,k+1}=e^{r_{n,k}}\); for \(k\ge y_n\), \(\hat y_{n,k+1}/\hat y_{n,k}=e^{r_{n,k}}\). Thus the sign of \(r_{n,k}\) exactly matches the sign of the directed probability violation. Normalizing the positive difference by the pair mass gives \((e^r-1)/(e^r+1)=\tanh(r/2)\).
\end{proof}

Assume that a reference distribution \(\mathbf p_n\) is unimodal with mode \(y_n\). Then the adjacent violation is bounded by the discrepancy from that reference.
\begin{equation}
 v_{n,k}\le |\hat y_{n,k}-p_{n,k}|+|\hat y_{n,k+1}-p_{n,k+1}|
 \le 2\operatorname{TV}(\mathbf p_n,\hat{\mathbf y}_n).
\end{equation}
Combining this with Pinsker's inequality yields
\begin{equation}
\begin{aligned}
D_{\mathrm{KL}}(\mathbf p_n\|\hat{\mathbf y}_n)
&\ge \frac12 v_{n,k}^2\\
&=\frac12 m_{n,k}^2\tanh^2(r_{n,k}/2),
\qquad r_{n,k}>0.
\end{aligned}
\end{equation}
The directed gap \(r_{n,k}\) therefore provides an input-specific ordinal uncertainty proxy: its sign detects adjacent ordinal violation, the sigmoid of \(r_{n,k}\) gives the farther-side mass fraction within the adjacent pair, and its value tracks the transition from ordered margin to near-boundary ambiguity and violation strength. Larger \(r_{n,k}\) indicates progressively weaker local ordering around \(y_n\); for \(r_{n,k}>0\), it certifies a structural deviation with the KL lower bound above.
\endgroup

\section{Gradient Derivation of the ORCU Loss}
\label{suppl_sec:gradients}

\textbf{Derivation overview.} This appendix provides a detailed derivation of the gradients of the ORCU loss with respect to each logit \(z_{n,k}\), which are reported in Table~I of the main text. The derivation is provided for a single sample \(n\); the overall gradient across a dataset of size \(N\) is obtained by averaging the gradients over all samples.

\subsection{Loss Definition}
\label{suppl_subsec:loss_def}
For each sample \(n\), the ORCU loss is given by
\begin{align}
\label{eq:orcu_loss_single}
\mathcal{L}_{\text{ORCU}}(n)
&= 
\underbrace{-\sum_{k=1}^C y'_{n,k} \,\log \hat{y}_{n,k}}_{\mathcal{L}_{\text{SoftCE}}(n)} \notag \\
&\quad+
\underbrace{\sum_{k=1}^{C-1}
\begin{cases}
\hat{I}\bigl(z_{n,k} - z_{n,k+1}\bigr), & k < y_n,\\
\hat{I}\bigl(z_{n,k+1} - z_{n,k}\bigr), & k \ge y_n,
\end{cases}}_{\mathcal{L}_{\text{LBE}}(n)},
\end{align}
where
\begin{equation}
\hat{y}_{n,k} \;=\; \frac{\exp(z_{n,k})}{\sum_{j=1}^C \exp(z_{n,j})},
\end{equation}
and
\begin{equation}
\label{eq:Ihat_function}
\hat{I}(r) 
\;=\;
\begin{cases}
-\dfrac{1}{s}\,\log\bigl(-r\bigr), 
& r \le -\tfrac{1}{s^2},\\[4pt]
s\,r \;-\; \dfrac{1}{s}\,\log\!\Bigl(\tfrac{1}{s^2}\Bigr) \;+\; \dfrac{1}{s},
& \text{otherwise}.
\end{cases}
\end{equation}
The adjacent logit gap is
\begin{equation}
\label{definition_of_r}
r_{n,k} 
\;=\;
\begin{cases}
z_{n,k} - z_{n,k+1}, & k < y_n,\\
z_{n,k+1} - z_{n,k}, & k \ge y_n.
\end{cases}
\end{equation}

\subsection{Gradient of the SoftCE Term}
\label{suppl_subsec:gradient_softce}
The first component,
\begin{equation}
\mathcal{L}_{\text{SoftCE}}(n)
\;=\;
-\sum_{k=1}^C y'_{n,k}\,\log \hat{y}_{n,k},
\end{equation}
can be differentiated with respect to \(z_{n,k}\) as follows.

\noindent\textbf{Step 1: Derivative of \(\log(\hat{y}_{n,m})\).}
By definition,
\begin{equation}
\hat{y}_{n,m} \;=\; 
\frac{\exp(z_{n,m})}{\sum_{j=1}^C \exp(z_{n,j})}.
\end{equation}
Hence,
\begin{equation}
\frac{\partial}{\partial z_{n,k}}
\,\log\bigl(\hat{y}_{n,m}\bigr)
\;=\;
\frac{1}{\hat{y}_{n,m}}
\,\frac{\partial \hat{y}_{n,m}}{\partial z_{n,k}}.
\end{equation}
Using the well-known derivative of the softmax function,
\begin{equation}
\frac{\partial \hat{y}_{n,m}}{\partial z_{n,k}}
=
\hat{y}_{n,m}(\delta_{mk} - \hat{y}_{n,k}),
\end{equation}
we obtain
\begin{equation}
\frac{\partial}{\partial z_{n,k}} \log(\hat{y}_{n,m})
=
\delta_{mk} - \hat{y}_{n,k}
=
\begin{cases}
1 - \hat{y}_{n,k}, & m = k,\\
-\hat{y}_{n,k}, & m \neq k.
\end{cases}
\end{equation}

\noindent\textbf{Step 2: Summation over \(m\).}
Thus,
\begin{equation}
\frac{\partial \mathcal{L}_{\text{SoftCE}}(n)}{\partial z_{n,k}}
\;=\;
-\sum_{m=1}^C
y'_{n,m}
\,\frac{\partial}{\partial z_{n,k}}
\,\log\bigl(\hat{y}_{n,m}\bigr).
\end{equation}
Substituting the piecewise derivative yields
\begin{equation}
\frac{\partial \mathcal{L}_{\text{SoftCE}}(n)}{\partial z_{n,k}}
\;=\;
-
\Bigl[
\,y'_{n,k}\,\bigl(1 - \hat{y}_{n,k}\bigr)
\;+\;
\sum_{\substack{m=1 \\ m \neq k}}^C
y'_{n,m}\,(-\hat{y}_{n,k})
\Bigr].
\end{equation}
Rearranging terms recovers the standard result:
\begin{equation}
\frac{\partial \mathcal{L}_{\text{SoftCE}}(n)}{\partial z_{n,k}}
\;=\;
\hat{y}_{n,k}
\;-\;
y'_{n,k}.
\end{equation}

\subsection{Gradient of the Regularization Term}
\label{suppl_subsec:gradient_reg} 
The regularization component is
\begin{equation}
\mathcal{L}_{\text{LBE}}(n)
\;=\;
\sum_{k=1}^{C-1}
\begin{cases}
\hat{I}(z_{n,k} - z_{n,k+1}), & k < y_n,\\
\hat{I}(z_{n,k+1} - z_{n,k}), & k \ge y_n,
\end{cases}
\end{equation}
which can be equivalently written with a dummy index \(i\) as
\begin{equation}
\label{eq:Lreg_with_i}
\mathcal{L}_{\text{LBE}}(n)
\;=\;
\sum_{i=1}^{C-1}
\hat{I}\bigl(r_{n,i}\bigr),
\end{equation}
where \(r_{n,i}\) is given by Equation~\ref{definition_of_r}.

\noindent\textbf{Step 1: Derivative of \(\hat{I}(r)\).}
By Equation~\ref{eq:Ihat_function},
\begin{equation}
\hat{I}(r) 
\;=\;
\begin{cases}
-\dfrac{1}{s}\,\log(-r), 
& r \le -\tfrac{1}{s^2},\\[4pt]
s\,r - \dfrac{1}{s}\,\log\bigl(\tfrac{1}{s^2}\bigr) + \dfrac{1}{s},
& \text{otherwise}.
\end{cases}
\end{equation}
Since \(\tfrac{d}{dr}\log(-r) = \tfrac{1}{r}\), we obtain
\begin{equation}
\label{eq:Ihat_prime}
\hat{I}'(r)
\;=\;
\begin{cases}
-\dfrac{1}{s\,r}, 
& r \le -\tfrac{1}{s^2},\\[4pt]
s, 
& \text{otherwise}.
\end{cases}
\end{equation}

\noindent\textbf{Step 2: Local contribution of a single pair \((i,i+1)\).}
For each adjacent pair \(i\), the gap \(r_{n,i}\) is
\begin{equation}
r_{n,i}
=
\begin{cases}
z_{n,i} - z_{n,i+1}, & i < y_n,\\
z_{n,i+1} - z_{n,i}, & i \ge y_n.
\end{cases}
\end{equation}
The partial derivatives of \(r_{n,i}\) with respect to the two logits in the pair are
\begin{equation}
\label{eq:drdz_local}
\frac{\partial r_{n,i}}{\partial z_{n,i}}
=
\begin{cases}
+1, & i < y_n,\\
-1, & i \ge y_n,
\end{cases}
\qquad
\frac{\partial r_{n,i}}{\partial z_{n,i+1}}
=
\begin{cases}
-1, & i < y_n,\\
+1, & i \ge y_n.
\end{cases}
\end{equation}
Applying the chain rule to the term \(\hat{I}\bigl(r_{n,i}\bigr)\) gives its contribution to the gradients of \(z_{n,i}\) and \(z_{n,i+1}\):
\begin{equation}
\label{eq:local_pair_contrib}
\frac{\partial}{\partial z_{n,i}} \hat{I}\bigl(r_{n,i}\bigr)
=
\hat{I}'(r_{n,i})\,
\frac{\partial r_{n,i}}{\partial z_{n,i}}
=
\begin{cases}
\hat{I}'(r_{n,i}), & i < y_n,\\
-\hat{I}'(r_{n,i}), & i \ge y_n,
\end{cases}
\end{equation}
\begin{equation}
\frac{\partial}{\partial z_{n,i+1}} \hat{I}\bigl(r_{n,i}\bigr)
=
\hat{I}'(r_{n,i})\,
\frac{\partial r_{n,i}}{\partial z_{n,i+1}}
=
\begin{cases}
-\hat{I}'(r_{n,i}), & i < y_n,\\
\hat{I}'(r_{n,i}), & i \ge y_n.
\end{cases}
\end{equation}
These expressions correspond to the single-pair behavior summarized in Table~I of the main text: for each pair \((i,i+1)\), the LBE contribution assigns opposite gradient signs to the two logits, so that a gradient-descent step increases one logit and decreases the other depending on whether the pair lies to the left (\(i<y_n\)) or to the right (\(i\ge y_n\)) of the target index.

\noindent\textbf{Step 3: Aggregating contributions from two adjacent pairs.}
Using Equation~\ref{eq:Lreg_with_i}, the full derivative of \(\mathcal{L}_{\text{LBE}}(n)\) with respect to a single logit \(z_{n,k}\) is obtained by summing over all pairs:
\begin{equation}
\label{eq:Lreg_full_sum}
\frac{\partial \mathcal{L}_{\text{LBE}}(n)}{\partial z_{n,k}}
=
\sum_{i=1}^{C-1}
\hat{I}'\bigl(r_{n,i}\bigr)\,
\frac{\partial r_{n,i}}{\partial z_{n,k}}.
\end{equation}
Because \(z_{n,k}\) appears only in the two pairs \((k-1,k)\) and \((k,k+1)\), the sum in Equation~\ref{eq:Lreg_full_sum} reduces to at most two terms:
\begin{equation}
\label{eq:Lreg_two_terms}
\resizebox{0.98\columnwidth}{!}{$\displaystyle
\frac{\partial \mathcal{L}_{\text{LBE}}(n)}{\partial z_{n,k}}
=
\begin{cases}
\frac{\partial}{\partial z_{n,1}} \hat{I}\bigl(r_{n,1}\bigr),
& k = 1,\\[6pt]
\frac{\partial}{\partial z_{n,k}} \hat{I}\bigl(r_{n,k-1}\bigr)
+
\frac{\partial}{\partial z_{n,k}} \hat{I}\bigl(r_{n,k}\bigr),
& 1 < k < C,\\[6pt]
\frac{\partial}{\partial z_{n,C}} \hat{I}\bigl(r_{n,C-1}\bigr),
& k = C.
\end{cases}$}
\end{equation}
where each term on the right-hand side is given by the local pair contributions in Equation~\ref{eq:local_pair_contrib} (or its counterpart for \(z_{n,i+1}\)). Explicitly, Equation~\ref{eq:Lreg_two_terms} yields:
\begin{equation}
\label{eq:Lreg_boundary_left}
\frac{\partial \mathcal{L}_{\text{LBE}}(n)}{\partial z_{n,1}}
=
\begin{cases}
\hat{I}'(r_{n,1}), & y_n > 1,\\
-\hat{I}'(r_{n,1}), & y_n = 1,
\end{cases}
\end{equation}
\begin{equation}
\label{eq:Lreg_boundary_right}
\frac{\partial \mathcal{L}_{\text{LBE}}(n)}{\partial z_{n,C}}
=
\begin{cases}
-\hat{I}'(r_{n,C-1}), & y_n = C,\\
\hat{I}'(r_{n,C-1}), & y_n < C,
\end{cases}
\end{equation}
and, for interior indices \(1<k<C\),
\begin{equation}
\label{eq:Lreg_interior_two_terms}
\frac{\partial \mathcal{L}_{\text{LBE}}(n)}{\partial z_{n,k}}
=
\begin{cases}
\hat{I}'(r_{n,k}) - \hat{I}'(r_{n,k-1}), 
& k < y_n,\\[4pt]
-\hat{I}'(r_{n,k}) - \hat{I}'(r_{n,k-1}), 
& k = y_n,\\[4pt]
-\hat{I}'(r_{n,k}) + \hat{I}'(r_{n,k-1}), 
& k > y_n.
\end{cases}
\end{equation}
Equations~\ref{eq:Lreg_boundary_left}--\ref{eq:Lreg_interior_two_terms} make explicit that each logit’s gradient receives contributions from the two adjacent gaps \(r_{n,k-1}\) and \(r_{n,k}\), whereas the expressions in Table~I of the main text correspond to the contribution of a single pair.

Substituting the piecewise form of \(\hat{I}'(\cdot)\) into Equations~\ref{eq:Lreg_boundary_left}--\ref{eq:Lreg_interior_two_terms} recovers the log-barrier behavior described in the main text: within the feasible region \(r \le -1/s^2\), the gradients scale as \(-1/(s\,r)\), while in the linear near-boundary/violation branch \(r>-1/s^2\), they saturate at \(\pm s\), providing stable yet informative corrections.

\subsection{Combined Gradient}
\label{suppl_subsec:combined_grad} 
Combining the two components yields the gradient of the ORCU loss for sample~\(n\):
\begin{equation}
\label{eq:orcu_grad_general}
\frac{\partial \mathcal{L}_{\text{ORCU}}(n)}{\partial z_{n,k}}
\;=\;
\bigl(\hat{y}_{n,k} - y'_{n,k}\bigr)
\;+\;
\frac{\partial \mathcal{L}_{\text{LBE}}(n)}{\partial z_{n,k}}.
\end{equation}
Each logit \(z_{n,k}\) appears in at most two adjacent gaps, \(r_{n,k-1}\) and \(r_{n,k}\), defined in Equation~\ref{definition_of_r}. Using the chain rule with the derivative \(\hat{I}'(r)\) in Equation~\ref{eq:Ihat_prime}, the regularization gradient can be compactly written as
\begin{equation}
\label{eq:Lreg_two_pair_compact}
\resizebox{0.98\columnwidth}{!}{$\displaystyle
\frac{\partial \mathcal{L}_{\text{LBE}}(n)}{\partial z_{n,k}}
=
\begin{cases}
\frac{\partial}{\partial z_{n,1}} \hat{I}\bigl(r_{n,1}\bigr), & k = 1,\\[4pt]
\frac{\partial}{\partial z_{n,k}} \hat{I}\bigl(r_{n,k-1}\bigr)
+
\frac{\partial}{\partial z_{n,k}} \hat{I}\bigl(r_{n,k}\bigr), & 1 < k < C,\\[4pt]
\frac{\partial}{\partial z_{n,C}} \hat{I}\bigl(r_{n,C-1}\bigr), & k = C.
\end{cases}$}
\end{equation}
From Section~\ref{suppl_subsec:gradient_reg}, the local derivatives of a single pair \((i,i{+}1)\) satisfy
\begin{equation}
\label{eq:pair_local_derivatives}
\frac{\partial}{\partial z_{n,i}} \hat{I}\bigl(r_{n,i}\bigr)
=
\begin{cases}
\hat{I}'(r_{n,i}), & i < y_n,\\
-\hat{I}'(r_{n,i}), & i \ge y_n,
\end{cases}
\end{equation}
\begin{equation}
\frac{\partial}{\partial z_{n,i+1}} \hat{I}\bigl(r_{n,i}\bigr)
=
\begin{cases}
-\hat{I}'(r_{n,i}), & i < y_n,\\
\hat{I}'(r_{n,i}), & i \ge y_n.
\end{cases}
\end{equation}

\noindent\textbf{Boundary indices.}
For \(k=1\), only the pair \((1,2)\) contributes. Using Equation~\ref{eq:pair_local_derivatives},
\begin{equation}
\label{eq:Lreg_boundary_left_final}
\frac{\partial \mathcal{L}_{\text{LBE}}(n)}{\partial z_{n,1}}
=
\begin{cases}
\hat{I}'(r_{n,1}), & y_n > 1,\\[4pt]
-\hat{I}'(r_{n,1}), & y_n = 1.
\end{cases}
\end{equation}
Similarly, for \(k=C\), only the pair \((C{-}1,C)\) contributes and
\begin{equation}
\label{eq:Lreg_boundary_right_final}
\frac{\partial \mathcal{L}_{\text{LBE}}(n)}{\partial z_{n,C}}
=
\begin{cases}
-\hat{I}'(r_{n,C-1}), & y_n = C,\\[4pt]
\hat{I}'(r_{n,C-1}), & y_n < C.
\end{cases}
\end{equation}

\noindent\textbf{Interior indices and two-pair interaction.}
For interior logits \(1<k<C\), the logit \(z_{n,k}\) participates in both pairs \((k{-}1,k)\) and \((k,k{+}1)\).  
Let
\begin{equation}
r^{-}_{n,k} \;\coloneqq\; r_{n,k-1},
\qquad
r^{+}_{n,k} \;\coloneqq\; r_{n,k},
\end{equation}
denote the left and right gaps around \(z_{n,k}\).  
Applying Equation~\ref{eq:pair_local_derivatives} to the two neighboring pairs and summing their contributions, we obtain
\begin{equation}
\label{eq:Lreg_interior_two_pair}
\frac{\partial \mathcal{L}_{\text{LBE}}(n)}{\partial z_{n,k}}
=
\begin{cases}
\hat{I}'\bigl(r^{+}_{n,k}\bigr) - \hat{I}'\bigl(r^{-}_{n,k}\bigr), 
& 1<k<y_n,\\[4pt]
-\,\hat{I}'\bigl(r^{+}_{n,k}\bigr) - \hat{I}'\bigl(r^{-}_{n,k}\bigr), 
& k = y_n,\\[4pt]
-\,\hat{I}'\bigl(r^{+}_{n,k}\bigr) + \hat{I}'\bigl(r^{-}_{n,k}\bigr), 
& y_n<k<C.
\end{cases}
\end{equation}
Equations~\ref{eq:Lreg_boundary_left_final}, \ref{eq:Lreg_boundary_right_final}, and~\ref{eq:Lreg_interior_two_pair} together give the complete form of \(\tfrac{\partial \mathcal{L}_{\text{LBE}}(n)}{\partial z_{n,k}}\) that appears in Equation~\ref{eq:orcu_grad_general}. Substituting these into Equation~\ref{eq:orcu_grad_general} yields the full two-pair gradient of \(\mathcal{L}_{\text{ORCU}}\) with respect to each logit \(z_{n,k}\).

\noindent\textbf{Single-pair view (for Table~I of the main text).}
For completeness, we also record the single-pair contribution used to construct Table~I of the main text.  
If we focus only on the pair \((k,k{+}1)\), the regularizer contributes
\begin{equation}
\frac{\partial \mathcal{L}_{\text{LBE}}(n)}{\partial z_{n,k}}
=
\begin{cases}
\hat{I}'\bigl(z_{n,k} - z_{n,k+1}\bigr), & k < y_n,\\[4pt]
-\,\hat{I}'\bigl(z_{n,k+1} - z_{n,k}\bigr), & k \ge y_n,
\end{cases}
\end{equation}
so that, using Equation~\ref{eq:Ihat_prime},
\begin{equation}
\resizebox{\columnwidth}{!}{$
\begin{aligned}
\frac{\partial \mathcal{L}_{\text{ORCU}}(n)}{\partial z_{n,k}}
&=\;
\begin{cases}
\bigl(\hat{y}_{n,k} - y'_{n,k}\bigr) - \dfrac{1}{s\,\bigl(z_{n,k} - z_{n,k+1}\bigr)}, 
& \begin{array}{l}
k < y_n,\\[2pt]
z_{n,k} - z_{n,k+1} \le -\tfrac{1}{s^2}
\end{array}\\[8pt]
\bigl(\hat{y}_{n,k} - y'_{n,k}\bigr) + s, 
& \begin{array}{l}
k < y_n,\\[2pt]
z_{n,k} - z_{n,k+1} > -\tfrac{1}{s^2}
\end{array}\\[8pt]
\bigl(\hat{y}_{n,k} - y'_{n,k}\bigr) + \dfrac{1}{s\,\bigl(z_{n,k+1} - z_{n,k}\bigr)}, 
& \begin{array}{l}
k \ge y_n,\\[2pt]
z_{n,k+1} - z_{n,k} \le -\tfrac{1}{s^2}
\end{array}\\[8pt]
\bigl(\hat{y}_{n,k} - y'_{n,k}\bigr) - s, 
& \begin{array}{l}
k \ge y_n,\\[2pt]
z_{n,k+1} - z_{n,k} > -\tfrac{1}{s^2}
\end{array}
\end{cases}
\end{aligned}
$}
\label{eq:orcu_grad_single_pair}
\end{equation}
which matches the expressions summarized in Table~I of the main text for a single adjacent pair.  
The compact two-pair form in Equations~\ref{eq:Lreg_boundary_left_final}--\ref{eq:Lreg_interior_two_pair} and the single-pair behavior in Equation~\ref{eq:orcu_grad_single_pair} together characterize how \(\mathcal{L}_{\text{ORCU}}\) keeps predictions anchored to the soft-encoded ordinal target while using LBE to adjust directed adjacent logits for unimodal structure and feasible-interior confidence refinement.

\input{tables/suppl_tb2_r34}

\subsection{Summary}
\label{suppl_subsec:summary}
This gradient analysis shows how the ORCU loss combines a SoftCE probability anchor with an ordinal-aware LBE response. The directed gap $r$ serves as an input-specific ordinal uncertainty proxy: it identifies adjacent violations, encodes the pairwise ordering margin or near-boundary ambiguity through mass imbalance, and determines the strength of the feasible-interior update. Deep in the ordered region, LBE gives small but nonzero updates that preserve the current geometry; near the boundary, its magnitude increases and focuses correction on weak adjacent relations; after a violation, the linear continuation gives a bounded correction that restores the required unimodal ordering. These regimes explain how ORCU preserves unimodal structure while providing the confidence-refinement channel evaluated in the main paper.

\input{tables/suppl_tb3_r101}

\section{Additional Results}
\label{suppl_sec:results}

\subsection{Generalization Performance}
\label{suppl_subsec:generalization_performance}
This section presents extended results evaluating the generalizability of the proposed $\mathcal{L}_{\text{ORCU}}$ loss function across different model architectures. Tables~\ref{suppl_table:r34} and ~\ref{suppl_table:r101} assess calibration performance and the ability to capture ordinal label relationships on the Image Aesthetics, Adience, LIMUC, and Diabetic Retinopathy datasets using ResNet-34 and ResNet-101 architectures. Across both architectures, $\mathcal{L}_{\text{ORCU}}$ consistently demonstrates low calibration errors while maintaining alignment with ordinal label structures, highlighting its robustness in producing well-calibrated, ordinal-aware predictions. These findings further validate the effectiveness of $\mathcal{L}_{\text{ORCU}}$ for ordinal classification across diverse architectures.

\input{tables/suppl_tb4_full_classification}

\subsection{Classification Performance}
\label{suppl_subsec:classification_performance}
Table~\ref{suppl_table:classification} presents a comprehensive comparison between \(\mathcal{L}_{\text{ORCU}}\) and other loss functions, demonstrating its capacity to balance calibration and accuracy effectively. Across all metrics, \(\mathcal{L}_{\text{ORCU}}\) consistently outperforms the baseline CE and achieves competitive performance relative to other ordinal loss functions, substantially exceeding calibration-focused methods by leveraging ordinal structures. These findings affirm \(\mathcal{L}_{\text{ORCU}}\)’s efficacy in optimizing the calibration-accuracy trade-off through the integration of ordinal relationships within its calibration framework.

\subsection{Visual Analyses}
\label{suppl_subsec:visual_analyses}
\subsubsection{Reliability Diagrams}
\input{figures/tex/suppl_fig2_reliabilitydiagram}
\label{suppl_subsec:visual_analyses_rd}

Fig.~\ref{suppl_figure:reliability_diagram} presents a comparison of reliability diagrams for \(\mathcal{L}_{\text{ORCU}}\) against both ordinal-focused and calibration-focused loss functions. CE and ordinal-focused baselines (Fig.~\ref{suppl_figure:reliability_diagram}(a), (c)--(e)) consistently exhibit overconfidence, with predictions frequently exceeding the expected calibration line. SORD (Fig.~\ref{suppl_figure:reliability_diagram}(b)), designed to align predictions with a soft-encoded distribution rather than a one-hot target, captures ordinal structure through a dispersed label distribution. In contrast, the proposed \(\mathcal{L}_{\text{ORCU}}\) addresses these limitations (Fig.~\ref{suppl_figure:reliability_diagram}(k)) by mitigating SORD’s underconfidence and extending calibration to high-confidence predictions, resulting in more reliable confidence estimates.

Although calibration-focused methods aim to improve calibration, they often struggle in ordinal tasks due to their disregard for ordinal dependencies (Fig.~\ref{suppl_figure:reliability_diagram}(f)--(j)). Consequently, most of these methods show limited calibration improvements and frequently exhibit severe overconfidence, similar to the baseline CE. Among these, LS achieves relatively lower ECE, as shown in Fig.~\ref{suppl_figure:reliability_diagram}(f). However, its lack of ordinal awareness diminishes its effectiveness in ordinal tasks, a limitation further examined in the softmax output analysis (Fig.~\ref{suppl_figure:ls_comparison}). These findings emphasize the effectiveness of \(\mathcal{L}_{\text{ORCU}}\) in integrating ordinal relationships to achieve superior calibration in ordinal tasks.

\subsubsection{t-SNE Visualizations}
\label{suppl_subsec:visual_analyses_tsne}
\input{figures/tex/suppl_fig3_tsne}
\input{figures/tex/suppl_fig4,5_comparison}
Fig.~\ref{suppl_figure:tsne} presents a comparison of t-SNE visualizations~\cite{zhang2023improving} for models trained with \(\mathcal{L}_{\text{ORCU}}\) and comparative loss functions on the Adience, LIMUC, and Diabetic Retinopathy datasets. The Image Aesthetics dataset is excluded from this analysis due to its highly imbalanced class distribution, which complicates t-SNE-based embedding comparisons. Compared to other losses, \(\mathcal{L}_{\text{ORCU}}\) (Fig.~\ref{suppl_figure:tsne}(d)) produces embeddings more closely aligned with the ordinal label structure. The comparison with SORD (Fig.~\ref{suppl_figure:tsne}(b)) demonstrates the effect of the regularization term, which operates based on label order. These label-order-consistent embeddings suggest that models trained with \(\mathcal{L}_{\text{ORCU}}\) effectively leverage inherent ordinal structures in the prediction process. 

\subsubsection{Softmax Output Distribution}
\label{suppl_subsec:visual_analyses_softmax}

Fig.~\ref{suppl_figure:ce_comparison} compares the softmax output distributions of models trained with CE and \(\mathcal{L}_{\text{ORCU}}\) on the Adience dataset. For correct predictions (Fig.~\ref{suppl_figure:ce_comparison}(a)), models trained with CE produce overly confident outputs, regardless of input-specific uncertainty. In contrast, models trained with \(\mathcal{L}_{\text{ORCU}}\) exhibit calibrated predictions that adjust confidence levels based on input uncertainty, providing more conservative estimates for uncertain inputs. For incorrect predictions (Fig.~\ref{suppl_figure:ce_comparison}(b)), CE assigns disproportionately high confidence to a single incorrect label, neglecting ordinal relationships. Conversely, models trained with \(\mathcal{L}_{\text{ORCU}}\) distribute confidence across adjacent labels, maintaining ordinal consistency and adequately reflecting uncertainty. In both cases, CE's reliance on one-hot encoding results in poor alignment with inherent ordinal relationships. By coupling ordinal anchoring with adjacent-gap refinement during training, \(\mathcal{L}_{\text{ORCU}}\) facilitates consistent and reasonable predictions that align with the natural order of labels.

Fig.~\ref{suppl_figure:ls_comparison} compares the softmax output distributions of models trained with LS and \(\mathcal{L}_{\text{ORCU}}\) on the Adience dataset. LS shows relatively strong calibration performance as measured by the ECE metric, outperforming other calibration-focused losses. However, it lacks consistency in preserving ordinal relationships, which are essential for ordinal classification. For correct predictions (Fig.~\ref{suppl_figure:ls_comparison}(a)), models trained with LS and \(\mathcal{L}_{\text{ORCU}}\) provide similar confidence estimates for the target label. However, LS predictions violate unimodality, resulting in order-inconsistent outputs. For incorrect samples (Fig.~\ref{suppl_figure:ls_comparison}(b)), models trained with LS produce predictions misaligned with the requirements of ordinal tasks due to the absence of an order-awareness mechanism. In contrast, \(\mathcal{L}_{\text{ORCU}}\) maintains balanced confidence distributions across adjacent likely labels, preserving order-aware unimodality and reflecting input-specific uncertainty. These findings demonstrate \(\mathcal{L}_{\text{ORCU}}\)’s effectiveness in delivering reliable confidence estimates while capturing ordinal structure.

\input{tables/suppl_tb5_ablation}

\subsubsection{Grad-CAM Visualizations}
\label{suppl_subsec:visual_analyses_gradcam}
\input{figures/tex/suppl_fig6_gradcam}
\input{figures/tex/suppl_fig7_valcurve}
Fig.~\ref{suppl_figure:gradcam} presents Grad-CAM activation maps~\cite{jacobgilpytorchcam} for models trained with CE, SORD, LS, and ORCU on the Image Aesthetics and Adience datasets. These datasets are selected as they represent general computer vision tasks, enabling qualitative evaluation of feature localization. These visualizations, generated from correctly predicted samples, reveal notable differences in feature localization across the various loss functions. Models trained with ORCU consistently generate broader activation maps than those trained with other loss functions. Notably, compared to SORD, ORCU integrates an ordinal-aware regularization term that enhances calibration and ordinal consistency by leveraging the inherent relationships among labels, resulting in superior feature representation quality and improved interpretability~\cite{suara2023grad, tan2024interpret} in ordinal classification tasks.

\subsection{Training Curve Comparison}
\label{suppl_subsec:training_curve}
Fig.~\ref{suppl_figure:validation_curve} presents the convergence behavior of ORCU across calibration and classification metrics. ORCU achieves faster convergence in ECE, SCE, and ACE (Fig.~\ref{suppl_figure:validation_curve}(a)--(c)) compared to SORD, which does not incorporate the ordinal-aware regularization term. Notably, ORCU maintains stable QWK performance throughout training (Fig.~\ref{suppl_figure:validation_curve}(d)), indicating that the improved calibration achieves its effect without compromising ordinal classification performance.

\subsection{Comprehensive Ablation Study}
\label{suppl_subsec:ablation}
Table~\ref{suppl_table:ablation} presents a comprehensive ablation study conducted on the Adience, Image Aesthetics, and LIMUC datasets, analyzing the effects of various regularization terms and distance metrics in \(\mathcal{L_{\text{SoftCE}}}\). The results demonstrate that conventional calibration regularization terms are insufficient for achieving both effective calibration and unimodality, revealing their limitations in addressing the unique challenges of ordinal tasks. In contrast, combining the Squared distance metric with the proposed ordinal-aware regularization term consistently achieves robust calibration performance and ordinal alignment across datasets. This study provides a detailed evaluation of the key components essential for jointly learning calibration and ordinal structure in classification tasks.

%% file: tables/suppl_tb1_dataset.tex
\begin{table}[h!]
\renewcommand{\arraystretch}{1.00}
\caption{Label-wise distribution of images across all datasets, presented as counts and proportions (\%). This table enables the observation of datasets with balanced and imbalanced label distributions.}
\vspace{-0.2cm}
\centering
\resizebox{\columnwidth}{!}{%
        {
        \begin{tabular}{cccccc}
        \toprule 
        \multicolumn{1}{c}{\multirow{2}{*}{Label}} & \multicolumn{5}{c}{Datasets} \\ 
        \cmidrule(lr){2-6}
        & Image Aesthetics & Adience & LIMUC  & Diabetic Retinopathy & SST-5 \\
        \midrule
            0 & \makebox[2.5em][r]{36} \makebox[4em][r]{(0.27\%)} & \makebox[2.5em][r]{2,518} \makebox[4em][r]{(14.45\%)}  & \makebox[2.5em][r]{6,105} \makebox[4em][r]{(54.14\%)}  & \makebox[2.5em][r]{1,914} \makebox[4em][r]{(20.00\%)} & \makebox[2.5em][r]{1,510} \makebox[4em][r]{(12.74\%)} \\ 
            1 & \makebox[2.5em][r]{654} \makebox[4em][r]{(4.89\%)} & \makebox[2.5em][r]{2,163} \makebox[4em][r]{(12.41\%)} & \makebox[2.5em][r]{3,052} \makebox[4em][r]{(27.07\%)}  & \makebox[2.5em][r]{1,914} \makebox[4em][r]{(20.00\%)} & \makebox[2.5em][r]{3,140} \makebox[4em][r]{(26.49\%)} \\ 
            2 & \makebox[2.5em][r]{8,157} \makebox[4em][r]{(61.04\%)} & \makebox[2.5em][r]{2,132} \makebox[4em][r]{(12.24\%)} & \makebox[2.5em][r]{1,254} \makebox[4em][r]{(11.12\%)} & \makebox[2.5em][r]{1,914} \makebox[4em][r]{(20.00\%)}  & \makebox[2.5em][r]{2,242} \makebox[4em][r]{(18.91\%)}\\ 
            3 & \makebox[2.5em][r]{4,211} \makebox[4em][r]{(31.51\%)} & \makebox[2.5em][r]{1,655} \makebox[4em][r]{(9.50\%)} & \makebox[2.5em][r]{865} \makebox[4em][r]{(7.67\%)}  & \makebox[2.5em][r]{1,914} \makebox[4em][r]{(20.00\%)} & \makebox[2.5em][r]{3,111} \makebox[4em][r]{(26.24\%)}\\ 
            4 & \makebox[2.5em][r]{306} \makebox[4em][r]{(2.29\%)} & \makebox[2.5em][r]{4,950} \makebox[4em][r]{(28.41\%)} & - & \makebox[2.5em][r]{1,914} \makebox[4em][r]{(20.00\%)} & \makebox[2.5em][r]{1,852} \makebox[4em][r]{(15.62\%)}\\ 
            5 & - & \makebox[2.5em][r]{2,300} \makebox[4em][r]{(13.20\%)} & -  & - & -\\ 
            6 & - & \makebox[2.5em][r]{830} \makebox[4em][r]{(4.76\%)} & -  & - & - \\ 
            7 & - & \makebox[2.5em][r]{875} \makebox[4em][r]{(5.02\%)} & -  & - & - \\ \midrule
            Total & 13,364 & 17,423 & 11,276 & 9,570 & 11,855\\ 
        \bottomrule
        \end{tabular}
        }
    }
    \vspace{-0.2cm}
    \label{suppl_table:dataset}
\end{table}

%% file: figures/tex/suppl_fig1_tfigure.tex
\begin{figure*}[!t]
\centering
\subfloat[Diabetic Retinopathy]{%
    \includegraphics[width=0.47\textwidth]{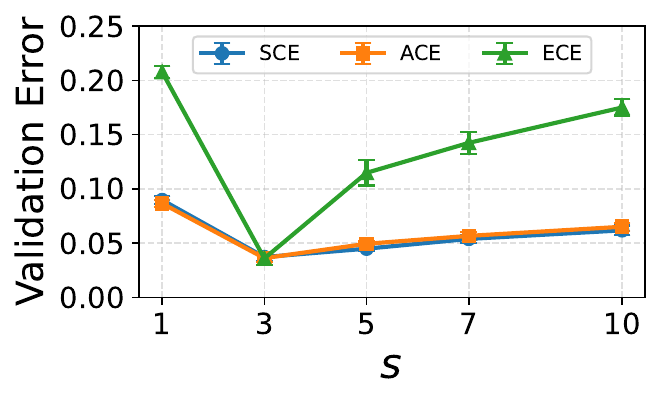}%
    \label{suppl_figure:t_dr}}
\hfill
\subfloat[Image Aesthetics]{%
    \includegraphics[width=0.47\textwidth]{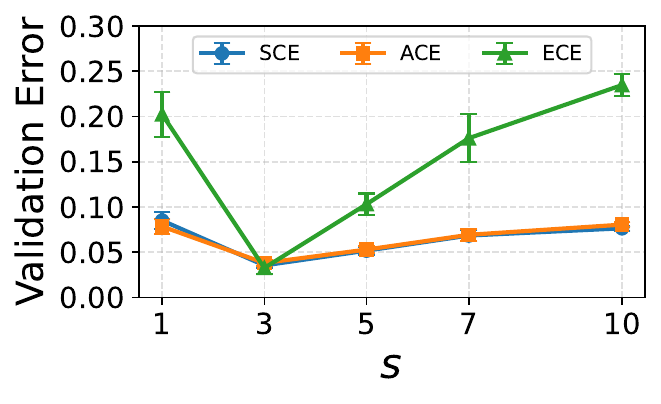}%
    \label{suppl_figure:t_ia}}
\caption{Effect of the hyperparameter $s$ in $\mathcal{L}_{\text{ORCU}}$ on SCE, ACE, and ECE. Although all three metrics are reported, $s$ was selected based on validation SCE. (a) Validation result on the balanced Diabetic Retinopathy dataset was used to determine $s$. (b) Validation on the imbalanced Image Aesthetics dataset, demonstrating the generalizability of $s$.}
\label{suppl_figure:t_figure}
\end{figure*}

%% file: tables/suppl_tb2_r34.tex
\begin{table*}[!t]
\renewcommand{\arraystretch}{1.0}
\caption{Calibration performance (SCE, ACE, and ECE) and ordinal structure representation (Unimodal) across various loss functions on four ordinal datasets using ResNet-34. The table shows the mean and standard deviation across folds, highlighting the best in bold and the second-best underlined.}
\centering
\resizebox{\textwidth}{!}{%
\begin{tabular}{rcccccccccccc}
\toprule
\multicolumn{1}{c}{\multirow{2}{*}{Metrics}} & \multicolumn{11}{c}{Methods} \\
\cmidrule(lr){2-12}
& CE
& SORD
& CDW-CE
& CO2
& POE
& LS
& FLSD
& MbLS
& MDCA
& ACLS 
& \cellcolor[gray]{.93} \textBF{ORCU (Ours)} \\
\midrule
\multicolumn{12}{c}{Image Aesthetics (\(n = 13{,}364\))} \\
SCE($10^{-2}$) $\downarrow$ & \phantom{0}8.10 $\pm$ 0.56 & \phantom{0}8.84 $\pm$ 0.11 & \phantom{0}\underline{3.80 $\pm$ 0.15} & \phantom{0}8.57 $\pm$ 0.92 & \phantom{0}6.17 $\pm$ 1.18 & \phantom{0}5.54 $\pm$ 0.75 & \phantom{0}6.06 $\pm$ 0.72 & \phantom{0}7.49 $\pm$ 0.46 & \phantom{0}7.53 $\pm$ 0.20 & \phantom{0}7.24 $\pm$ 0.95 & \cellcolor[gray]{.93} \phantom{0}\textBF{3.66 $\pm$ 0.86} \\
ACE($10^{-2}$) $\downarrow$ & \phantom{0}7.86 $\pm$ 0.53 & \phantom{0}8.82 $\pm$ 0.08 & \phantom{0}\underline{3.83 $\pm$ 0.14} & \phantom{0}8.30 $\pm$ 0.97 & \phantom{0}6.03 $\pm$ 1.18 & \phantom{0}5.43 $\pm$ 0.75 & \phantom{0}5.93 $\pm$ 0.65 & \phantom{0}7.32 $\pm$ 0.45 & \phantom{0}7.34 $\pm$ 0.19 & \phantom{0}7.11 $\pm$ 0.93 & \cellcolor[gray]{.93} \phantom{0}\textBF{3.81 $\pm$ 0.84} \\
ECE(\%) $\downarrow$ & 19.65 $\pm$ 1.54 & 18.46 $\pm$ 0.89 & \phantom{0}\underline{8.60 $\pm$ 0.59} & 20.89 $\pm$ 2.14 & 14.60 $\pm$ 2.87 & 10.10 $\pm$ 1.65 & 13.95 $\pm$ 2.10 & 18.00 $\pm$ 1.02 & 18.09 $\pm$ 0.52 & 17.42 $\pm$ 2.38 & \cellcolor[gray]{.93} \phantom{0}\textBF{6.08 $\pm$ 4.01} \\
Unimodal(\%) $\uparrow$ & 99.55 $\pm$ 0.54 & \textBF{100.0 $\pm$ 0.00} & 93.02 $\pm$ 8.91 & 99.44 $\pm$ 0.49 & 97.35 $\pm$ 2.24 & 82.82 $\pm$ 1.68 & \underline{99.80 $\pm$ 0.24} & 90.30 $\pm$ 0.99 & 99.89 $\pm$ 0.09 & 90.54 $\pm$ 3.50 & \cellcolor[gray]{.93} \textBF{100.0 $\pm$ 0.00} \\
\midrule
\multicolumn{12}{c}{Adience (\(n = 17{,}423\))} \\
SCE($10^{-2}$) $\downarrow$ & \phantom{0}8.76 $\pm$ 0.77 & \phantom{0}\underline{4.26 $\pm$ 0.83} & \phantom{0}8.09 $\pm$ 0.58 & \phantom{0}9.21 $\pm$ 1.02 & \phantom{0}8.41 $\pm$ 1.14 & \phantom{0}6.56 $\pm$ 0.56 & \phantom{0}8.33 $\pm$ 0.94 & \phantom{0}8.31 $\pm$ 0.85 & \phantom{0}8.85 $\pm$ 0.82 & \phantom{0}8.34 $\pm$ 0.95 & \cellcolor[gray]{.93} \phantom{0}\textBF{3.80 $\pm$ 0.22} \\
ACE($10^{-2}$) $\downarrow$ & \phantom{0}8.31 $\pm$ 0.83 & \phantom{0}\underline{4.35 $\pm$ 0.76} & \phantom{0}7.75 $\pm$ 0.55 & \phantom{0}8.69 $\pm$ 1.03 & \phantom{0}7.92 $\pm$ 1.17 & \phantom{0}6.33 $\pm$ 0.49 & \phantom{0}7.92 $\pm$ 0.93 & \phantom{0}7.90 $\pm$ 0.91 & \phantom{0}8.47 $\pm$ 0.88 & \phantom{0}7.91 $\pm$ 0.99 & \cellcolor[gray]{.93} \phantom{0}\textBF{3.65 $\pm$ 0.27} \\
ECE(\%) $\downarrow$ & 33.65 $\pm$ 3.09 & \phantom{0}\underline{7.04 $\pm$ 1.60} & 30.92 $\pm$ 2.48 & 35.63 $\pm$ 4.09 & 32.70 $\pm$ 4.65 & 24.37 $\pm$ 2.65 & 32.16 $\pm$ 3.98 & 32.20 $\pm$ 3.50 & 34.34 $\pm$ 3.15 & 32.22 $\pm$ 3.66 & \cellcolor[gray]{.93} \phantom{0}\textBF{6.97 $\pm$ 2.32} \\
Unimodal(\%) $\uparrow$ & 84.27 $\pm$ 2.71 & \underline{97.85 $\pm$ 0.77} & 88.43 $\pm$ 1.83 & 81.86 $\pm$ 1.65 & 72.04 $\pm$ 1.34 & 66.65 $\pm$ 8.32 & 85.03 $\pm$ 2.23 & 69.77 $\pm$ 1.09 & 83.14 $\pm$ 1.94 & 68.91 $\pm$ 1.35 & \cellcolor[gray]{.93} \textBF{99.98 $\pm$ 0.01} \\
\midrule
\multicolumn{12}{c}{LIMUC (\(n = 11{,}276\))} \\
SCE($10^{-2}$) $\downarrow$ & \phantom{0}4.11 $\pm$ 1.11 & \phantom{0}9.03 $\pm$ 0.35 & \phantom{0}4.65 $\pm$ 1.63 & \phantom{0}6.87 $\pm$ 0.77 & \phantom{0}6.27 $\pm$ 0.64 & \phantom{0}4.68 $\pm$ 0.48 & \phantom{0}\textBF{2.90 $\pm$ 0.61} & \phantom{0}6.36 $\pm$ 0.88 & \phantom{0}\underline{3.84 $\pm$ 0.85} & \phantom{0}6.29 $\pm$ 0.80 & \cellcolor[gray]{.93} \phantom{0}5.05 $\pm$ 1.02 \\
ACE($10^{-2}$) $\downarrow$ & \phantom{0}3.69 $\pm$ 1.15 & \phantom{0}8.94 $\pm$ 0.32 & \phantom{0}4.34 $\pm$ 1.65 & \phantom{0}6.50 $\pm$ 0.74 & \phantom{0}6.02 $\pm$ 0.73 & \phantom{0}4.74 $\pm$ 0.62 & \phantom{0}\textBF{2.45 $\pm$ 0.58} & \phantom{0}6.02 $\pm$ 0.84 & \phantom{0}\underline{3.45 $\pm$ 0.82} & \phantom{0}5.98 $\pm$ 0.90 & \cellcolor[gray]{.93} \phantom{0}4.99 $\pm$ 1.03 \\
ECE(\%) $\downarrow$ & \phantom{0}7.24 $\pm$ 2.34 & 16.85 $\pm$ 1.17 & \phantom{0}7.50 $\pm$ 3.19 & 13.00 $\pm$ 1.66 & 11.80 $\pm$ 1.35 & \phantom{0}\underline{6.06 $\pm$ 1.46} & \phantom{0}\textBF{4.27 $\pm$ 1.33} & 12.11 $\pm$ 1.84 & \phantom{0}6.67 $\pm$ 1.92 & 12.09 $\pm$ 1.78 & \cellcolor[gray]{.93} \phantom{0}9.46 $\pm$ 2.74 \\
Unimodal(\%) $\uparrow$ & 99.65 $\pm$ 0.26 & \textBF{100.0 $\pm$ 0.00} & 99.85 $\pm$ 0.37 & 99.68 $\pm$ 0.32 & 93.01 $\pm$ 1.29 & 81.12 $\pm$ 2.07 & \underline{99.85 $\pm$ 0.10} & 86.84 $\pm$ 2.00 & 99.80 $\pm$ 0.12 & 87.57 $\pm$ 2.26 & \cellcolor[gray]{.93} \textBF{100.0 $\pm$ 0.00} \\
\midrule
\multicolumn{12}{c}{Diabetic Retinopathy (\(n = 9{,}570\))} \\
SCE($10^{-2}$) $\downarrow$ & \phantom{0}8.73 $\pm$ 1.19 & \phantom{0}\underline{3.99 $\pm$ 0.41} & 11.88 $\pm$ 1.58 & 11.46 $\pm$ 0.47 & \phantom{0}7.18 $\pm$ 0.96 & \phantom{0}7.25 $\pm$ 0.50 & \phantom{0}5.26 $\pm$ 0.49 & \phantom{0}8.87 $\pm$ 0.67 & \phantom{0}8.17 $\pm$ 0.79 & \phantom{0}9.09 $\pm$ 0.70 & \cellcolor[gray]{.93} \phantom{0}\textBF{3.76 $\pm$ 0.40} \\
ACE($10^{-2}$) $\downarrow$ & \phantom{0}8.48 $\pm$ 1.16 & \phantom{0}\underline{4.10 $\pm$ 0.38} & 11.72 $\pm$ 1.66 & 11.24 $\pm$ 0.48 & \phantom{0}7.00 $\pm$ 0.93 & \phantom{0}7.10 $\pm$ 0.50 & \phantom{0}5.19 $\pm$ 0.45 & \phantom{0}8.63 $\pm$ 0.60 & \phantom{0}7.95 $\pm$ 0.76 & \phantom{0}8.86 $\pm$ 0.69 & \cellcolor[gray]{.93} \phantom{0}\textBF{3.89 $\pm$ 0.38} \\
ECE(\%) $\downarrow$ & 20.56 $\pm$ 2.78 & \phantom{0}\underline{5.38 $\pm$ 0.38} & 24.74 $\pm$ 2.56 & 27.61 $\pm$ 1.52 & 16.99 $\pm$ 2.22 & 17.04 $\pm$ 1.15 & 12.00 $\pm$ 1.31 & 20.80 $\pm$ 1.86 & 19.27 $\pm$ 2.11 & 21.45 $\pm$ 1.60 & \cellcolor[gray]{.93} \phantom{0}\textBF{4.10 $\pm$ 0.43} \\
Unimodal(\%) $\uparrow$ & 92.87 $\pm$ 0.95 & \underline{99.92 $\pm$ 0.05} & 99.44 $\pm$ 0.69 & 88.86 $\pm$ 1.99 & 89.18 $\pm$ 1.36 & 81.36 $\pm$ 0.95 & 94.52 $\pm$ 1.18 & 87.21 $\pm$ 0.80 & 92.92 $\pm$ 0.94 & 87.78 $\pm$ 1.32 & \cellcolor[gray]{.93} \textBF{99.98 $\pm$ 0.01} \\
\bottomrule
\end{tabular}%
}
\label{suppl_table:r34}
\end{table*}

%% file: tables/suppl_tb3_r101.tex
\begin{table*}[!t]
\renewcommand{\arraystretch}{1.0}
\caption{Calibration performance (SCE, ACE, and ECE) and ordinal structure representation (Unimodal) across various loss functions on four ordinal datasets using ResNet-101. The table shows the mean and standard deviation across folds, highlighting the best in bold and the second-best underlined.}
\centering
\resizebox{\textwidth}{!}{%
\begin{tabular}{rcccccccccccc}
\toprule
\multicolumn{1}{c}{\multirow{2}{*}{Metrics}} & \multicolumn{11}{c}{Methods} \\
\cmidrule(lr){2-12}
& CE
& SORD 
& CDW-CE
& CO2
& POE
& LS
& FLSD
& MbLS
& MDCA 
& ACLS
& \cellcolor[gray]{.93} \textBF{ORCU (Ours)} \\
\midrule
\multicolumn{12}{c}{Image Aesthetics (\(n = 13{,}364\))} \\
SCE($10^{-2}$) $\downarrow$ & \phantom{0}8.33 $\pm$ 0.85 & \phantom{0}8.59 $\pm$ 0.25 & \phantom{0}6.78 $\pm$ 0.63 & \phantom{0}9.22 $\pm$ 0.32 & \phantom{0}7.93 $\pm$ 0.44 & \phantom{0}\underline{6.12 $\pm$ 0.26} & \phantom{0}7.32 $\pm$ 0.87 & \phantom{0}7.82 $\pm$ 0.29 & \phantom{0}8.48 $\pm$ 0.22 & \phantom{0}8.03 $\pm$ 0.29 & \cellcolor[gray]{.93} \phantom{0}\textBF{2.82 $\pm$ 0.22} \\
ACE($10^{-2}$) $\downarrow$ & \phantom{0}8.00 $\pm$ 0.84 & \phantom{0}8.50 $\pm$ 0.27 & \phantom{0}6.60 $\pm$ 0.61 & \phantom{0}8.85 $\pm$ 0.33 & \phantom{0}7.72 $\pm$ 0.39 & \phantom{0}\underline{6.15 $\pm$ 0.32} & \phantom{0}6.95 $\pm$ 0.78 & \phantom{0}7.66 $\pm$ 0.32 & \phantom{0}8.19 $\pm$ 0.18 & \phantom{0}7.82 $\pm$ 0.35 & \cellcolor[gray]{.93} \phantom{0}\textBF{2.96 $\pm$ 0.16} \\
ECE(\%) $\downarrow$ & 20.49 $\pm$ 2.20 & 18.13 $\pm$ 1.01 & 16.42 $\pm$ 1.61 & 22.58 $\pm$ 0.81 & 19.20 $\pm$ 1.12 & \underline{11.35 $\pm$ 0.57} & 17.76 $\pm$ 2.16 & 18.93 $\pm$ 0.79 & 20.55 $\pm$ 0.56 & 19.43 $\pm$ 0.86 & \cellcolor[gray]{.93} \phantom{0}\textBF{2.20 $\pm$ 0.31} \\
Unimodal(\%) $\uparrow$ & 95.63 $\pm$ 1.38 & \textBF{100.0 $\pm$ 0.00} & \underline{98.47 $\pm$ 1.45} & 94.29 $\pm$ 0.73 & 91.23 $\pm$ 1.59 & 78.68 $\pm$ 0.18 & 97.64 $\pm$ 0.47 & 85.40 $\pm$ 1.06 & 96.01 $\pm$ 0.86 & 84.64 $\pm$ 0.65 & \cellcolor[gray]{.93} \textBF{100.0 $\pm$ 0.00} \\
\midrule
\multicolumn{12}{c}{Adience (\(n = 17{,}423\))} \\
SCE($10^{-2}$) $\downarrow$ & \phantom{0}8.78 $\pm$ 1.20 & \phantom{0}\underline{4.43 $\pm$ 0.63} & \phantom{0}8.07 $\pm$ 0.83 & \phantom{0}9.42 $\pm$ 1.02 & \phantom{0}7.42 $\pm$ 0.66 & \phantom{0}5.84 $\pm$ 0.88 & \phantom{0}8.46 $\pm$ 1.05 & \phantom{0}7.70 $\pm$ 1.01 & \phantom{0}8.84 $\pm$ 0.88 & \phantom{0}7.46 $\pm$ 0.87 & \cellcolor[gray]{.93} \phantom{0}\textBF{4.10 $\pm$ 0.21} \\
ACE($10^{-2}$) $\downarrow$ & \phantom{0}8.33 $\pm$ 1.20 & \phantom{0}\underline{4.45 $\pm$ 0.60} & \phantom{0}7.81 $\pm$ 0.79 & \phantom{0}8.91 $\pm$ 1.07 & \phantom{0}7.06 $\pm$ 0.67 & \phantom{0}5.76 $\pm$ 0.88 & \phantom{0}8.01 $\pm$ 1.01 & \phantom{0}7.33 $\pm$ 0.96 & \phantom{0}8.41 $\pm$ 0.85 & \phantom{0}7.09 $\pm$ 0.90 & \cellcolor[gray]{.93} \phantom{0}\textBF{3.96 $\pm$ 0.27} \\
ECE(\%) $\downarrow$ & 34.06 $\pm$ 4.86 & \phantom{0}\textBF{7.52 $\pm$ 2.95} & 30.53 $\pm$ 3.53 & 36.47 $\pm$ 4.01 & 28.22 $\pm$ 2.86 & 20.50 $\pm$ 4.13 & 32.56 $\pm$ 0.63 & 29.68 $\pm$ 4.18 & 34.17 $\pm$ 3.63 & 28.65 $\pm$ 3.45 & \cellcolor[gray]{.93} \phantom{0}\underline{8.96 $\pm$ 2.32} \\
Unimodal(\%) $\uparrow$ & 84.17 $\pm$ 1.51 & \underline{98.15 $\pm$ 0.82} & 91.75 $\pm$ 2.04 & 83.04 $\pm$ 1.34 & 70.59 $\pm$ 1.03 & 67.47 $\pm$ 0.76 & 85.50 $\pm$ 0.98 & 69.73 $\pm$ 0.93 & 83.99 $\pm$ 1.84 & 68.79 $\pm$ 0.81 & \cellcolor[gray]{.93} \textBF{99.92 $\pm$ 0.04} \\
\midrule
\multicolumn{12}{c}{LIMUC (\(n = 11{,}276\))} \\
SCE($10^{-2}$) $\downarrow$ & \phantom{0}6.77 $\pm$ 0.61 & \phantom{0}8.60 $\pm$ 0.34 & \phantom{0}6.15 $\pm$ 0.43 & \phantom{0}7.81 $\pm$ 0.43 & \phantom{0}7.01 $\pm$ 0.45 & \phantom{0}\underline{4.56 $\pm$ 0.21} & \phantom{0}6.39 $\pm$ 0.40 & \phantom{0}6.83 $\pm$ 0.50 & \phantom{0}6.60 $\pm$ 0.58 & \phantom{0}7.06 $\pm$ 0.60 & \cellcolor[gray]{.93} \phantom{0}\textBF{4.16 $\pm$ 0.78} \\
ACE($10^{-2}$) $\downarrow$ & \phantom{0}6.40 $\pm$ 0.58 & 8.39 $\pm$ 0.33 & \phantom{0}5.80 $\pm$ 0.49 & \phantom{0}7.40 $\pm$ 0.41 & \phantom{0}6.69 $\pm$ 0.48 & \phantom{0}\underline{4.47 $\pm$ 0.36} & \phantom{0}6.00 $\pm$ 0.40 & \phantom{0}6.48 $\pm$ 0.48 &\phantom{0}6.23 $\pm$ 0.62 & \phantom{0}6.63 $\pm$ 0.63 & \cellcolor[gray]{.93} \phantom{0}\textBF{3.82 $\pm$ 0.90} \\
ECE(\%) $\downarrow$ & 13.13 $\pm$ 1.22 & 16.14 $\pm$ 0.78 & 11.76 $\pm$ 0.94 & 15.17 $\pm$ 0.75 & 13.60 $\pm$ 1.04 & \phantom{0}\textBF{5.84 $\pm$ 0.82} & 12.30 $\pm$ 0.80 & 13.31 $\pm$ 1.06 & 12.72 $\pm$ 1.38 & 13.51 $\pm$ 1.19 & \cellcolor[gray]{.93} \phantom{0}\underline{7.17 $\pm$ 2.09} \\
Unimodal(\%) $\uparrow$ & 97.92 $\pm$ 0.98 & \textBF{100.0 $\pm$ 0.00} & \underline{99.85 $\pm$ 0.28} & 97.54 $\pm$ 1.52 & 89.62 $\pm$ 1.68 & 76.83 $\pm$ 0.90 & 98.36 $\pm$ 1.60 & 82.97 $\pm$ 0.92 & 98.40 $\pm$ 1.28 & 82.91 $\pm$ 1.80 & \cellcolor[gray]{.93} \textBF{100.0 $\pm$ 0.00} \\
\midrule
\multicolumn{12}{c}{Diabetic Retinopathy (\(n = 9{,}570\))} \\
SCE($10^{-2}$) $\downarrow$ & 10.36 $\pm$ 0.81 & \phantom{0}\underline{4.40 $\pm$ 0.34} & \phantom{0}9.15 $\pm$ 0.58 & 11.76 $\pm$ 0.80 & \phantom{0}8.90 $\pm$ 0.26 & \phantom{0}6.88 $\pm$ 0.46 & \phantom{0}7.43 $\pm$ 0.77 & \phantom{0}9.51 $\pm$ 0.43 & \phantom{0}9.29 $\pm$ 0.66 & \phantom{0}9.94 $\pm$ 0.65 & \cellcolor[gray]{.93} \phantom{0}\textBF{3.78 $\pm$ 0.30} \\
ACE($10^{-2}$) $\downarrow$ & 10.15 $\pm$ 0.79 & \phantom{0}\underline{4.44 $\pm$ 0.29} & \phantom{0}8.86 $\pm$ 0.52 & 11.53 $\pm$ 0.73 & \phantom{0}8.62 $\pm$ 0.30 & \phantom{0}6.83 $\pm$ 0.45 & \phantom{0}7.23 $\pm$ 0.75 & \phantom{0}9.32 $\pm$ 0.44 & \phantom{0}9.11 $\pm$ 0.77 & \phantom{0}9.72 $\pm$ 0.58 & \cellcolor[gray]{.93} \phantom{0}\textBF{3.74 $\pm$ 0.23} \\
ECE(\%) $\downarrow$ & 25.05 $\pm$ 2.13 & \phantom{0}\textBF{3.54 $\pm$ 1.00} & 21.32 $\pm$ 1.98 & 28.39 $\pm$ 1.90 & 21.15 $\pm$ 0.89 & 16.49 $\pm$ 1.09 & 17.52 $\pm$ 2.04 & 22.90 $\pm$ 1.19 & 21.87 $\pm$ 1.46 & 23.79 $\pm$ 1.52 & \cellcolor[gray]{.93} \phantom{0}\underline{4.13 $\pm$ 0.39} \\
Unimodal(\%) $\uparrow$ & 90.75 $\pm$ 1.35 & 99.88 $\pm$ 0.04 & \underline{99.97 $\pm$ 0.05} & 88.03 $\pm$ 1.99 & 91.03 $\pm$ 2.43 & 78.37 $\pm$ 0.73 & 93.30 $\pm$ 1.11 & 83.94 $\pm$ 0.96 & 92.11 $\pm$ 1.36 & 84.34 $\pm$ 1.52 & \cellcolor[gray]{.93} \textBF{99.99 $\pm$ 0.02} \\
\bottomrule
\end{tabular}%
}
\label{suppl_table:r101}
\end{table*}

%% file: tables/suppl_tb4_full_classification.tex
\begin{table*}[!t]
\renewcommand{\arraystretch}{1.0}
\caption{Classification results across all datasets using ResNet-50. Each value denotes the mean and standard deviation across folds. Best and second-best results are shown in bold and underlined, respectively.}
\centering
\resizebox{\textwidth}{!}{%
\begin{tabular}{rcccccccccccc}
\toprule
\multicolumn{1}{c}{\multirow{2}{*}{Metrics}} & \multicolumn{11}{c}{Methods} \\
\cmidrule(lr){2-12}
& CE
& SORD
& CDW-CE
& CO2
& POE 
& LS
& FLSD
& MbLS
& MDCA
& ACLS
& \cellcolor[gray]{.93} \textBF{ORCU (Ours)} \\
\midrule
\multicolumn{12}{c}{Image Aesthetics (\(n = 13{,}364\))} \\
Acc(\%) $\uparrow$
& 70.30 $\pm$ 0.80 & \textBF{70.92 $\pm$ 0.44} & 70.41 $\pm$ 0.64 & 69.80 $\pm$ 0.72 & 70.15 $\pm$ 0.58 & 70.63 $\pm$ 0.54 & 70.15 $\pm$ 0.59 & 70.40 $\pm$ 0.57 & 70.30 $\pm$ 0.57 & 70.31 $\pm$ 0.75 & \cellcolor[gray]{.93} \underline{70.74 $\pm$ 0.64} \\
QWK($10^{-2}$) $\uparrow$
& 49.61 $\pm$ 1.99 & \underline{50.30 $\pm$ 0.30} & 48.37 $\pm$ 1.48 & 49.19 $\pm$ 1.58 & 48.54 $\pm$ 1.50 & 49.90 $\pm$ 1.22 & 49.92 $\pm$ 1.45 & 49.42 $\pm$ 1.59 & 49.93 $\pm$ 1.29 & 49.53 $\pm$ 1.47 & \cellcolor[gray]{.93} \textBF{51.52 $\pm$ 1.13} \\
MAE($10^{-2}$) $\downarrow$ 
& 31.06 $\pm$ 0.91 & \textBF{30.02 $\pm$ 0.31} & 30.62 $\pm$ 0.63 & 31.55 $\pm$ 0.66 & 31.48 $\pm$ 0.75 & 30.63 $\pm$ 0.11 & 31.11 $\pm$ 0.57 & 31.02 $\pm$ 0.63 & 31.10 $\pm$ 0.53 & 31.02 $\pm$ 0.71 & \cellcolor[gray]{.93} \underline{30.27 $\pm$ 0.53} \\
\midrule
\multicolumn{12}{c}{Adience (\(n = 17{,}423\))} \\
Acc(\%) $\uparrow$
& 56.39 $\pm$ 4.86 & \textBF{59.10 $\pm$ 4.39} & \underline{57.89 $\pm$ 3.62} & 56.37 $\pm$ 4.89 & 57.29 $\pm$ 4.47 & 57.92 $\pm$ 4.95 & 57.18 $\pm$ 5.08 & 57.94 $\pm$ 4.67 & 56.18 $\pm$ 5.27 & 57.62 $\pm$ 4.88 & \cellcolor[gray]{.93} 57.50 $\pm$ 4.27 \\
QWK($10^{-2}$) $\uparrow$ 
& 87.95 $\pm$ 3.45 & \textBF{89.95 $\pm$ 2.91} & \underline{89.88 $\pm$ 2.57} & 87.28 $\pm$ 0.73 & 87.55 $\pm$ 3.04 & 88.24 $\pm$ 3.71 & 88.04 $\pm$ 3.44 & 88.25 $\pm$ 3.24 & 87.78 $\pm$ 3.64 & 87.88 $\pm$ 3.66 & \cellcolor[gray]{.93} 89.74 $\pm$ 2.99 \\
MAE($10^{-2}$) $\downarrow$ 
& 55.04 $\pm$ 7.04 & \textBF{48.75 $\pm$ 5.85} & \underline{49.48 $\pm$ 4.86} & 56.04 $\pm$ 7.65 & 54.73 $\pm$ 6.03 & 53.00 $\pm$ 7.39 & 54.15 $\pm$ 7.26 & 52.91 $\pm$ 6.39 & 55.38 $\pm$ 7.40 & 53.83 $\pm$ 7.04 & \cellcolor[gray]{.93} 50.71 $\pm$ 6.03 \\
\midrule
\multicolumn{12}{c}{LIMUC (\(n = 11{,}276\))} \\
Acc(\%) $\uparrow$
& 77.02 $\pm$ 0.66 & 77.49 $\pm$ 0.60 & \underline{77.73 $\pm$ 0.58} & 76.62 $\pm$ 0.76 & 76.60 $\pm$ 0.74 & 76.33 $\pm$ 0.55 & 77.40 $\pm$ 0.55 & 76.55 $\pm$ 0.79 & 76.92 $\pm$ 0.48 & 76.83 $\pm$ 0.89 & \cellcolor[gray]{.93} \textBF{78.38 $\pm$ 0.73} \\
QWK($10^{-2}$) $\uparrow$ 
& 84.61 $\pm$ 0.90 & 85.39 $\pm$ 0.62 & \underline{85.51 $\pm$ 0.69} & 84.11 $\pm$ 0.81 & 83.92 $\pm$ 1.11 & 84.02 $\pm$ 0.86 & 84.93 $\pm$ 0.77 & 84.09 $\pm$ 0.81 & 84.13 $\pm$ 0.60 & 84.54 $\pm$ 0.92 & \cellcolor[gray]{.93} \textBF{86.21 $\pm$ 0.67} \\
MAE($10^{-2}$) $\downarrow$ 
& 23.87 $\pm$ 0.83 & 23.14 $\pm$ 0.63 & \underline{22.88 $\pm$ 0.67} & 24.32 $\pm$ 0.86 & 24.49 $\pm$ 1.00 & 24.67 $\pm$ 0.67 & 23.47 $\pm$ 0.72 & 24.49 $\pm$ 0.87 & 24.15 $\pm$ 0.49 & 24.07 $\pm$ 0.99 & \cellcolor[gray]{.93} \textBF{22.10 $\pm$ 0.82} \\
\midrule
\multicolumn{12}{c}{Diabetic Retinopathy (\(n = 9{,}570\))} \\
Acc(\%) $\uparrow$ 
& 54.11 $\pm$ 1.15 & 53.07 $\pm$ 0.86 & 53.47 $\pm$ 1.14 & 53.90 $\pm$ 0.65 & 54.46 $\pm$ 0.61 & 54.66 $\pm$ 1.26 & 54.58 $\pm$ 0.90 & \underline{54.73 $\pm$ 0.80} & \textBF{54.99 $\pm$ 0.83} & 54.10 $\pm$ 0.48 & \cellcolor[gray]{.93} 54.25 $\pm$ 1.39 \\
QWK($10^{-2}$) $\uparrow$
& 76.63 $\pm$ 1.53 & \underline{79.20 $\pm$ 0.92} & \textBF{79.68 $\pm$ 0.81} & 76.78 $\pm$ 1.04 & 76.78 $\pm$ 1.04 & 76.55 $\pm$ 1.72 & 76.55 $\pm$ 1.72 & 77.00 $\pm$ 1.41 & 77.39 $\pm$ 1.42 & 77.08 $\pm$ 1.41 & \cellcolor[gray]{.93} 78.86 $\pm$ 0.88 \\
MAE($10^{-2}$) $\downarrow$ 
& 59.40 $\pm$ 2.23 & 56.77 $\pm$ 1.32 & \textBF{55.62 $\pm$ 1.44} & 58.81 $\pm$ 1.15 & 58.73 $\pm$ 0.95 & 58.97 $\pm$ 2.36 & 57.96 $\pm$ 1.16 & 58.48 $\pm$ 1.93 & 58.00 $\pm$ 1.63 & 58.93 $\pm$ 1.42 & \cellcolor[gray]{.93} \underline{56.23 $\pm$ 1.75} \\
\bottomrule
\end{tabular}%
}
\label{suppl_table:classification}
\end{table*}

%% file: figures/tex/suppl_fig2_reliabilitydiagram.tex
\begin{figure*}[!t]
\begin{center}
\includegraphics[width=\textwidth]{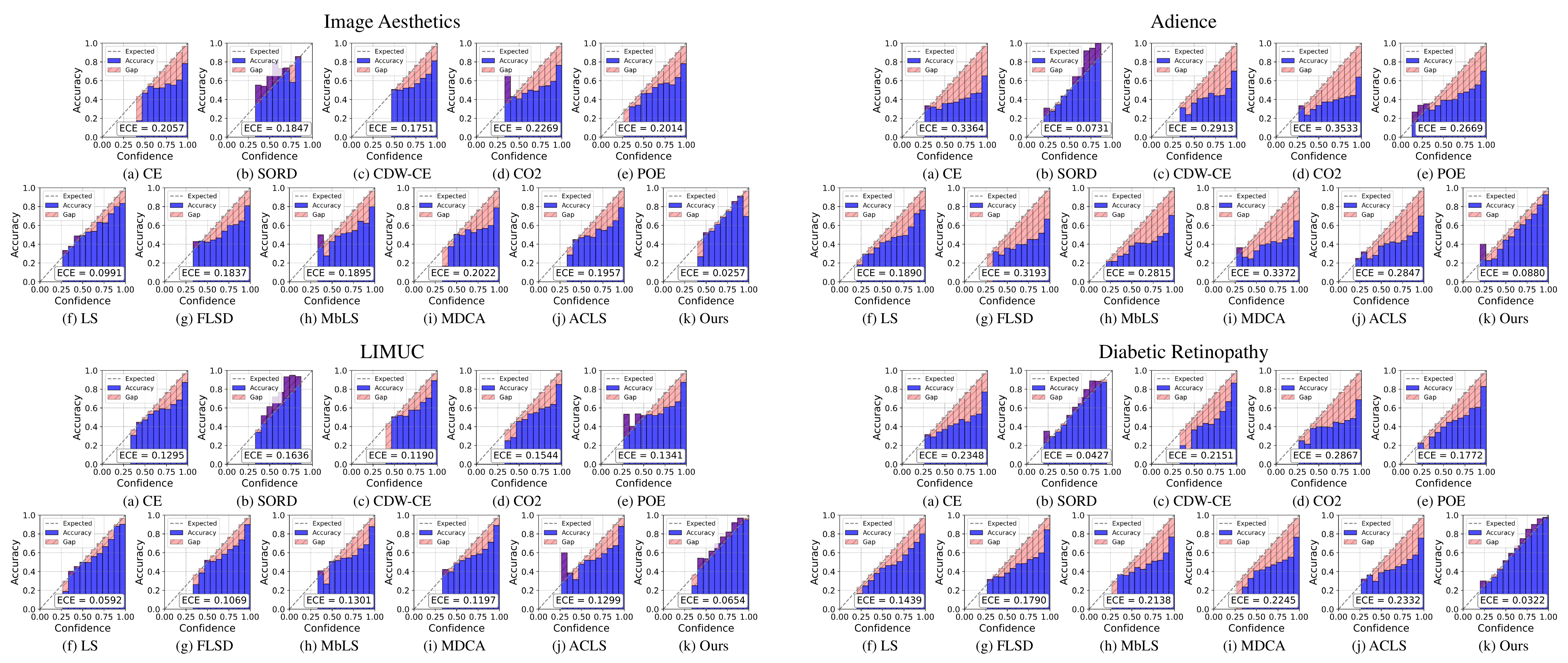}
\end{center}
\caption{Reliability diagrams comparing ordinal loss functions on the test splits of Image Aesthetics, Adience, LIMUC, and Diabetic Retinopathy. The diagrams show the calibration gap between model confidence and accuracy. Bars above the expected line indicate underconfidence (\( P(Y = y \mid \hat{P} = p) > p \)), while those below indicate overconfidence (\( P(Y = y \mid \hat{P} = p) < p \)). ECE is computed using 15 bins.}
\label{suppl_figure:reliability_diagram}
\end{figure*}

%% file: figures/tex/suppl_fig3_tsne.tex
\begin{figure*}[!t]
\centering
\includegraphics[width=0.72\textwidth]{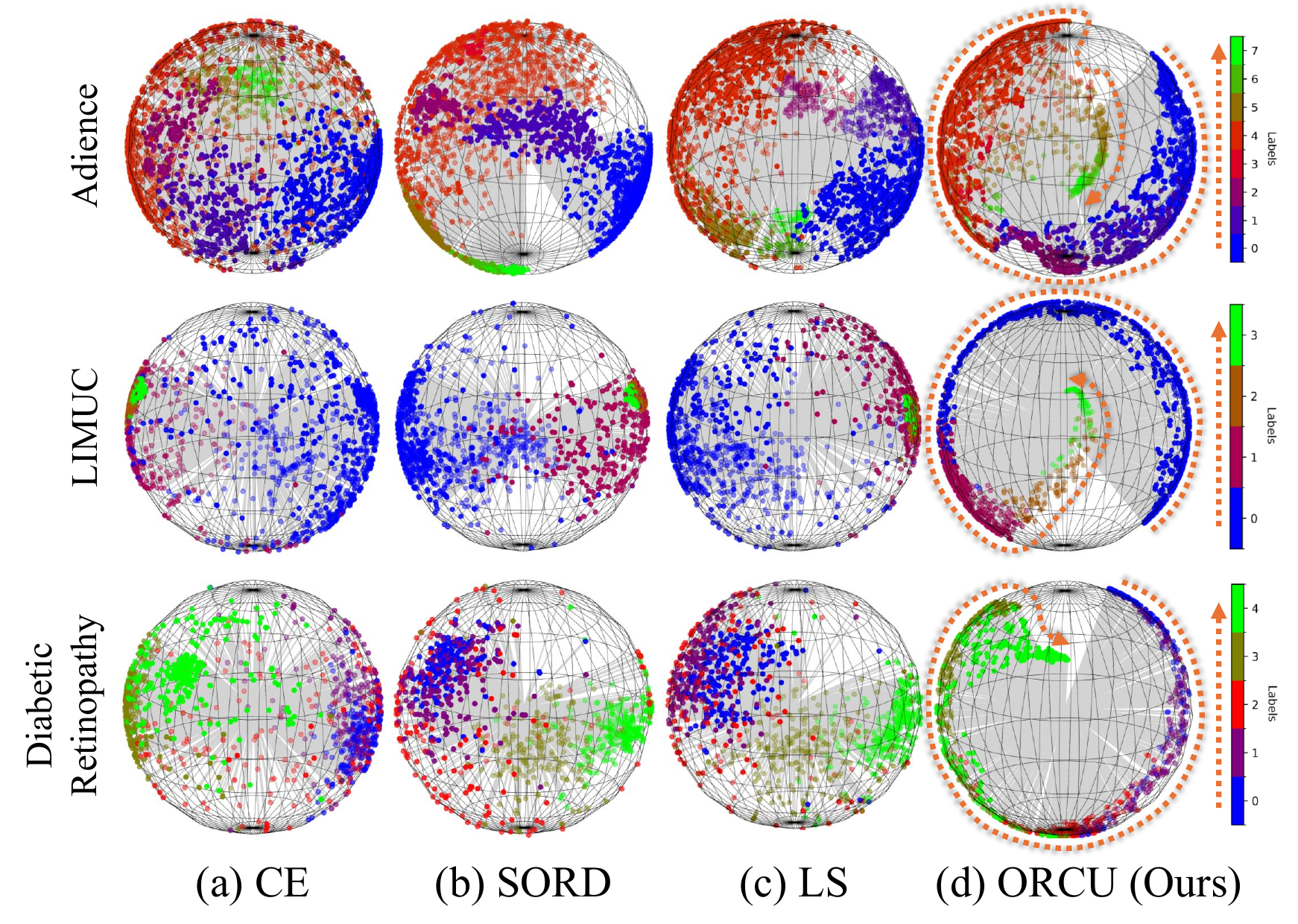}
\caption{t-SNE plots of ResNet-50 embeddings on Adience, LIMUC, and Diabetic Retinopathy. Colors denote class order. The orange dashed line indicates ordinal alignment.}
\label{suppl_figure:tsne}
\end{figure*}

%% file: figures/tex/suppl_fig4,5_comparison.tex
\begin{figure}[!t]
\centering
\includegraphics[width=\columnwidth]{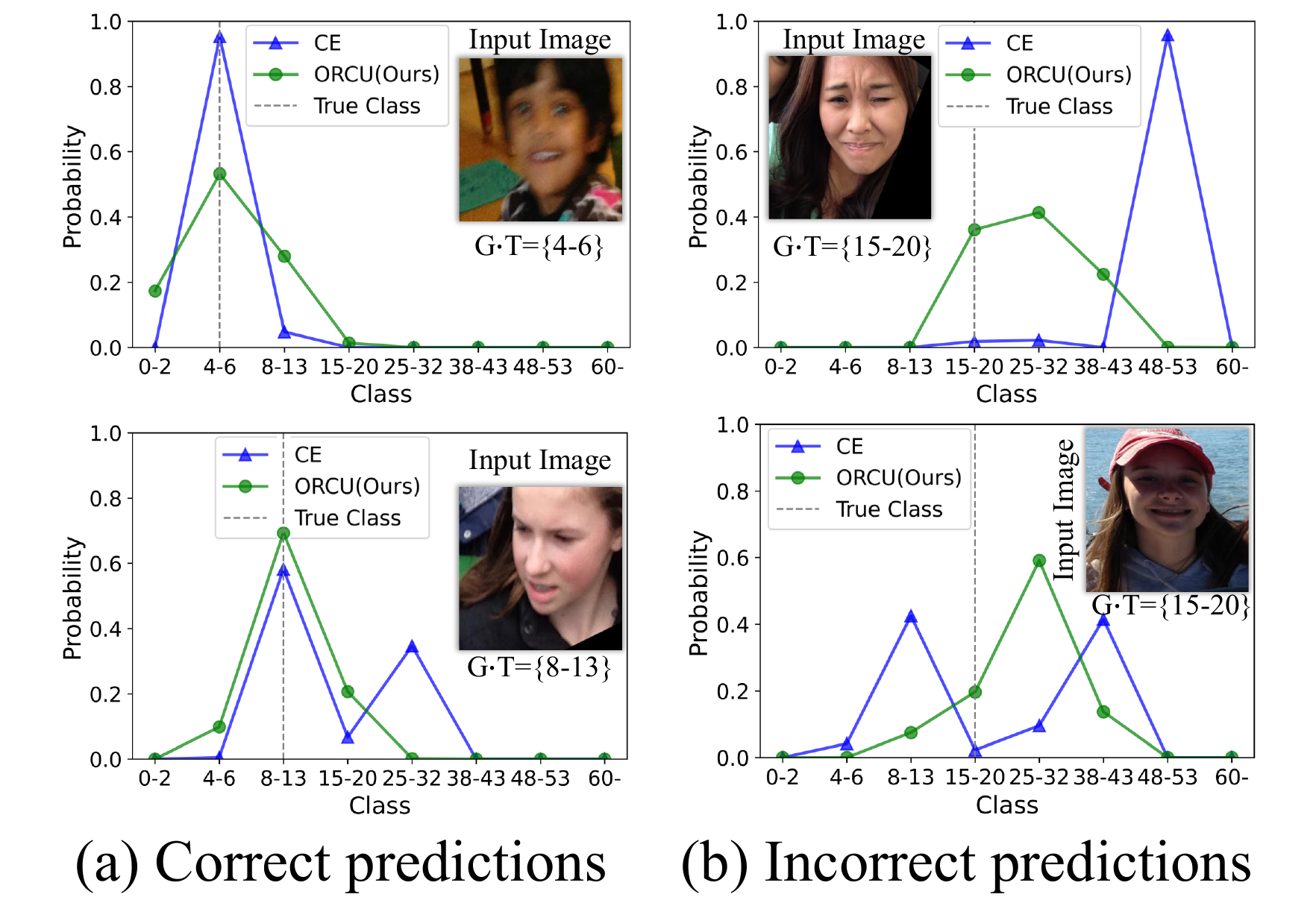}
\caption{Output distributions of models trained with CE and ORCU on the Adience test split, showing (a) correct and (b) incorrect predictions. Each subfigure compares ORCU (green) with CE (blue), with the gray dashed line representing the ground-truth age group.}
\label{suppl_figure:ce_comparison}
\end{figure}

\begin{figure}[!t]
\centering
\includegraphics[width=\columnwidth]{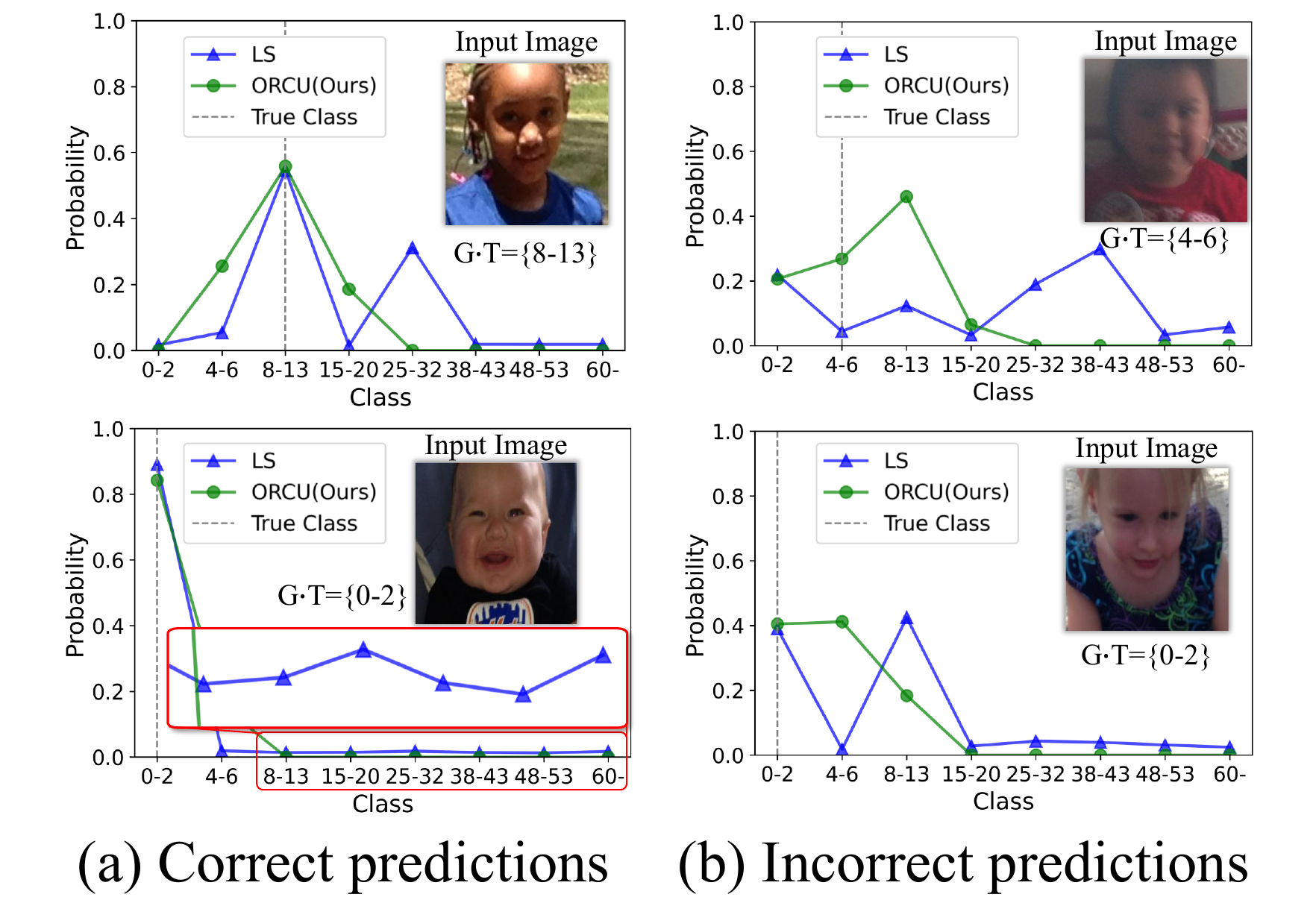}
\caption{Output distributions of models trained with LS and ORCU on the Adience test split, showing (a) correct and (b) incorrect predictions. Each subfigure compares ORCU (green) with LS (blue), with the gray dashed line representing the ground-truth age group.}
\label{suppl_figure:ls_comparison}
\end{figure}

%% file: tables/suppl_tb5_ablation.tex
\begin{table*}[!t]
\renewcommand{\arraystretch}{1.0}
\caption{Full table of ablation studies on the choice of $\mathcal{L}_{\text{REG}}$ and distance metrics used in $\mathcal{L}_{\text{SoftCE}}$, providing detailed results for $\mathcal{L}_{\text{ORCU}}$ and competing regularization methods ($\mathcal{L}_{\text{MbLS}}$, $\mathcal{L}_{\text{MDCA}}$, and $\mathcal{L}_{\text{ACLS}}$). The first row for each dataset corresponds to $\mathcal{L}_{\text{SoftCE}}$ (Squared) as in SORD. Metrics include Static Calibration Error (SCE), Adaptive Calibration Error (ACE), Expected Calibration Error (ECE), and \%Unimodality (\%Unimodal). Symbol \space denotes configurations that remain unchanged from the previous setting.}
\centering
\resizebox{0.72\textwidth}{!}{%
\begin{tabular}{lcccc}
\toprule
\multicolumn{1}{c}{Combinations} & \multicolumn{4}{c}{Evaluation} \\
\cmidrule(lr){2-5}
\makebox[6em][c]{$\mathcal{L}_{\text{SoftCE}{(\phi=\{\})}}$} \makebox[3em]{+} \makebox[6em][c]{$\mathcal{L}_{\{\}}$} & SCE ($10^{-2}$) $\downarrow$ & ACE ($10^{-2}$) $\downarrow$ & ECE (\%) $\downarrow$ & Unimodal(\%) $\uparrow$ \\
\midrule
\multicolumn{5}{c}{Image Aesthetics (\(n=\)13,364)} \\
\makebox[6em][c]{Squared} \makebox[3em]{+} \makebox[6em][c]{-} & 8.49 $\pm$ 0.08 & 8.42 $\pm$ 0.08 & 18.46 $\pm$ 0.26 & 100.0 $\pm$ 0.00 \\
\makebox[6em][c]{} \makebox[3em]{+} \makebox[6em][c]{MbLS} & 8.72 $\pm$ 0.16 & 8.59 $\pm$ 0.14 & 19.21 $\pm$ 0.57 & 100.0 $\pm$ 0.00 \\
\makebox[6em][c]{} \makebox[3em]{+} \makebox[6em][c]{MDCA} & 6.50 $\pm$ 0.23 & 6.44 $\pm$ 0.21 & 15.01 $\pm$ 0.80 & 100.0 $\pm$ 0.00 \\
\makebox[6em][c]{} \makebox[3em]{+} \makebox[6em][c]{ACLS} & 8.59 $\pm$ 0.07 & 8.47 $\pm$ 0.09 & 18.29 $\pm$ 0.81 & 100.0 $\pm$ 0.00 \\
\rowcolor[gray]{.9}\makebox[6em][c]{} \makebox[3em]{+} \makebox[6em][c]{ORCU(Ours)} & 2.82 $\pm$ 0.23 & 3.07 $\pm$ 0.26 & \phantom{0}2.57 $\pm$ 0.40 & 100.0 $\pm$ 0.00 \\
\makebox[6em][c]{Absolute} \makebox[3em]{+} \makebox[6em][c]{} & 3.26 $\pm$ 0.19 & 3.34 $\pm$ 0.19 & \phantom{0}3.29 $\pm$ 0.90 & 100.0 $\pm$ 0.00 \\
\makebox[6em][c]{Huber} \makebox[3em]{+} \makebox[6em][c]{} & 4.04 $\pm$ 0.13 & 4.05 $\pm$ 0.11 & \phantom{0}6.18 $\pm$ 0.60 & 100.0 $\pm$ 0.00 \\
\makebox[6em][c]{Exponential} \makebox[3em]{+} \makebox[6em][c]{} & 5.03 $\pm$ 0.29 & 5.00 $\pm$ 0.24 & \phantom{0}9.65 $\pm$ 1.05 & 99.99 $\pm$ 0.01 \\
\midrule
\multicolumn{5}{c}{Adience (\(n=\)17,423)} \\
\makebox[6em][c]{Squared} \makebox[3em]{+} \makebox[6em][c]{-} & 4.26 $\pm$ 0.77 & 4.33 $\pm$ 0.71 & 7.31 $\pm$ 2.40 & 98.76 $\pm$ 0.95 \\
\makebox[6em][c]{} \makebox[3em]{+} \makebox[6em][c]{MbLS} & 4.29 $\pm$ 0.59 & 4.40 $\pm$ 0.50 & 7.11 $\pm$ 2.17 & 92.12 $\pm$ 1.41 \\
\makebox[6em][c]{} \makebox[3em]{+} \makebox[6em][c]{MDCA} & 4.06 $\pm$ 0.55 & 4.05 $\pm$ 0.55 & 6.77 $\pm$ 3.15 & 97.71 $\pm$ 0.89 \\
\makebox[6em][c]{} \makebox[3em]{+} \makebox[6em][c]{ACLS} & 4.32 $\pm$ 0.65 & 4.39 $\pm$ 0.60 & 7.46 $\pm$ 2.39 & 92.04 $\pm$ 1.80 \\
\rowcolor[gray]{.9}\makebox[6em][c]{} \makebox[3em]{+} \makebox[6em][c]{ORCU(Ours)} & 4.20 $\pm$ 0.16 & 4.04 $\pm$ 0.21 & 8.80 $\pm$ 2.27 & 99.94 $\pm$ 0.04 \\
\makebox[6em][c]{Absolute} \makebox[3em]{+} \makebox[6em][c]{} & 3.96 $\pm$ 0.43 & 3.85 $\pm$ 0.51 & 6.81 $\pm$ 1.85 & 99.98 $\pm$ 0.02 \\
\makebox[6em][c]{Huber} \makebox[3em]{+} \makebox[6em][c]{} & 4.02 $\pm$ 0.65 & 4.00 $\pm$ 0.70 & 4.61 $\pm$ 1.45 & 99.99 $\pm$ 0.01 \\
\makebox[6em][c]{Exponential} \makebox[3em]{+} \makebox[6em][c]{} & 4.92 $\pm$ 1.06 & 4.90 $\pm$ 1.00 & 5.51 $\pm$ 3.41 & 99.97 $\pm$ 0.01 \\
\midrule
\multicolumn{5}{c}{LIMUC (\(n=\)11,276)} \\
\makebox[6em][c]{Squared} \makebox[3em]{+} \makebox[6em][c]{-} & 8.74 $\pm$ 0.30 & 8.55 $\pm$ 0.24 & 16.36 $\pm$ 0.64 & 100.0 $\pm$ 0.00 \\
\makebox[6em][c]{} \makebox[3em]{+} \makebox[6em][c]{MbLS} & 8.69 $\pm$ 0.20 & 8.55 $\pm$ 0.25 & 16.62 $\pm$ 0.56 & 100.0 $\pm$ 0.00 \\
\makebox[6em][c]{} \makebox[3em]{+} \makebox[6em][c]{MDCA} & 7.74 $\pm$ 0.35 & 7.55 $\pm$ 0.39 & 13.73 $\pm$ 0.65 & 100.0 $\pm$ 0.00 \\
\makebox[6em][c]{} \makebox[3em]{+} \makebox[6em][c]{ACLS} & 8.59 $\pm$ 0.29 & 8.40 $\pm$ 0.26 & 15.90 $\pm$ 0.67 & 100.0 $\pm$ 0.00 \\
\rowcolor[gray]{.9}\makebox[6em][c]{} \makebox[3em]{+} \makebox[6em][c]{ORCU(Ours)} & 3.98 $\pm$ 0.41 & 3.62 $\pm$ 0.47 & \phantom{0}6.54 $\pm$ 0.73 & 100.0 $\pm$ 0.00 \\
\makebox[6em][c]{Absolute} \makebox[3em]{+} \makebox[6em][c]{} & 5.09 $\pm$ 0.30 & 4.67 $\pm$ 0.28 & \phantom{0}6.35 $\pm$ 1.28 & 100.0 $\pm$ 0.00 \\
\makebox[6em][c]{Huber} \makebox[3em]{+} \makebox[6em][c]{} & 6.95 $\pm$ 0.41 & 6.57 $\pm$ 0.41 & 11.22 $\pm$ 1.37 & 100.0 $\pm$ 0.00 \\
\makebox[6em][c]{Exponential} \makebox[3em]{+} \makebox[6em][c]{} & 8.81 $\pm$ 0.42 & 8.67 $\pm$ 0.43 & 15.24 $\pm$ 1.30 & 99.99 $\pm$ 0.01 \\
\midrule
\multicolumn{5}{c}{Diabetic Retinopathy (\(n=\)9,570)} \\
\makebox[6em][c]{Squared} \makebox[3em]{+} \makebox[6em][c]{-} & 4.08 $\pm$ 0.17 & 4.17 $\pm$ 0.19 & 4.27 $\pm$ 0.23 & 99.90 $\pm$ 0.04 \\
\makebox[6em][c]{} \makebox[3em]{+} \makebox[6em][c]{MbLS} & 3.95 $\pm$ 0.24 & 3.99 $\pm$ 0.18 & 4.65 $\pm$ 0.66 & 99.27 $\pm$ 0.18 \\
\makebox[6em][c]{} \makebox[3em]{+} \makebox[6em][c]{MDCA} & 4.06 $\pm$ 0.30 & 4.10 $\pm$ 0.22 & 4.71 $\pm$ 1.08 & 99.93 $\pm$ 0.02 \\
\makebox[6em][c]{} \makebox[3em]{+} \makebox[6em][c]{ACLS} & 3.99 $\pm$ 0.29 & 4.13 $\pm$ 0.25 & 5.07 $\pm$ 1.10 & 99.43 $\pm$ 0.17 \\
\rowcolor[gray]{.9}\makebox[6em][c]{} \makebox[3em]{+} \makebox[6em][c]{ORCU(Ours)} & 3.75 $\pm$ 0.43 & 3.67 $\pm$ 0.30 & 3.22 $\pm$ 0.80 & 99.96 $\pm$ 0.02 \\
\makebox[6em][c]{Absolute} \makebox[3em]{+} \makebox[6em][c]{} & 4.22 $\pm$ 0.53 & 4.15 $\pm$ 0.45 & 4.00 $\pm$ 0.38 & 99.98 $\pm$ 0.02 \\
\makebox[6em][c]{Huber} \makebox[3em]{+} \makebox[6em][c]{} & 4.20 $\pm$ 0.18 & 4.28 $\pm$ 0.25 & 4.57 $\pm$ 1.28 & 99.98 $\pm$ 0.01 \\
\makebox[6em][c]{Exponential} \makebox[3em]{+} \makebox[6em][c]{} & 5.35 $\pm$ 0.27 & 5.32 $\pm$ 0.29 & 6.68 $\pm$ 1.74 & 99.98 $\pm$ 0.01 \\
\bottomrule
\end{tabular}
}
\label{suppl_table:ablation}
\end{table*}

%% file: figures/tex/suppl_fig6_gradcam.tex
\begin{figure*}[!t]
\begin{center}
\includegraphics[width=0.8\textwidth]{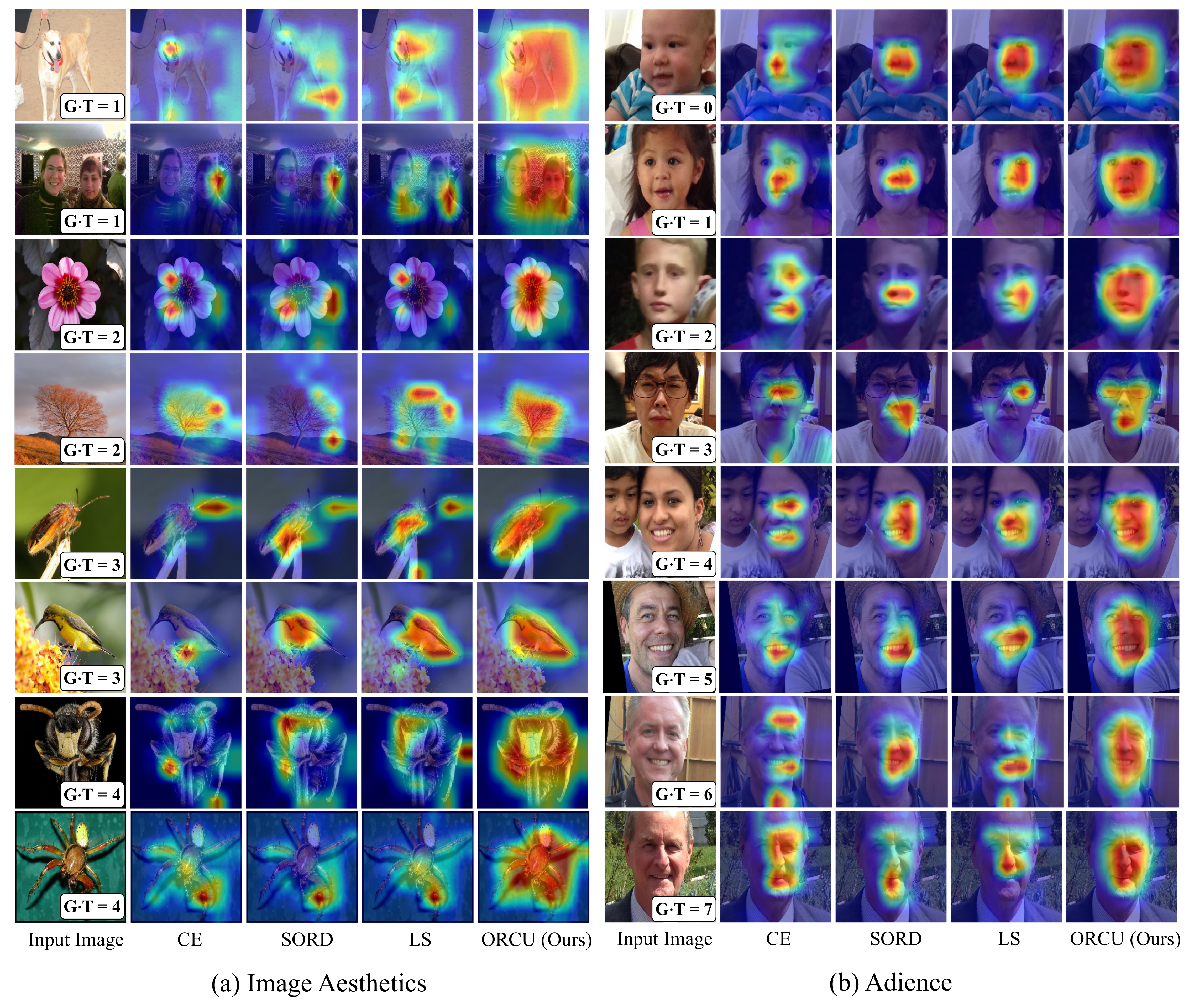}
\end{center}
\caption{Grad-CAM visualizations from the final convolutional layer of ResNet-50 trained with CE, SORD, LS, and \(\mathcal{L}_{\text{ORCU}}\) on (a) Image Aesthetics and (b) Adience datasets, representing a general computer vision context. Activation maps indicate regions contributing to model predictions, with colors showing activation intensity (red: high, blue: low). Visualizations are derived from correctly predicted samples.}
\label{suppl_figure:gradcam}
\end{figure*}

%% file: figures/tex/suppl_fig7_valcurve.tex
\setlength{\intextsep}{10pt}%
\setlength{\columnsep}{10pt}%

\begin{figure*}[!t]
\begin{center}
\includegraphics[width=0.95\textwidth]{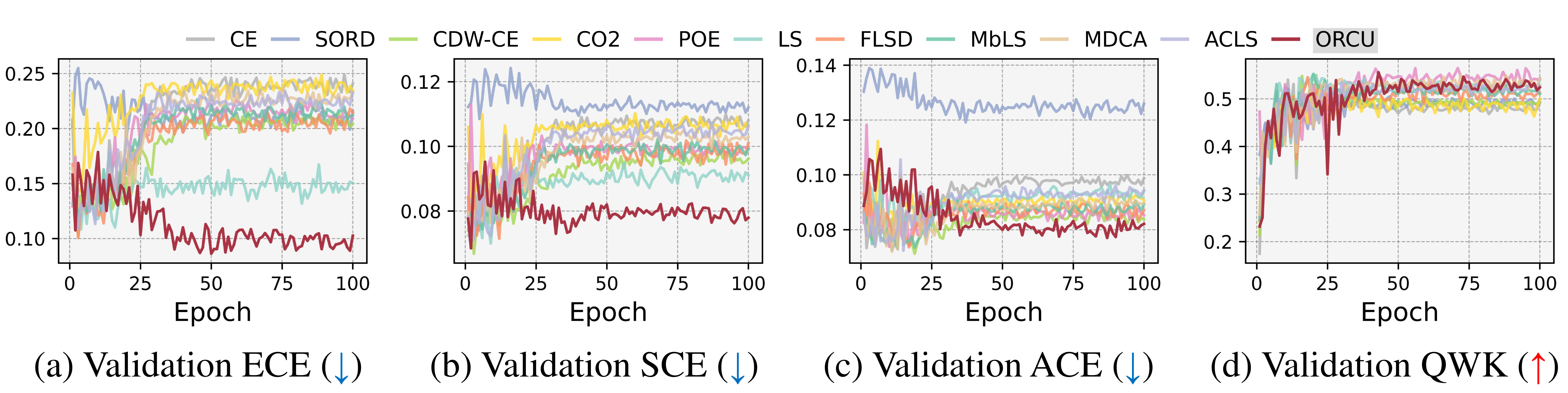}
\end{center}
\vspace{-0.3cm}
\caption{
Validation curves of ECE, SCE, ACE, and QWK on the Image Aesthetics dataset, illustrating the convergence behavior of different loss functions.
}
\label{suppl_figure:validation_curve}
\end{figure*}